\documentclass{article}

\usepackage{arxiv,times}
\iclrfinalcopy
\usepackage{amsmath,amsfonts,bm}

\def\eqref#1{equation~\ref{#1}}
\def\1{\bm{1}}

\DeclareMathAlphabet{\mathsfit}{\encodingdefault}{\sfdefault}{m}{sl}
\SetMathAlphabet{\mathsfit}{bold}{\encodingdefault}{\sfdefault}{bx}{n}

\usepackage[utf8]{inputenc} % allow utf-8 input
\usepackage[T1]{fontenc}    % use 8-bit T1 fonts
\usepackage{xr-hyper}
\usepackage{hyperref}
\usepackage{url}            % simple URL typesetting
\usepackage{amsfonts}       % blackboard math symbols
\usepackage{nicefrac}       % compact symbols for 1/2, etc.
\usepackage{microtype}
\usepackage{xcolor}

\usepackage{graphicx}
\usepackage{subfigure}
\usepackage{booktabs} % for professional tables
\usepackage{stackengine} % for stackunder
\usepackage{float} % for [H]
\usepackage{wrapfig} % wrapfigure

\usepackage{multirow}
\usepackage{array}
\newcolumntype{P}[1]{>{\centering\arraybackslash}p{#1}}
\usepackage{arydshln}

\usepackage{amsmath}
\mathchardef\mhyphen="2D % compact hyphen in the math mode
\usepackage{dsfont} % for indicator function

\definecolor{medblue}{rgb}{0,0,.75}
\definecolor{burntorange}{rgb}{0.8, 0.33, 0.0}
\usepackage[noend]{algpseudocode}
\usepackage{algorithm,algorithmicx}
\algrenewcommand\alglinenumber[1]{\sf\tiny\color{medblue}{#1}\quad}
\algrenewcommand\algorithmicrequire{\textbf{Input:}}
\algrenewcommand\algorithmicensure{\textbf{Output:}}
\algdef{SE}[DOWHILE]{Do}{doWhile}{\algorithmicdo}[1]{\algorithmicwhile\ #1}%

\usepackage{amsmath}
\usepackage{amssymb}
\usepackage{mathtools}
\usepackage{amsthm}

\usepackage{enumitem} % format the enumerated items

\usepackage[capitalize,noabbrev]{cleveref}

\theoremstyle{plain}

\theoremstyle{definition}

\theoremstyle{remark}

\newcommand{\name}{OPRO}

\usepackage[textsize=tiny]{todonotes}

\makeatletter
\newcommand*{\addFileDependency}[1]{% argument=file name and extension
  \typeout{(#1)}% latexmk will find this if $recorder=0 (however, in that case, it will ignore #1 if it is a .aux or .pdf file etc and it exists! if it doesn't exist, it will appear in the list of dependents regardless)
  \@addtofilelist{#1}% if you want it to appear in \listfiles, not really necessary and latexmk doesn't use this
  \IfFileExists{#1}{}{\typeout{No file #1.}}% latexmk will find this message if #1 doesn't exist (yet)
}
\makeatother

\newcommand{\myparagraph}[1]{\noindent\textbf{#1.}\,}

\definecolor{purple}{rgb}{0.3,0.0,.4}

\def \figurepath {figures/}

\title{Large Language Models as Optimizers}

\author{
Chengrun Yang\textsuperscript{*} \hspace{.5em}
Xuezhi Wang \hspace{.5em}
Yifeng Lu \hspace{.5em}
Hanxiao Liu \hspace{.5em}\\
\textbf{
Quoc V. Le \hspace{.5em}
Denny Zhou \hspace{.5em} 
Xinyun Chen\textsuperscript{*}
}
\\
[1ex]
\texttt{\{chengrun, xuezhiw, yifenglu, hanxiaol\}@google.com} \\
\texttt{\{qvl, dennyzhou, xinyunchen\}@google.com}\\
[1ex]
Google DeepMind \quad \textsuperscript{*} Equal contribution\\
}

\renewcommand{\headrulewidth}{1pt}
\fancypagestyle{firstpage}{
  \lhead{\begin{picture}(0,0)\put(0,-3){\includegraphics[width=0.25\linewidth]{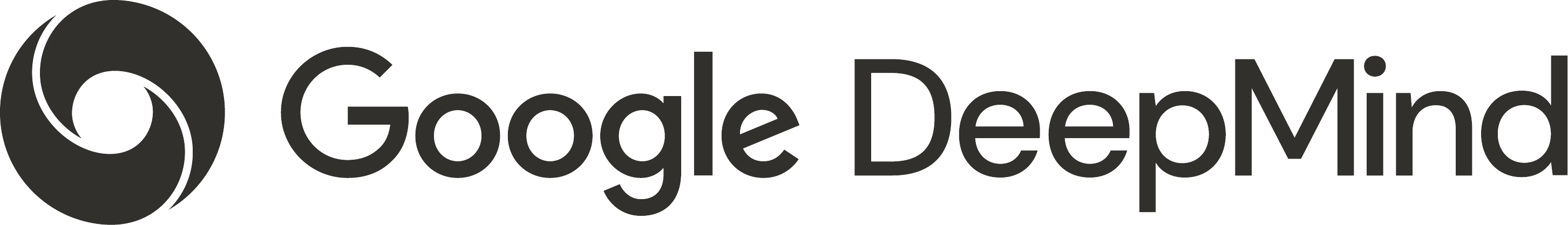}}\end{picture}}
}

\begin{document}
\maketitle
\thispagestyle{firstpage}
\vspace{-1em}
\begin{abstract}
Optimization is ubiquitous. While derivative-based algorithms have been powerful tools for various problems, the absence of gradient imposes challenges on many real-world applications.
In this work, we propose Optimization by PROmpting (\name{}), a simple and effective approach to leverage large language models (LLMs) as optimizers, where the optimization task is described in natural language. In each optimization step, the LLM generates new solutions from the prompt that contains previously generated solutions with their values, then the new solutions are evaluated and added to the prompt for the next optimization step. We first showcase \name{} on linear regression and traveling salesman problems, then move on to our main application in prompt optimization, where the goal is to find instructions that maximize the task accuracy. With a variety of LLMs, we demonstrate that the best prompts optimized by \name{} outperform human-designed prompts by up to $8\%$ on GSM8K, and by up to $50\%$ on Big-Bench Hard tasks.
Code at \url{https://github.com/google-deepmind/opro}.
\end{abstract}

\vspace{-1em}
\begin{figure}[H]
\centering
\scalebox{0.79}{
\subfigure[GSM8K]{\label{fig:prompt_optimization_graph_in_intro_gsm8k}\includegraphics[width=.4\linewidth]{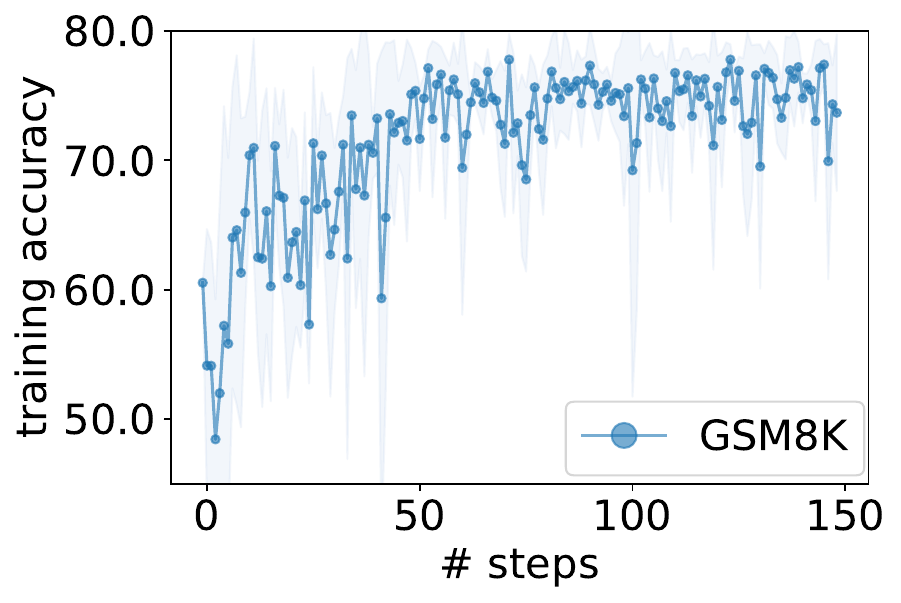}}
\hspace{.01\linewidth}
\subfigure[BBH movie\_recommendation]{\label{fig:prompt_optimization_graph_in_intro_bbh}\includegraphics[width=.43\linewidth]{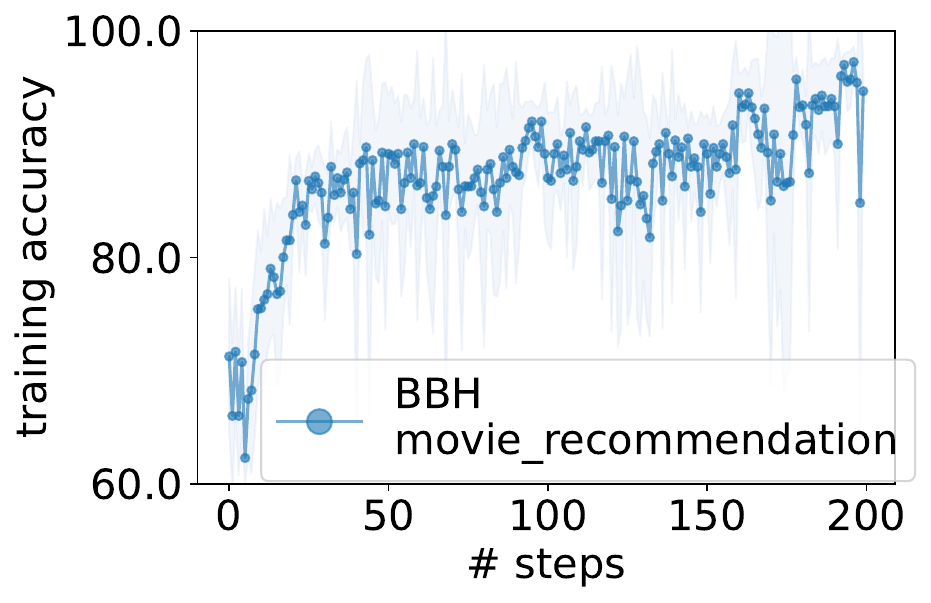}}
}
\caption{Prompt optimization on GSM8K~\citep{cobbe2021training} and BBH~\citep{suzgun2022challenging} movie\_recommendation. 
The optimization on GSM8K has pre-trained \texttt{PaLM 2-L} as the scorer and the instruction-tuned \texttt{PaLM 2-L} (denoted \texttt{PaLM 2-L-IT}) as the optimizer; the optimization on BBH movie\_recommendation has \texttt{text-bison} as the scorer and \texttt{PaLM 2-L-IT} as the optimizer.
Each dot is the average accuracy across all (up to 8) generated instructions in the single step, and the shaded region represents standard deviation.
See Section~\ref{sec:exp} for more details on experimental setup.}
\label{fig:prompt_optimization_graph_in_intro}
\end{figure}

\vspace{-1em}
\begin{table}[H]
\caption{Top instructions with the highest GSM8K zero-shot test accuracies from prompt optimization with different optimizer LLMs. All results use the pre-trained \texttt{PaLM 2-L} as the scorer. 
}
\centering
\scalebox{0.81}{
\begin{tabular}{P{3.5cm}P{10.5cm}P{1cm}}
\toprule
Source & Instruction & Acc \\
\midrule
\multicolumn{3}{l}{\textit{Baselines}} \\
\hdashline\noalign{\vskip 0.5ex}
\citep{kojima2022large} & Let's think step by step. & 71.8 \\
\citep{zhou2022large} & Let’s work this out in a step by step way to be sure we have the right answer. & 58.8 \\
& (empty string) & 34.0 \\
\midrule
\multicolumn{3}{l}{\textit{Ours}} \\
\hdashline\noalign{\vskip 0.5ex}
\texttt{PaLM 2-L-IT} & Take a deep breath and work on this problem step-by-step. & \textbf{80.2} \\
\texttt{PaLM 2-L} & Break this down. & 79.9 \\
\texttt{gpt-3.5-turbo} & A little bit of arithmetic and a logical approach will help us quickly arrive at the solution to this problem. & 78.5 \\
\texttt{gpt-4} & Let's combine our numerical command and clear thinking to quickly and accurately decipher the answer. & 74.5\\
\bottomrule
\end{tabular}}
\label{table:gsm8k_top_instructions_in_intro}
\end{table}

\section{Introduction}
\label{sec:intro}

Optimization is critical for all areas. Many optimization techniques are iterative: the optimization starts from an initial solution, then iteratively updates the solution to optimize the objective function~\citep{amari1993backpropagation,qian1999momentum,kingma2014adam,back1993overview,rios2013derivative,reeves1993modern}.
The optimization algorithm typically needs to be customized for an individual task to deal with the specific challenges posed by the decision space and the performance landscape, especially for derivative-free optimization.

In this work, we propose Optimization by PROmpting (\name{}), a simple and effective approach to utilize large language models (LLMs) as optimizers.
With the advancement of prompting techniques, LLMs have achieved impressive performance in various domains~\citep{wei2022chain,kojima2022large,wang2022self,zhou2022least,madaan2023self,bai2022constitutional,chen2023teaching}.
Their ability to understand natural language lays out a new possibility for optimization: instead of formally defining the optimization problem and deriving the update step with a programmed solver, we describe the optimization problem in natural language, then instruct the LLM to iteratively generate new solutions based on the problem description and the previously found solutions.
Optimization with LLMs enables quick adaptation to different tasks by changing the problem description in the prompt, and the optimization process can be customized by adding instructions to specify the desired properties of the solutions.

To demonstrate the potential of LLMs for optimization, we first present case studies on linear regression and the traveling salesman problem, which are two classic optimization problems that underpin many others in mathematical optimization, computer science, and operations research.
On small-scale optimization problems, we show that LLMs are able to find good-quality solutions simply through prompting, and sometimes match or surpass hand-designed heuristic algorithms.

Next, we demonstrate the ability of LLMs to optimize prompts: the goal is to find a prompt that maximizes the task accuracy. 
Specifically, we focus on natural language tasks where both the task input and output are texts. 
LLMs are shown to be sensitive to the prompt format~\citep{zhao2021calibrate,lu2021fantastically,wei2023larger,madaan2022text}; in particular, semantically similar prompts may have drastically different performance~\citep{kojima2022large,zhou2022large,zhang2022tempera}, and the optimal prompt formats can be model-specific and task-specific~\citep{ma2023let,chen2023you}.
Therefore, prompt engineering is often important for LLMs to achieve good performance~\citep{reynolds2021prompt}.
However, the large and discrete prompt space makes it challenging for optimization, especially when only API access to the LLM is available. 
Following prior work on continuous and discrete prompt optimization~\citep{lester2021power,li2021prefix,zhou2022large,pryzant2023automatic}, we assume a training set is available to compute the training accuracy as the objective value for optimization, and we show in experiments that optimizing the prompt for accuracy on a small training set is sufficient to reach high performance on the test set.

The prompt to the LLM serves as a call to the optimizer, and we name it the \emph{meta-prompt}. 
Figure~\ref{fig:meta_prompt_example} shows an example.
The meta-prompt contains two core pieces of information.
The first piece is previously generated prompts with their corresponding training accuracies. 
The second piece is the optimization problem description, which includes several exemplars randomly selected from the training set to exemplify the task of interest.
We also provide instructions for the LLM to understand the relationships among different parts and the desired output format.
Different from recent work on using LLMs for automatic prompt generation~\citep{zhou2022large,pryzant2023automatic}, each optimization step in our work \emph{generates} new prompts that aim to increase the test accuracy based on a trajectory of previously generated prompts, instead of \emph{editing} one input prompt according to natural language feedback~\citep{pryzant2023automatic} or requiring the new prompt to follow the same semantic meaning~\citep{zhou2022large}.
Making use of the full optimization trajectory, \name{} enables the LLM to gradually generate new prompts that improve the task accuracy throughout the optimization process, where the initial prompts have low task accuracies.

We conduct comprehensive evaluation on several LLMs, including \texttt{text-bison} and \texttt{Palm 2-L} in the PaLM-2 model family~\citep{anil2023palm}, as well as \texttt{gpt-3.5-turbo} and \texttt{gpt-4} in the GPT model family. 
We optimize prompts on GSM8K~\citep{cobbe2021training} and Big-Bench Hard~\citep{suzgun2022challenging}, which are reasoning benchmarks where prompting techniques have achieved remarkable performance breakthrough~\citep{wei2022chain,kojima2022large,suzgun2022challenging}. Starting from initial prompts with low task accuracies, we show that all LLMs in our evaluation are able to serve as optimizers, which consistently improve the performance of the generated prompts through iterative optimization until convergence (see Figure~\ref{fig:prompt_optimization_graph_in_intro}). 
In particular, while these LLMs generally produce instructions of different styles (see Table~\ref{table:gsm8k_top_instructions_in_intro}), with zero-shot prompting, their best generated instructions match the few-shot chain-of-thought prompting performance when applied to \texttt{PaLM 2-L}, outperforming the zero-shot performance with human-designed prompts by up to $8\%$ on GSM8K.
Additionally, we observe that the \name{}-optimized prompts transfer to other benchmarks of the same domain and also deliver notable performance gain.
\section{\name{}: LLM as the Optimizer}
\label{sec:approach}

\begin{figure}
\centering
\subfigure{\includegraphics[width=.55\linewidth]{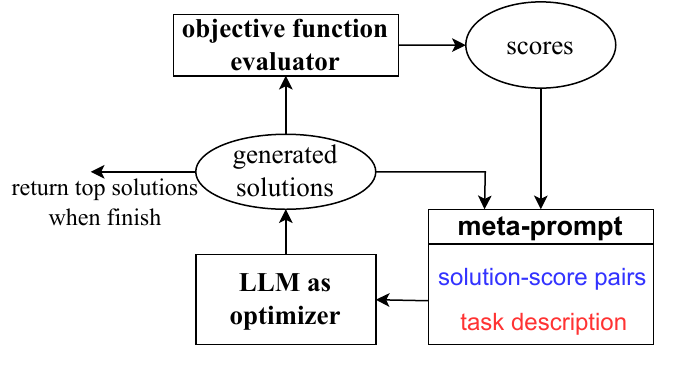}}
\caption{An overview of the \name{} framework. Given the meta-prompt as the input, the LLM generates new solutions to the objective function, then the new solutions and their scores are added into the meta-prompt for the next optimization step. The meta-prompt contains the solution-score pairs obtained throughout optimization, a natural language description of the task, and (in prompt optimization) a few task exemplars. Figure~\ref{fig:meta_prompt_example} shows a sample meta-prompt for prompt optimization.
}
\vspace{-.5em}
\label{fig:overview}
\end{figure}

Figure~\ref{fig:overview} illustrates the overall framework of \name{}. 
In each optimization step, the LLM generates candidate solutions to the optimization task based on the optimization problem description and previously evaluated solutions in the meta-prompt. 
Then the new solutions are evaluated and added to the meta-prompt for the subsequent optimization process. 
The optimization process terminates when the LLM is unable to propose new solutions with better optimization scores, or a maximum number of optimization steps has reached. 
We first outline the desired features of LLMs for optimization, then describe the key design choices based on these desirables.

\subsection{Desirables of Optimization by LLMs}

\myparagraph{Making use of natural language descriptions} The main advantage of LLMs for optimization is their ability of understanding natural language, which allows people to describe their optimization tasks without formal specifications. 
For instance, in prompt optimization where the goal is to find a prompt that optimizes the task accuracy, the task can be described with a high-level text summary along with input-output examples.

\myparagraph{Trading off exploration and exploitation} The exploration-exploitation trade-off is a fundamental challenge in optimization, and it is important for LLMs serving as optimizers to balance these two competing goals. 
This means that the LLM should be able to exploit promising areas of the search space where good solutions are already found, while also exploring new regions of the search space so as to not miss potentially better solutions.

\subsection{Meta-prompt Design}

As the input to the optimizer LLM, the meta-prompt contains the following two essential parts.

\myparagraph{Optimization problem description} 
The first part is the text description of the optimization problem, including the objective function and solution constraints. 
For example, for prompt optimization, the LLM can be instructed to ``generate a new instruction that achieves a higher accuracy'', and we denote such instructions in the meta-prompt as \emph{meta-instructions}. 
We can also provide customized meta-instructions as an informal regularization of the generated solutions, such as ``the instruction should be concise and generally applicable''.

\myparagraph{Optimization trajectory} 
Besides understanding natural language instructions, LLMs are also shown to be able to recognize patterns from in-context demonstrations~\citep{wei2023larger,madaan2022text,mirchandani2023large}. 
Our meta-prompt makes use of this property and instructs the LLM to leverage the optimization trajectory for generating new solutions. 
Specifically, the optimization trajectory includes past solutions and their optimization scores, sorted in the ascending order. 
Including optimization trajectory in the meta-prompt allows the LLM to identify similarities of solutions with high scores, encouraging the LLM to build upon existing good solutions to construct potentially better ones without the need of explicitly defining how the solution should be updated.

\subsection{Solution Generation}

At the solution generation step, the LLM generates new solutions with the meta-prompt as input. 
The following are the key optimization challenges we address in this stage.

\myparagraph{Optimization stability} In the optimization process, not all solutions achieve high scores and monotonically improve over prior ones. 
Due to the sensitivity of in-context learning to the prompt, LLM output can be drastically affected by low-quality solutions in the input optimization trajectory, especially at the beginning when the solution space has not been adequately explored. 
This sometimes results in optimization instability and large variance. 
To improve stability, we prompt the LLM to generate multiple solutions at each optimization step, allowing the LLM to simultaneously explore multiple possibilities and quickly discover promising directions to move forward.

\myparagraph{Exploration-exploitation trade-off} 
We tune the LLM sampling temperature to balance between exploration and exploitation. 
A lower temperature encourages the LLM to exploit the solution space around the previously found solutions and make small adaptations, while a high temperature allows the LLM to more aggressively explore solutions that can be notably different.
\section{Motivating Example: Mathematical Optimization}
\label{sec:motivating_example}

We first demonstrate the potential of LLMs in serving as optimizers for mathematical optimization. 
In particular, we present a case study on linear regression as an example of continuous optimization, and on the Traveling Salesman Problem (TSP) as an example of discrete optimization. 
On both tasks, we see LLMs properly capture the optimization directions on small-scale problems merely based on the past optimization trajectory provided in the meta-prompt.

\subsection{Linear Regression}

In linear regression problems, the goal is to find the linear coefficients that probabilistically best explain the response from the input variables.
We study the setting in which the independent and dependent variables $X$ and $y$ are both one-dimensional and an intercept $b$ is present, so that there are two one-dimensional variables $w$, $b$ to optimize over.
In a synthetic setting, we sample ground truth values for one-dimensional variables $w_\text{true}$ and $b_\text{true}$, and generate 50 data points by $y = w_\text{true} x + b_\text{true} + \epsilon$, in which $x$ ranges from 1 to 50 and $\epsilon$ is the standard Gaussian noise.
Our optimization starts from 5 randomly sampled $(w, b)$ pairs.
In each step, we prompt an instruction-tuned LLM with a meta-prompt that includes the best 20 $(w, b)$ pairs in history and their sorted objective values.
The meta-prompt then asks for a new $(w, b)$ pair that further decreases the objective value.
A sample meta-prompt is shown in Figure~\ref{fig:meta_prompt_example_linear_regression} of Appendix~\ref{appsec:meta_prompts_for_math_opt}.
We prompt the meta-prompt 8 times to generate at most 8 new $(w, b)$ pairs in each step to improve optimization stability.
Then we evaluate the objective value of the proposed pair and add it to history.
We do black-box optimization: the analytic form does not appear in the meta-prompt text.
This is because the LLM can often calculate the solution directly from the analytic form.

\begin{table}
\centering
\caption{Linear regression by optimizer LLMs: the mean $\pm$ standard deviation of the number of steps and the number of unique $(w, b)$ pairs explored before reaching the global optima.
Both $w$ and $b$ start from 5 random starting points in $[10, 20]$.
We use temperature 1.0 for all models.
We run each setting 5 times.
The starting points are the same across optimizer LLMs but are different across 5 runs, and are grouped by: within the starting region, outside and close to the starting region, and outside and farther from the starting region.
Bold numbers indicate the best among three LLMs in each setting.
}
\scalebox{0.8}{
\begin{tabular}{cccccccc}
\toprule
\multirow{2}{*}{$w_\text{true}$} & \multirow{2}{*}{$b_\text{true}$} & \multicolumn{3}{c}{number of steps} & \multicolumn{3}{c}{number of unique $(w, b)$ pairs explored} \\ \cmidrule(lr){3-5} \cmidrule(lr){6-8}
& & \texttt{text-bison} & \texttt{gpt-3.5-turbo} & \texttt{gpt-4} & \texttt{text-bison} & \texttt{gpt-3.5-turbo} & \texttt{gpt-4} \\
\midrule
15 & 14 & 5.8 \scriptsize{$\pm$ 2.6} & 7.6 \scriptsize{$\pm$ 4.5} & \textbf{4.0} \scriptsize{$\pm$ 1.5} & 40.0 \scriptsize{$\pm$ 12.4}
 & 36.0 \scriptsize{$\pm$ 15.2} & \textbf{17.2} \scriptsize{$\pm$ 5.1} \\
17 & 17 & \textbf{4.0} \scriptsize{$\pm$ 1.8} & 12.6 \scriptsize{$\pm$ 6.0} & 6.0 \scriptsize{$\pm$ 3.7} & 33.4 \scriptsize{$\pm$ 11.7} & 53.8 \scriptsize{$\pm$ 16.9} & \textbf{26.0} \scriptsize{$\pm$ 10.6} \\
16 & 10 & \textbf{3.8} \scriptsize{$\pm$ 2.2} & 10.4 \scriptsize{$\pm$ 5.4} & 6.2 \scriptsize{$\pm$ 3.1} & 30.2 \scriptsize{$\pm$ 13.4} & 42.8 \scriptsize{$\pm$ 16.3} & \textbf{24.2} \scriptsize{$\pm$ 8.2} \\
\hdashline\noalign{\vskip 0.5ex}
3 & 5 & \textbf{9.8} \scriptsize{$\pm$ 2.8} & 10.8 \scriptsize{$\pm$ 2.7} & 12.2 \scriptsize{$\pm$ 2.0} & 55.8 \scriptsize{$\pm$ 16.1} & 39.6 \scriptsize{$\pm$ 10.1} & \textbf{33.0} \scriptsize{$\pm$ 4.0} \\
25 & 23 & 19.6 \scriptsize{$\pm$ 11.4} & 26.4 \scriptsize{$\pm$ 18.3} & \textbf{12.2} \scriptsize{$\pm$ 3.7} & 104.0 \scriptsize{$\pm$ 52.3} & 78.6 \scriptsize{$\pm$ 26.2} & \textbf{44.2} \scriptsize{$\pm$ 8.3} \\
\hdashline\noalign{\vskip 0.5ex}
2 & 30 & \textbf{31.4} \scriptsize{$\pm$ 6.3} & 42.8 \scriptsize{$\pm$ 9.7} & 38.0 \scriptsize{$\pm$ 15.9} & 126.4 \scriptsize{$\pm$ 17.7} & 125.6 \scriptsize{$\pm$ 21.7} & \textbf{99.0} \scriptsize{$\pm$ 24.6} \\
36 & -1 & \textbf{35.8} \scriptsize{$\pm$ 6.4} & 45.4 \scriptsize{$\pm$ 16.9} & 50.4 \scriptsize{$\pm$ 18.8} & 174.0 \scriptsize{$\pm$ 28.2} & 142.2 \scriptsize{$\pm$ 31.2} & \textbf{116.4} \scriptsize{$\pm$ 32.7} \\
\bottomrule
\end{tabular}
}
\label{table:linear_regression_results_in_main_paper}
\end{table}

Table~\ref{table:linear_regression_results_in_main_paper} summarizes the results with one of the following optimizer LLMs: \texttt{text-bison}, \texttt{gpt-3.5-turbo}, and \texttt{gpt-4}.
We study three settings of $w_\text{true}$ and $b_\text{true}$: within the starting region $[10, 20] \times [10, 20]$, ``near outside'' (each of $w_\text{true}$ and $b_\text{true}$ is outside the starting region but the distance is less than 10), and ``far outside'' (each of $w_\text{true}$ and $b_\text{true}$ is outside the starting region and the distance is greater than 10).
We see:
\begin{itemize}[leftmargin=2em,topsep=0pt,partopsep=1ex,parsep=0ex]
\item The number of unique $(w, b)$ pairs explored by each model is fewer than exhaustive search, indicating these models are able to to do black-box optimization: compare the numbers and propose a descent direction.
\item The \texttt{text-bison} and \texttt{gpt-4} models outperform \texttt{gpt-3.5-turbo} in convergence speed: they arrive at the optima with fewer steps.
The \texttt{gpt-4} model also outperforms in finding the optima with fewer explored unique points. 
Taking a closer look at the optimization trajectory, we see \texttt{gpt-4} is the best at proposing a reasonable next step from the history: for example, when the history shows the objective values of $(w, b) = (8, 7)$, $(w, b) = (8, 6)$, and $(w, b) = (8, 5)$ are decreasing, it has a highest chance to propose $(w, b) = (8, 4)$ for evaluation.
\item The problem becomes harder for all models when the ground truth moves farther from the starting region: all models need more explorations and more steps.
\end{itemize}

\subsection{Traveling Salesman Problem (TSP)}
Next, we consider the Traveling Salesman Problem (TSP)~\citep{junger1995traveling,gutin2006traveling}, a classical combinatorial optimization problem with numerous algorithms proposed in literature, including heuristic algorithms and solvers~\citep{rosenkrantz1977analysis,golden1980approximate,optimization2020gurobi,applegate2006concorde,helsgaun2017extension}, and approaches based on training deep neural networks~\citep{kool2018attention,deudon2018learning,chen2019learning,nazari2018reinforcement}. 
Specifically, given a set of $n$ nodes with their coordinates, the TSP task is to find the shortest route that traverses all nodes from the starting node and finally returns to the starting node.

Our optimization process with LLMs starts from 5 randomly generated solutions, and each optimization step produces at most 8 new solutions. 
We present the meta-prompt in Figure~\ref{fig:meta_prompt_example_tsp} of Appendix~\ref{appsec:meta_prompts_for_math_opt}. 
We generate the problem instances by sampling $n$ nodes with both $x$ and $y$ coordinates in $[-100, 100]$. 
We use the Gurobi solver~\citep{optimization2020gurobi} to construct the oracle solutions and compute the optimality gap for all approaches, where the optimality gap is defined as the difference between the distance in the solution constructed by the evaluated approach and the distance achieved by the oracle solution, divided by the distance of the oracle solution. 
Besides evaluating \name{} with different LLMs including \texttt{text-bison}, \texttt{gpt-3.5-turbo} and \texttt{gpt-4}, we also compare \name{} to the following heuristics:

\begin{itemize}[leftmargin=2em,topsep=0pt,partopsep=1ex,parsep=0ex]

\item \texttt{Nearest Neighbor (NN)}. Starting from an initial node, the solution is constructed with the nearest neighbor heuristic: At each step, among the remaining nodes that are not included in the current partial solution, NN selects the node with the shortest distance to the end node of the partial solution, and adds it as the new end node. The process finishes when all nodes have been added to the solution.

\item \texttt{Farthest Insertion (FI)}. One caveat of the nearest neighbor heuristic is that it does not take the distance between the start and end node into consideration when constructing partial solutions. To address this issue, FI aims to optimize the cost of inserting new nodes into the partial solution at each step. Define the minimal insertion cost of adding a new node $k$ as $c(k) = \min_{(i, j)} d(i, k) + d(k, j) - d(i, j)$, where $i$ and $j$ are adjacent nodes in the current tour, and $d(\cdot, \cdot)$ represents the distance between two nodes. At each step, FI adds a new node that maximizes the minimal insertion cost.

\end{itemize}

\begin{table}
\centering
\caption{Results of the Traveling Salesman Problem (TSP) with different number of nodes $n$, where each $n$ contains 5 problems. ``\# steps'' calculates the mean $\pm$ standard error of optimization steps for successful runs that find the optimal solution. ``\# successes'' counts the number of problems that \name{} results in the optimal solution. When no optimal solution is found for any evaluated problem, the corresponding number of steps is N/A.
}
\scalebox{0.7}{
\begin{tabular}{ccccccccc}
\toprule
\multirow{2}{*}{$n$} & \multicolumn{5}{c}{optimality gap (\%)} & \multicolumn{3}{c}{\# steps (\# successes)} \\ \cmidrule(lr){2-6} \cmidrule(lr){7-9}
& NN & FI & \texttt{text-bison} & \texttt{gpt-3.5-turbo} & \texttt{gpt-4} & \texttt{text-bison} & \texttt{gpt-3.5-turbo} & \texttt{gpt-4} \\
\midrule
10 & 13.0 \scriptsize{$\pm$ 1.3} & 3.2 \scriptsize{$\pm$ 1.4} & \textbf{0.0} \scriptsize{$\pm$ 0.0} & \textbf{0.0} \scriptsize{$\pm$ 0.0} & \textbf{0.0} \scriptsize{$\pm$ 0.0} & 40.4 \scriptsize{$\pm$ 5.6} \textbf{\footnotesize{ (5)}} & 46.8 \scriptsize{$\pm$ 9.3} \textbf{\footnotesize{ (5)}} & \textbf{9.6} \scriptsize{$\pm$ 3.0} \textbf{\footnotesize{ (5)}}\\
15 & 9.4 \scriptsize{$\pm$ 3.7} & 1.2 \scriptsize{$\pm$ 0.6} & 4.4 \scriptsize{$\pm$ 1.3} & 1.2 \scriptsize{$\pm$ 1.1} & \textbf{0.2} \scriptsize{$\pm$ 0.2} & N/A (0) & 202.0 \scriptsize{$\pm$ 41.1} \textbf{\footnotesize{ (4)}} & \textbf{58.5} \scriptsize{$\pm$ 29.0} \textbf{\footnotesize{ (4)}} \\
20 & 16.0\scriptsize{$\pm$ 3.9} & \textbf{0.2}\scriptsize{$\pm$ 0.1} & 30.4 \scriptsize{$\pm$ 10.6}  & 4.4 \scriptsize{$\pm$ 2.5} & 1.4 \scriptsize{$\pm$ 0.6} & N/A (0) & 438.0 \scriptsize{$\pm$ 0.0} \footnotesize{ (1)} &  \textbf{195.5} \scriptsize{$\pm$ 127.6} \textbf{\footnotesize{ (2)}} \\
50 & 19.7 \scriptsize{$\pm$ 3.1} & \textbf{9.8} \scriptsize{$\pm$ 1.5} & 219.8 \scriptsize{$\pm$ 13.7}  & 133.0 \scriptsize{$\pm$ 6.8} & 11.0 \scriptsize{$\pm$ 2.6} & N/A (0) & N/A (0) &  N/A (0)\\
\bottomrule
\end{tabular}
}
\label{table:tsp_main_results}
\end{table}

We present the results in Table~\ref{table:tsp_main_results}. We randomly generate 5 problem instances for each number of nodes $n$. In addition to measuring the optimality gap, on problems where the LLM finds the optimal solutions, we also show the number of optimization steps taken to reach the global optimum. First, we observe that \texttt{gpt-4} significantly outperforms \texttt{gpt-3.5-turbo} and \texttt{text-bison} across all problem sizes. Specifically, on smaller-scale problems, \texttt{gpt-4} reaches the global optimum about $4 \times$ faster than other LLMs. On larger-scale problems, especially with $n=50$, \texttt{gpt-4} still finds solutions with a comparable quality to heuristic algorithms, while both \texttt{text-bison} and \texttt{gpt-3.5-turbo} get stuck at local optima with up to $20\times$ worse optimality gaps.

On the other hand, the performance of \name{} degrades dramatically on problems with larger sizes. When $n=10$, all LLMs find the optimal solutions for every evaluated problem; as the problem size gets larger, the \name{} optimality gaps increase quickly, and the farthest insertion heuristic starts to outperform all LLMs in the optimality gap.

\paragraph{Limitations.} We would like to note that \name{} is designed for neither outperforming the state-of-the-art gradient-based optimization algorithms for continuous mathematical optimization, nor surpassing the performance of specialized solvers for classical combinatorial optimization problems such as TSP. Instead, the goal is to demonstrate that LLMs are able to optimize different kinds of objective functions simply through prompting, and reach the global optimum for some small-scale problems.
Our evaluation reveals several limitations of \name{} for mathematical optimization.
Specifically, the length limit of the LLM context window makes it hard to fit large-scale optimization problem descriptions in the prompt, e.g., linear regression with high-dimensional data, and traveling salesman problems with a large set of nodes to visit. In addition, the optimization landscape of some objective functions are too bumpy for the LLM to propose a correct descending direction, causing the optimization to get stuck halfway. We further elaborate our observed failure cases in Appendix~\ref{appsec:some_failure_cases}.
\section{Application: Prompt Optimization}
\label{sec:application_prompt_opt}

\begin{figure}[t]
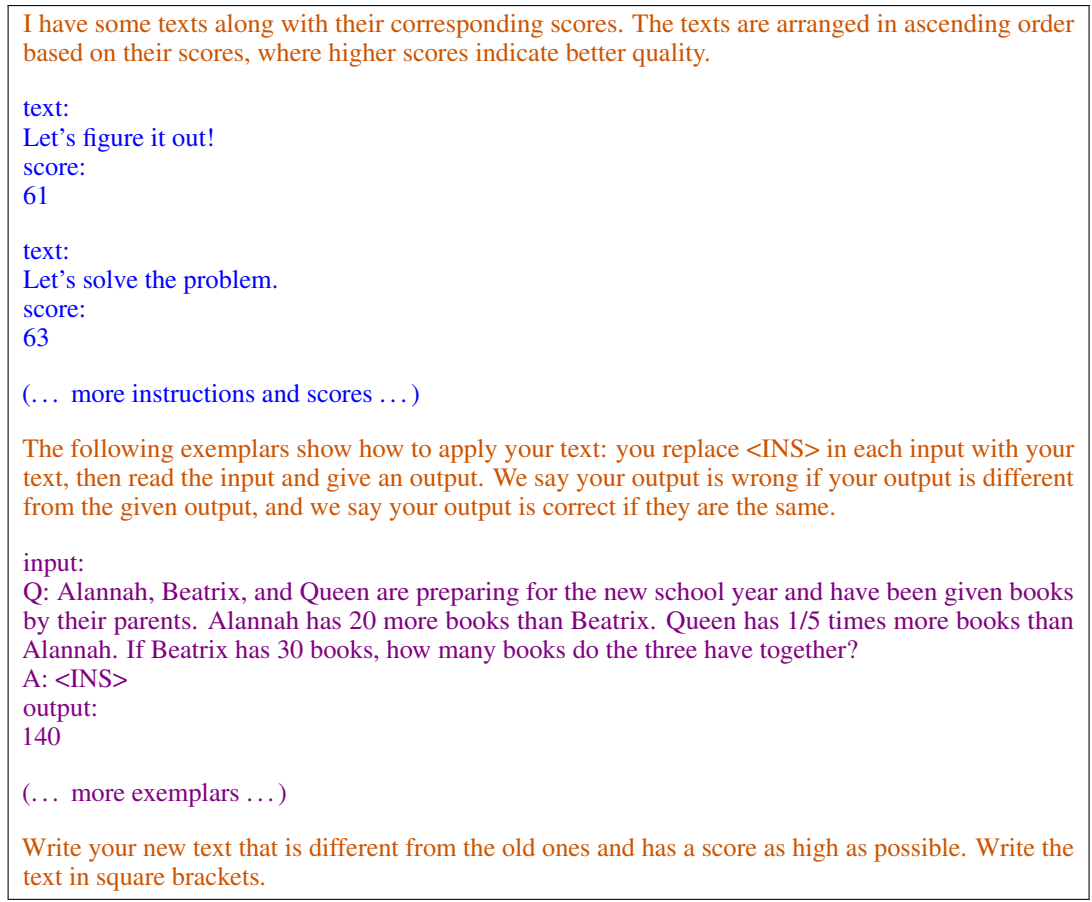

\noindent\fbox{
\parbox{\textwidth}{
\color{burntorange}{
I have some texts along with their corresponding scores. The texts are arranged in ascending order based on their scores, where higher scores indicate better quality.
}

\vspace{1em}
\color{blue}{
text:

Let's figure it out!

score:

61

\vspace{1em}

text:

Let's solve the problem.

score:

63

\vspace{1em}
(… more instructions and scores …)
\vspace{1em}
}

\color{burntorange}{
The following exemplars show how to apply your text: you replace <INS> in each input with your text, then read the input and give an output. We say your output is wrong if your output is different from the given output, and we say your output is correct if they are the same.
}
\color{violet}{
\vspace{1em}

input:

Q: Alannah, Beatrix, and Queen are preparing for the new school year and have been given books by their parents. Alannah has 20 more books than Beatrix. Queen has 1/5 times more books than Alannah. If Beatrix has 30 books, how many books do the three have together?

A: <INS>

output:

140

\vspace{1em}
(… more exemplars …)
\vspace{1em}
}

\color{burntorange}{
Write your new text that is different from the old ones and has a score as high as possible. Write the text in square brackets.}
}
}
\caption{An example of the meta-prompt for prompt optimization with instruction-tuned \texttt{PaLM 2-L} (\texttt{PaLM 2-L-IT}) on GSM8K, where the generated instruction will be prepended to the beginning of ``A:'' in the scorer LLM output (\emph{A\_begin} in Section~\ref{sec:setup}). <INS> denotes the position where the generated instruction will be added. The \textcolor{blue}{blue} text contains solution-score pairs; the \textcolor{violet}{purple} text describes the optimization task and output format; the \textcolor{burntorange}{orange} text are meta-instructions.}
\label{fig:meta_prompt_example}
\end{figure}

Next, we demonstrate the effectiveness of \name{} on prompt optimization, where the objective is to find the prompt that maximizes task accuracy. We first introduce the problem setup, then illustrate the meta-prompt design.

\subsection{Problem Setup}
\label{sec:setup}

We focus on prompt optimization for natural language tasks, where both the input and output are in the text format. 
The task is represented as a dataset with training and test splits, where the training set is used to calculate the training accuracy as the objective value during the optimization process, and we compute the test accuracy on the test set after the optimization finishes. 
While traditional optimization often requires a decently large training set, our experiment shows that a small number or fraction of training samples (e.g., 3.5\% of the training set for GSM8K~\citep{cobbe2021training}, 20\% for Big-Bench Hard~\citep{suzgun2022challenging}) is sufficient. 
The objective function evaluator is an LLM to which the optimized prompt will be applied, and it can be the same or different from the LLM for optimization. 
We denote the LLM for objective function evaluation as the \emph{scorer LLM}, and the LLM for optimization as the \emph{optimizer LLM}.

The output of the optimizer LLM is an \emph{instruction}, which is concatenated to the question part of every exemplar and prompts the scorer LLM. 
We consider the following positions to insert the instruction:

\begin{itemize}[leftmargin=2em,topsep=0pt,partopsep=1ex,parsep=0ex]
\item \emph{Q\_begin}: the instruction is added before the original question.
\item \emph{Q\_end}: the instruction is added after the original question.
\item \emph{A\_begin}: the instruction is added to the beginning of the scorer LLM output. This is applicable to pretrained LLMs without instruction tuning, where the prompt is formatted as a sequence of QA pairs.
\end{itemize}

We exemplify these prompting formats in Appendix~\ref{appsec:scorer_prompting_formats}.

\subsection{Meta-Prompt Design}

Figure~\ref{fig:meta_prompt_example} shows an example of the meta-prompt for prompt optimization on GSM8K~\citep{cobbe2021training}.
More details are as follows.

\myparagraph{Optimization problem examples} The problem description includes a few examples taken from the training set to demonstrate the task for the generated instructions. 
For example, from the input-output pair in Figure~\ref{fig:meta_prompt_example}, we can infer this is a math word problem. 
The input-output pair also demonstrates the position where the generated instruction will be added to, and this is essential for the optimizer LLM to generate instructions of the same style.
In each optimization step, we add several (three for example) training examples to the meta-prompt by random sampling the training set or choose the ones the previous instructions fall short of.

\myparagraph{Optimization trajectory} 
The optimization trajectory includes instructions generated from the past optimization steps, along with their scores.
The old instructions and scores are sorted by the score in ascending order.
The score is the training accuracy in prompt optimization. 
We only keep instructions with the highest scores in the meta-prompt in consideration of the LLM context length limit.

\myparagraph{Meta-instructions}
We also add \emph{meta-instructions}: the instructions to the optimizer LLM that explain the optimization goal and instruct the model how to use the above information.
The meta-instructions may also specify the desired generated instruction format for easier parsing.
\section{Prompt Optimization Experiments}
\label{sec:exp}

We present the evaluation results for prompt optimization in this section. Our experiments demonstrate that \name{} brings a significant performance gain across the board, with different combinations of LLMs as the optimizer and the scorer.

Section~\ref{sec:evaluation_setup} describes the experiment setup.
Section~\ref{sec:main_results} shows main results on reasoning tasks like GSM8K and BBH.
Section~\ref{sec:ablation} shows ablation studies. 
Section~\ref{sec:overfitting_analysis_in_prompt_optimization} analyzes overfitting in prompt optimization.
Section~\ref{sec:comparison_with_evoprompt} compares the prompt optimization performance of meta-prompts in \name{} and EvoPrompt~\citep{guo2023connecting}.

\subsection{Evaluation Setup}
\label{sec:evaluation_setup}

\myparagraph{Models}
The LLMs we use as the optimizer and the scorer are:

\begin{itemize}[leftmargin=2em,topsep=0pt,partopsep=1ex,parsep=0ex]
\item Optimizer LLM: Pre-trained \texttt{PaLM 2-L}~\citep{anil2023palm}, instruction-tuned \texttt{PaLM 2-L} (denoted \texttt{PaLM 2-L-IT}), \texttt{text-bison}, \texttt{gpt-3.5-turbo}, and \texttt{gpt-4}.
\item Scorer LLM: Pre-trained \texttt{PaLM 2-L} and \texttt{text-bison}.
\end{itemize}

With pre-trained \texttt{PaLM 2-L} as the scorer, the optimizer LLM generates A\_begin instructions.
Since \texttt{text-bison} has been instruction-tuned, the optimizer LLM generates Q\_begin and Q\_end instructions when \texttt{text-bison} is used as the scorer.

\myparagraph{Benchmarks}
Our primary evaluation benchmarks are GSM8K~\citep{cobbe2021training} and Big-Bench Hard (BBH)~\citep{suzgun2022challenging}. GSM8K is a benchmark of grade school math word problems with 7,473 training samples and 1,319 test samples, where chain-of-thought prompting~\citep{wei2022chain} and the zero-shot instruction ``Let's think step by step.''~\citep{kojima2022large} have drastically improved the performance over the standard prompting. BBH is a suite of 23 challenging BIG-Bench tasks~\citep{srivastava2022beyond} that covers a wide range of topics beyond arithmetic reasoning, including symbolic manipulation and commonsense reasoning. Each task contains up to 250 examples in total.

To examine the transferability of the optimized instructions, we also evaluate the instructions optimized for GSM8K on two other mathematical reasoning datasets, i.e.,  MultiArith~\citep{roy2016solving} and AQuA~\citep{ling2017program}.

\myparagraph{Implementation details}
We set the temperature to be 0 when evaluating the performance of generated instructions, in which case the scorer LLM greedily decodes.
Unless otherwise specified, we set the default temperature to be 1.0 for optimizer LLMs to generate diverse and creative instructions.
At each optimization step, we prompt the optimizer LLM with the meta-prompt 8 times to generate 8 instructions, then we add these instructions with their training scores to the optimization trajectory in the meta-prompt.
Our meta-prompt at each step contains the best 20 instructions so far and 3 randomly picked exemplars from the training set.
We study the effect of different hyperparameters in ablation studies (Section~\ref{sec:ablation}). Appendix~\ref{appsec:meta_prompts_for_prompt_opt} presents the full meta-prompts for different optimizer LLMs.

\subsection{Main Results}
\label{sec:main_results}
We show prompt optimization curves on GSM8K and two BBH tasks in this section.
The curves on other BBH tasks are deferred to Appendix~\ref{appsec:bbh_optimization_curves}, and the tables containing all accuracy numbers are in Appendix~\ref{appsec:bbh_taskwise_detailed_results}.

\subsubsection{GSM8K}

\begin{table}[t]
\footnotesize
\caption{Test accuracies on GSM8K. We show the instruction with the highest test accuracy for each scorer-optimizer pair. 
}
\begin{center}
\scalebox{0.86}{
\begin{tabular}{cP{2cm}P{1.5cm}P{8cm}c}
\toprule
Scorer & Optimizer / Source & Instruction position & Top instruction & Acc \\
\midrule
\multicolumn{3}{l}{\textit{Baselines}} \\
\hdashline\noalign{\vskip 0.5ex}
\texttt{PaLM 2-L} & \citep{kojima2022large} & A\_begin & Let's think step by step. & 71.8 \\ [1ex]
\texttt{PaLM 2-L} & \citep{zhou2022large} & A\_begin & Let’s work this out in a step by step way to be sure we have the right answer. & 58.8 \\ [3ex]
\texttt{PaLM 2-L} & & A\_begin & Let's solve the problem. & 60.8 \\ [1ex]
\texttt{PaLM 2-L} & & A\_begin & (empty string) & 34.0 \\ [1ex]
\texttt{text-bison} & \citep{kojima2022large} & Q\_begin & Let's think step by step. & 64.4 \\ [1ex]
\texttt{text-bison} & \citep{zhou2022large} & Q\_begin & Let’s work this out in a step by step way to be sure we have the right answer. & 65.6 \\ [3ex]
\texttt{text-bison} & & Q\_begin & Let's solve the problem. & 59.1 \\ [1ex]
\texttt{text-bison} & & Q\_begin & (empty string) & 56.8 \\
\midrule
\multicolumn{3}{l}{\textit{Ours}} \\
\hdashline\noalign{\vskip 0.5ex}
\texttt{PaLM 2-L} & \texttt{PaLM 2-L-IT} & A\_begin & Take a deep breath and work on this problem step-by-step. & \textbf{80.2} \\ [1ex]
\texttt{PaLM 2-L} & \texttt{PaLM 2-L} & A\_begin & Break this down. & 79.9 \\ [1ex]
\texttt{PaLM 2-L} & \texttt{gpt-3.5-turbo} & A\_begin & A little bit of arithmetic and a logical approach will help us quickly arrive at the solution to this problem. & 78.5 \\ [3ex]
\texttt{PaLM 2-L} & \texttt{gpt-4} & A\_begin & Let's combine our numerical command and clear thinking to quickly and accurately decipher the answer. & 74.5 \\ [3ex]
\texttt{text-bison} & \texttt{PaLM 2-L-IT} & Q\_begin & Let's work together to solve math word problems! First, we will read and discuss the problem together to make sure we understand it. Then, we will work together to find the solution. I will give you hints and help you work through the problem if you get stuck. & 64.4 \\ [3ex]
\texttt{text-bison} & \texttt{text-bison} & Q\_end & Let's work through this problem step-by-step: & \textbf{68.5} \\ [1ex]
\texttt{text-bison} & \texttt{gpt-3.5-turbo} & Q\_end & Analyze the given information, break down the problem into manageable steps, apply suitable mathematical operations, and provide a clear, accurate, and concise solution, ensuring precise rounding if necessary. Consider all variables and carefully consider the problem's context for an efficient solution. & 66.5 \\ [5ex]
\texttt{text-bison} & \texttt{gpt-4} & Q\_begin & Start by dissecting the problem to highlight important numbers and their relations. Decide on the necessary mathematical operations like addition, subtraction, multiplication, or division, required for resolution. Implement these operations, keeping in mind any units or conditions. Round off by ensuring your solution fits the context of the problem to ensure accuracy. & 62.7 \\
\bottomrule
\end{tabular}}
\end{center}
\label{table:top_instructions_on_gsm8k}
\end{table}

For prompt optimization, we randomly sample 3.5\% examples from the GSM8K training set.
The same subset is used throughout optimization, so that the task accuracies computed at intermediate optimization steps are approximations of the training accuracy on all 7,473 training examples.
This balances the evaluation cost with the generalization performance.
After the optimization procedure finishes, we evaluate the found instructions on the entire GSM8K test set.

Figure~\ref{fig:prompt_optimization_graph_in_intro_gsm8k} in Section~\ref{sec:intro} shows prompt optimization curves with pre-trained \texttt{PaLM 2-L} as scorer and \texttt{PaLM 2-L-IT} as optimizer, and the initial instruction is ``Let's solve the problem'' with a (approximated, and same below) training accuracy of 60.5.
We observe that the optimization curve shows an overall upward trend with several leaps throughout the optimization process, for example:
\begin{itemize}[leftmargin=2em,topsep=0pt,partopsep=1ex,parsep=0ex]
\item ``Let's think carefully about the problem and solve it together.'' at Step 2 with the training accuracy 63.2;
\item ``Let's break it down!'' at Step 4 with training accuracy 71.3;
\item ``Let's calculate our way to the solution!'' at Step 5 with training accuracy 73.9;
\item ``Let's do the math!'' at Step 6 with training accuracy 78.2.
\end{itemize}

The optimization curves also generally show a decrease of the variance among the accuracies of instructions generated at each step, indicating that the optimizer LLM generates \emph{distributionally} better instructions throughout the optimization.

\begin{figure}
\centering
\subfigure[\texttt{PaLM 2-L-IT} optimizer]{\label{fig:prompt_optimization_graph_gsm8k_text_bison}\includegraphics[width=.35\linewidth]{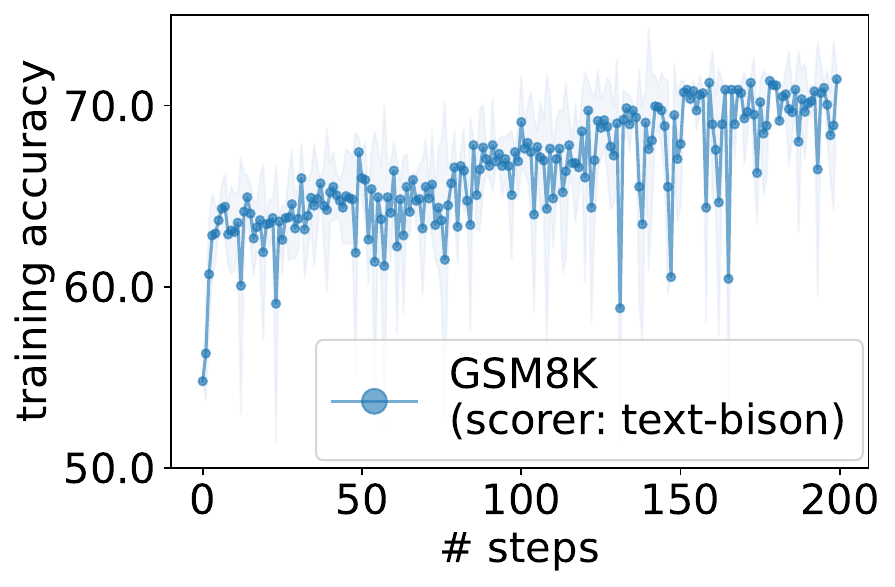}}
\hspace{.01\linewidth}
\subfigure[pre-trained \texttt{PaLM 2-L} optimizer]{\label{fig:prompt_optimization_graph_gsm8k_all_pretrained}\includegraphics[width=.35\linewidth]{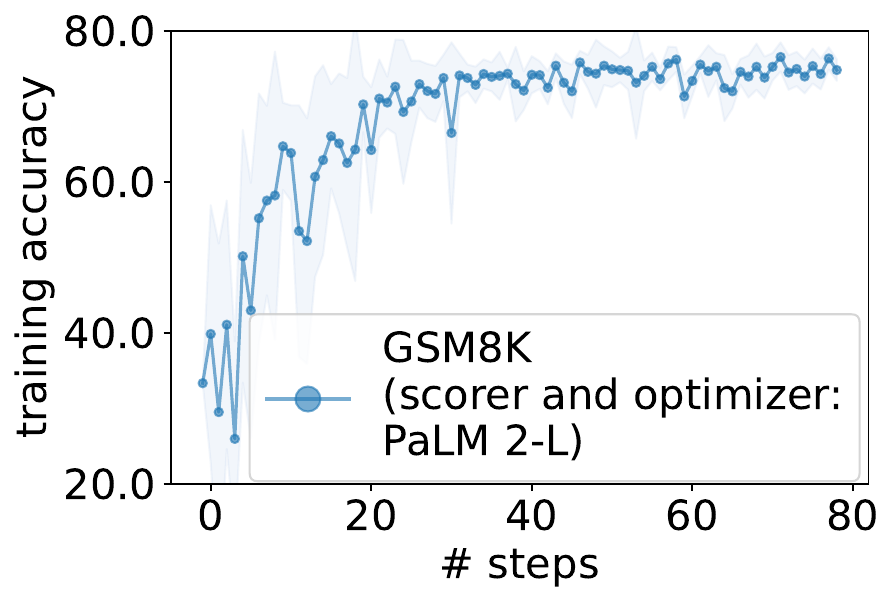}}

\caption{Prompt optimization on GSM8K with \subref{fig:prompt_optimization_graph_gsm8k_text_bison} the \texttt{text-bison} scorer and the \texttt{PaLM 2-L-IT} optimizer, and \subref{fig:prompt_optimization_graph_gsm8k_all_pretrained} pre-trained \texttt{PaLM 2-L} as both scorer and optimizer.
}
\label{fig:prompt_optimization_in_main_results_gsm8k_more}
\end{figure}

Next, we present the results of generating Q\_begin instructions with the \texttt{text-bison} scorer and the \texttt{PaLM 2-L-IT} optimizer, starting from an empty instruction with a 57.1 training accuracy.
The optimization curve in Figure~\ref{fig:prompt_optimization_graph_gsm8k_text_bison} shows a similar upward trend, during which a few leaps in the training accuracy include:
\begin{itemize}[leftmargin=2em,topsep=0pt,partopsep=1ex,parsep=0ex]
\item ``Solve the following problems using the given information.'' at Step 2 with training accuracy 59.8;
\item ``Solve the following problems by applying the given information and using the appropriate mathematical operations.'' at Step 3 with training accuracy 64.0;
\item ``Let's read the problem carefully and identify the given information. Then, we can create an equation and solve for the unknown variable.'' at Step 4 with training accuracy 67.0;
\item ``I'm always down for solving a math word problem together. Just give me a moment to read and understand the problem. Then, I'll create an equation that models the problem, which I'll solve for the unknown variable. I also may or may not use some helpful diagrams or visuals to understand the problem. Lastly, be sure to allow me some time to carefully check my work before submitting any responses!'' at Step 29 with training accuracy 70.1.
\end{itemize}

Note that although our default setting is to run \name{} for 200 steps in prompt optimization, we need much fewer steps if the goal is to find some outstanding instructions.
An example is that the Figure~\ref{fig:prompt_optimization_graph_in_intro_gsm8k} experiment found ``Let's do the math!'' at Step 6 with training accuracy 78.2, almost matching the ``Take a deep breath and work on this problem step-by-step.'' found at the 107th step with training accuracy 80.2, at a point where the optimization curve is still trending upwards.
This is because a leap in our optimization curve does not always correspond to a much better instruction being discovered; instead, it can be due to a large qualitative improvement of all 8 generated instructions in this step.
The latter usually happens several steps after the former: after a much better instruction is discovered in one step, the meta-prompt gradually gets rid of worse instructions in the latter steps by generating instructions similar to the much-better one. 
The top instructions kept in the meta-prompt gradually improves in this procedure.
At a point when the meta-prompt only triggers higher quality instructions, the leap happens.

Finally, Figure~\ref{fig:prompt_optimization_graph_gsm8k_all_pretrained} shows that the pre-trained \texttt{PaLM 2-L} can also serve as the optimizer LLM and improve its own prediction performance.
Different from other optimizer LLMs that are instruction-tuned, the pre-trained \texttt{PaLM 2-L} performs better when the prompt is formatted in a few-shot manner. Therefore, we include two initial instructions to start the optimization: the empty instruction (with a training accuracy 32.2) and ``The answer is'' (with a training accuracy 33.3).
See Figure~\ref{fig:meta_prompt_example_pretrained_palm_2_l} in Appendix~\ref{appsec:meta_prompts} for the meta-prompt format.
The generated instructions follow the same style as ``The answer is'': most instructions are also phrases suitable as the prefix of a sentence, like ``Here you go:'' (generated at Step 11 with training accuracy 61.3) and ``Let's do it:'' (generated at Step 13 with training accuracy 75.1).

Table~\ref{table:top_instructions_on_gsm8k} summarizes top instructions found on GSM8K with different scorer and optimizer LLMs.
We observe that:
\begin{itemize}[leftmargin=2em,topsep=0pt,partopsep=1ex,parsep=0ex]
\item The styles of instructions found by different optimizer LLMs vary a lot: \texttt{PaLM 2-L-IT} and \texttt{text-bison} ones are concise, while GPT ones are long and detailed.
\item Although some top instructions contain the ``step-by-step'' phrase, most others achieve a comparable or better accuracy with different semantic meanings.
\end{itemize}

\begin{figure}[t]
\centering
\subfigure[\scriptsize \texttt{PaLM 2-L} scorer, ours minus ``Let's think step by step.'']{\label{fig:accuracy_comparison_palm_2_l_found_minus_step_by_step}\includegraphics[width=.48\linewidth]{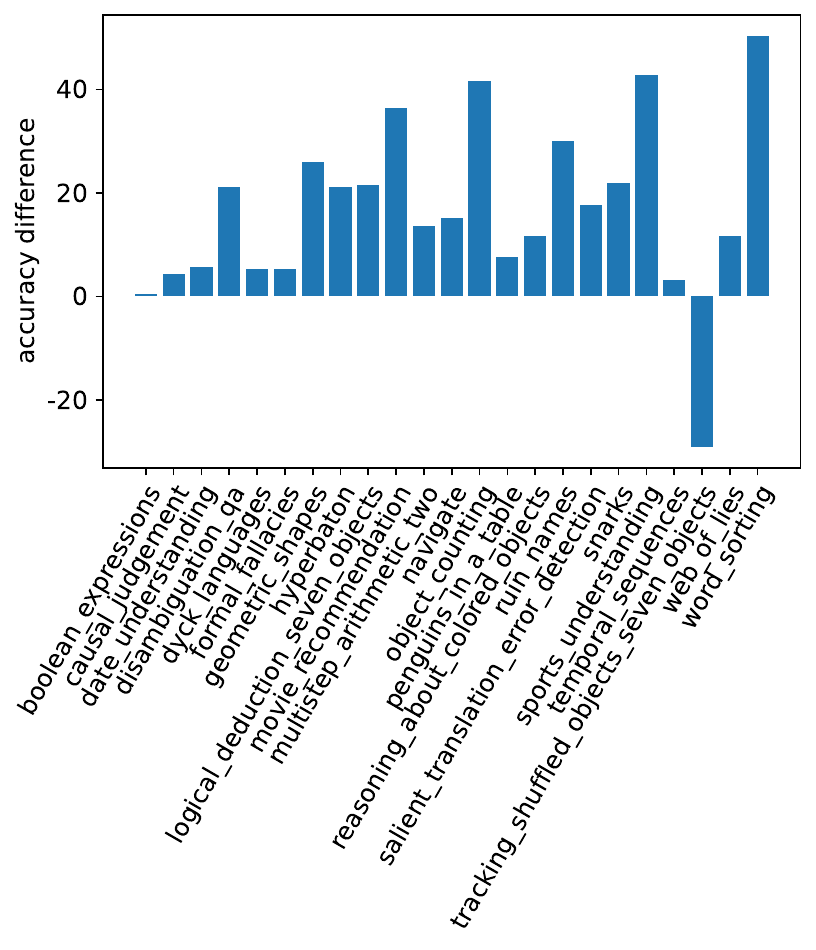}}
\hspace{.01\linewidth}
\subfigure[\scriptsize \texttt{PaLM 2-L} scorer, ours minus empty starting point]{\label{fig:accuracy_comparison_palm_2_l_found_minus_empty}\includegraphics[width=.48\linewidth]{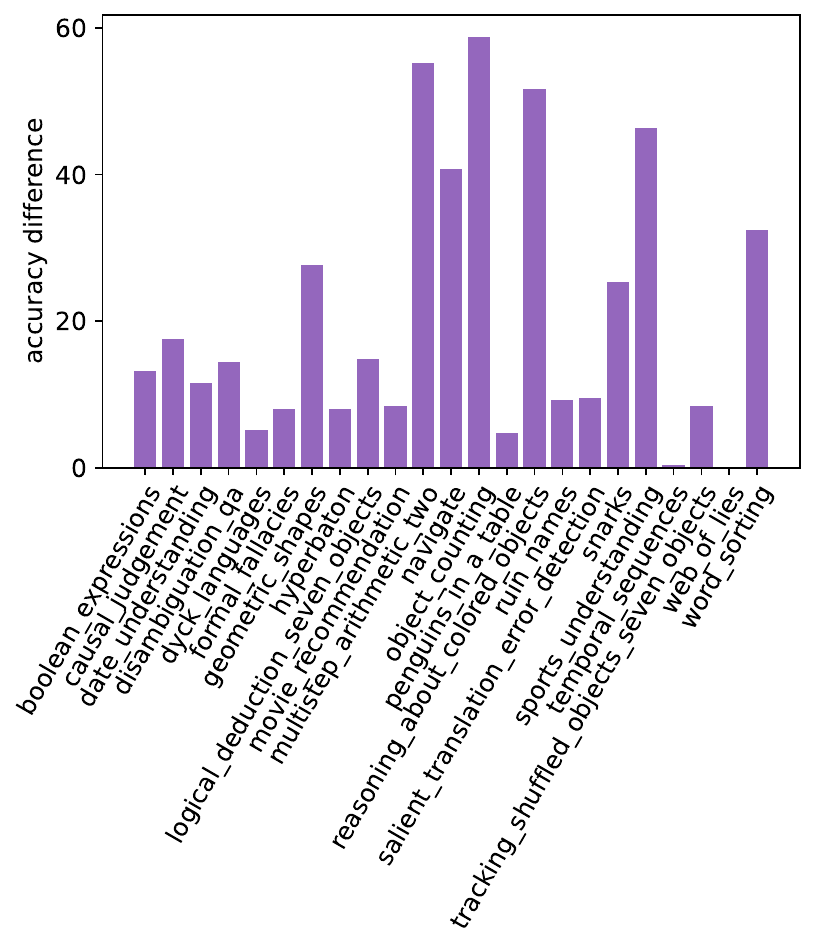}}

\subfigure[\scriptsize \texttt{text-bison} scorer, ours minus ``Let's think step by step.'']{\label{fig:accuracy_comparison_text_bison_found_minus_step_by_step}\includegraphics[width=.48\linewidth]{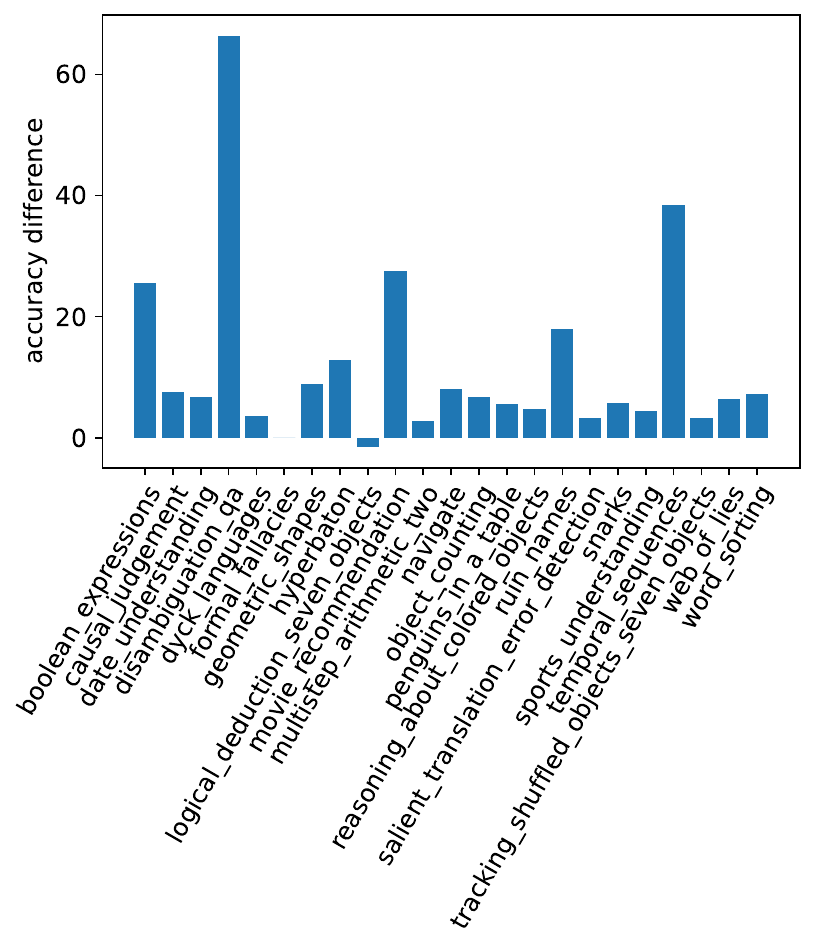}}
\hspace{.01\linewidth}
\subfigure[\scriptsize \texttt{text-bison} scorer, ours minus empty starting point]{\label{fig:accuracy_comparison_text_bison_found_minus_empty}\includegraphics[width=.48\linewidth]{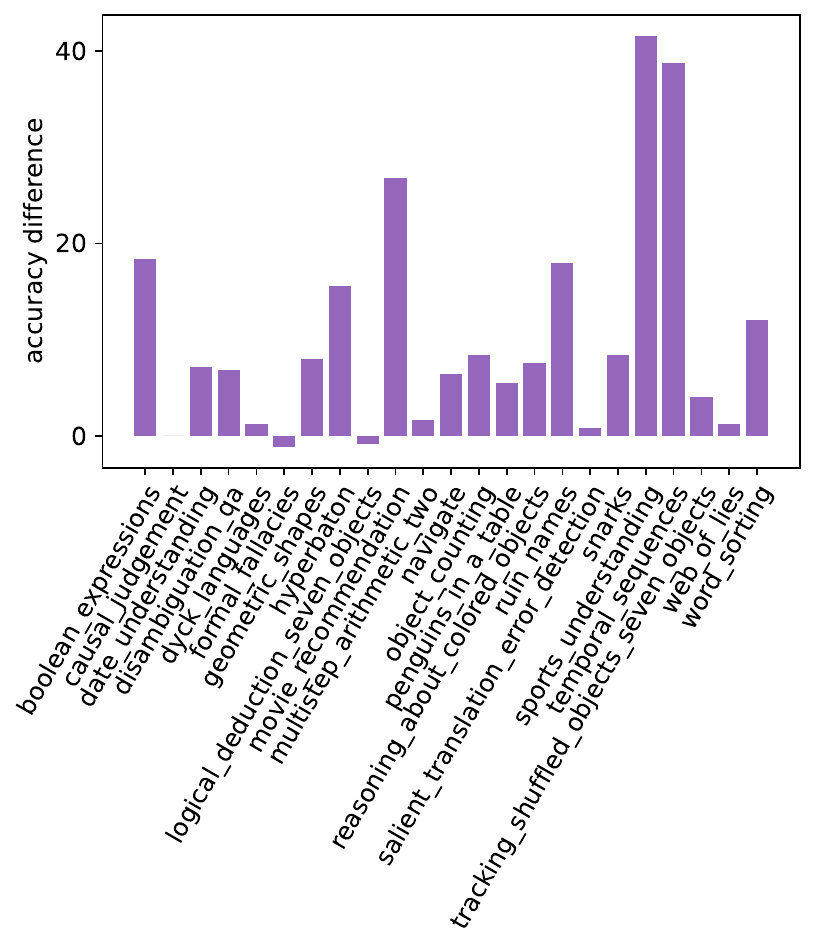}}

\caption{On 23 BBH tasks, the accuracy differences among instructions found by prompt optimization (with the \texttt{PaLM 2-L-IT} optimizer), ``Let's think step by step.'', and the empty string (optimization starting point).
}
\label{fig:accuracy_comparison_bar_charts_o_palm_2_l_it}
\end{figure}

\begin{figure}[t]
\centering
\subfigure[BBH ruin\_names]{\label{fig:prompt_optimization_graph_bbh_ruin_names}\includegraphics[width=.35\linewidth]{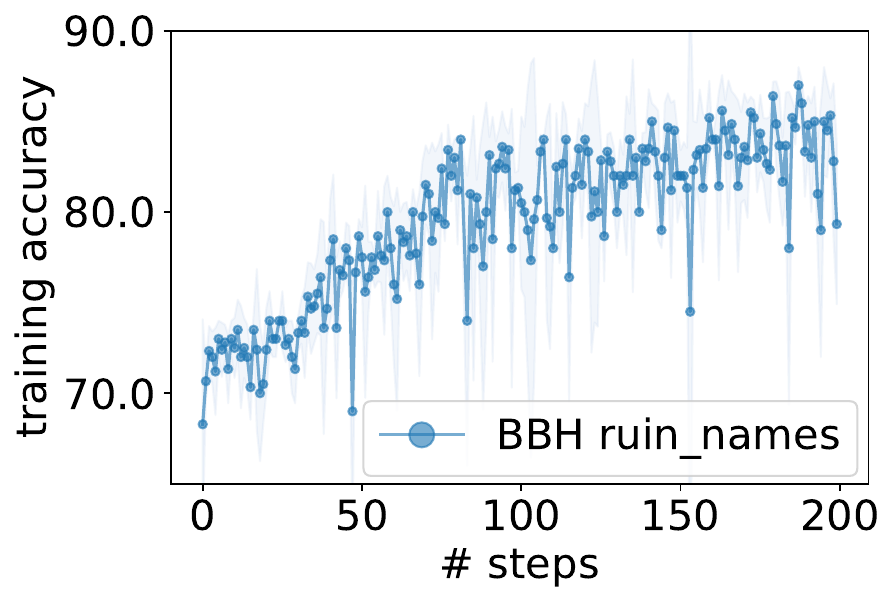}}
\hspace{.01\linewidth}
\subfigure[BBH temporal\_sequences]{\label{fig:prompt_optimization_graph_bbh_temporal_sequences}\includegraphics[width=.35\linewidth]{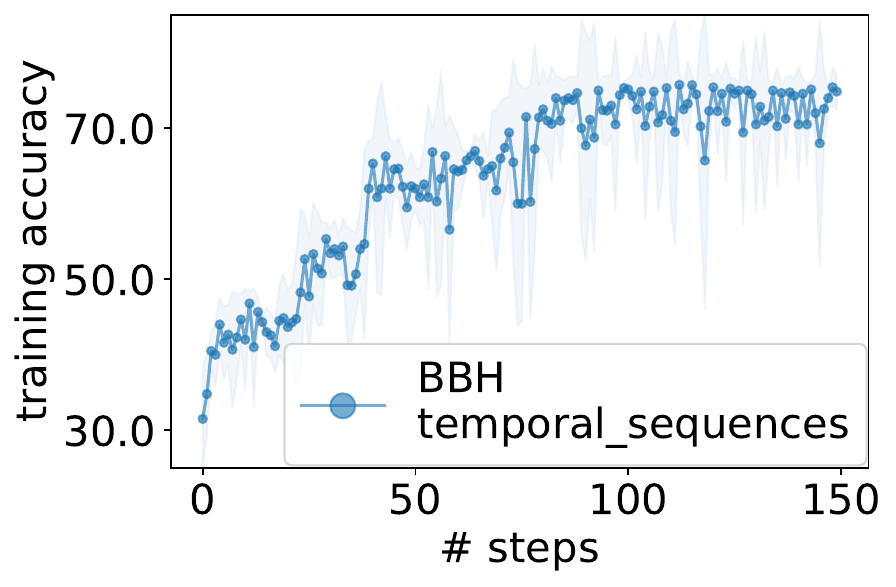}}

\caption{Training accuracy curves of prompt optimization on BBH ruin\_names and temporal\_sequences with the \texttt{text-bison} scorer and the \texttt{PaLM 2-L-IT} optimizer.
The optimizations start from the empty string.
}
\label{fig:prompt_optimization_in_main_results_bbh}
\end{figure}

\subsubsection{BBH}
On BBH, the optimization starts from an empty string as the initial instruction by default. The instructions are placed at A\_begin when the scorer is \texttt{PaLM 2-L}, and at Q\_begin when the scorer is \texttt{text-bison}. For each task, we utilize a subset of 20\% examples for prompt optimization, and the rest examples are for testing. We show experimental results on more variants of the instruction position and initialization in Appendix~\ref{appsec:bbh_taskwise_detailed_results}.

Figure~\ref{fig:accuracy_comparison_bar_charts_o_palm_2_l_it} visualizes the per-task accuracy difference on all 23 BBH tasks compared to the instruction ``Let's think step by step.''~\citep{kojima2022large} and the empty instruction, and we present the concrete accuracies in Table~\ref{table:palm2_scores_on_bbh_tasks} of Appendix~\ref{appsec:bbh_taskwise_detailed_results}. We show that the instructions found by \name{} outperform ``Let's think step by step.'' on almost all tasks by a large margin: our instructions outperform by over 5\% on 19/23 tasks with the \texttt{PaLM 2-L} scorer, and on 15/23 tasks with the \texttt{text-bison} scorer.
Our prompt optimization algorithm also improves instructions from the empty starting point by over 5\% on most tasks: 20/23 with the \texttt{PaLM 2-L} scorer and 15/23 with the \texttt{text-bison} scorer.

Similar to GSM8K, we observe upward trends in optimization curves on almost all BBH tasks, as shown in Figure~\ref{fig:prompt_optimization_in_main_results_bbh}.
See Figure~\ref{fig:prompt_optimization_curves_bbh_text_bison_scorer_all_tasks_appendix_part_one} and~\ref{fig:prompt_optimization_curves_bbh_text_bison_scorer_all_tasks_appendix_part_two} in Appendix~\ref{appsec:bbh_optimization_curves} for more curves on other BBH tasks. 

We next show some examples of instructions found through the course of optimization.
On the task ruin\_names, starting from the empty instruction (with 64.0 training accuracy), with the \texttt{text-bison} scorer and the \texttt{PaLM 2-L-IT} optimizer, the following instructions are generated:
\begin{itemize}[leftmargin=2em,topsep=0pt,partopsep=1ex,parsep=0ex]
\item ``Consider the following when editing artist or movie names humorously:'' at Step 1 with training accuracy 72.0;
\item ``When making humorous edits of artist or movie names, you can change one or more letters or even create puns by adding new words that sound similar.'' at Step 18 with training accuracy 80.0;
\item ``We can make humorous edits of artist/movie names by changing letters to create new words that are similar in sound but have different meanings. For example, The Police can be changed to The Polite, The Abyss can be changed to Toe Abyss, and Schindler’s List can be changed to Schindler’s Lost.'' at Step 38 with training accuracy 82.0.
\end{itemize}
Although the above instructions are semantically similar, a paraphrase by the optimizer LLM offers a notable accuracy improvement. We further highlight this observation in Section~\ref{sec:ins-acc-variance}.

Below are some instructions generated when performing prompt optimization on temporal\_sequences, starting from the empty instruction (with the training accuracy of 64.0):
\begin{itemize}[leftmargin=2em,topsep=0pt,partopsep=1ex,parsep=0ex]
\item ``To solve this problem, we need to first identify the time period when the person was not seen doing anything else. Then, we need to check if the place they went to was open during that time period. If it was, then that is the time period when they could have gone to that place.'' at Step 2 with training accuracy 42.0;
\item ``To find the time period when a person could have gone to a place, identify the time periods when they were not seen doing anything else and the place was open. If there are multiple time periods that match these criteria, then the person could have gone to the place during any of these time periods.'' at Step 18 with training accuracy 54.0;
\item ``To determine the possible time period when a person went to a place, first identify all the time periods when the person was not seen doing anything else and the place was open. Then, rule out any time periods during which the person was seen doing something else. The remaining time periods are the possible times when the person could have gone to the place.'' at Step 41 with training accuracy 72.0.
\end{itemize}

Table~\ref{table:top_instructions_on_bbh_tasks_main_paper} presents the best instructions generated on movie\_recommendation, ruin\_names, and temporal\_sequences tasks with different combinations of the optimizer and the scorer LLMs.
Again, different optimizer LLMs produce instructions of different styles. See Appendix~\ref{appsec:bbh_taskwise_detailed_results} for results on more BBH tasks.

\begin{table}
\centering
\caption{Top instructions with the highest accuracies found in prompt optimization on BBH movie\_recommendation, ruin\_names, and temporal\_sequences.}
\scalebox{0.85}{
\begin{tabular}{P{2.0cm}P{2.5cm}P{1.2cm}P{7.5cm}c}
\toprule
Scorer & Optimizer & Instruction position & Instruction & Acc \\
\midrule
\multicolumn{3}{l}{\textit{movie\_recommendation}} \\
\hdashline\noalign{\vskip 0.5ex}
\texttt{PaLM 2-L} & \texttt{PaLM 2-L-IT} & A\_begin & Based on your input, I have analyzed the given movies in terms of genre, plot, tone, audience rating, year of release, director, cast, and reviews. I have also taken into account the given options. The movie that is most similar to the given movies in terms of all these factors is: & 90.8 \\ [13ex]
\texttt{PaLM 2-L} & \texttt{PaLM 2-L} & A\_begin & The best film: & 88.4 \\ [1ex]
\texttt{PaLM 2-L} & \texttt{gpt-3.5-turbo} & A\_begin & Let's uncover the perfect movie recommendation from the options provided, ensuring an exceptional cinematic experience together as we select the most captivating and satisfying choice that will keep us thoroughly engaged and immersed until the very end. & 88.0 \\ [12ex]
\texttt{text-bison} & \texttt{PaLM 2-L-IT} & Q\_begin & What is the highest-rated movie similar to the given movies, with a similar IMDb rating and released in the same year? & 91.6 \\ [7ex]
\texttt{text-bison} & \texttt{gpt-3.5-turbo} & Q\_begin & Based on the movie list provided, carefully consider your preferences and make a well-informed decision. & 70.8 \\
\midrule
\multicolumn{3}{l}{\textit{ruin\_names}} \\
\hdashline\noalign{\vskip 0.5ex}
\texttt{PaLM 2-L} & \texttt{PaLM 2-L-IT} & A\_begin & Which is the funniest pun on the artist or movie name? & 88.0 \\ [1ex]
\texttt{PaLM 2-L} & \texttt{PaLM 2-L} & A\_begin & Answer for ruin: & 83.6 \\ [1ex]
\texttt{PaLM 2-L} & \texttt{gpt-3.5-turbo} & A\_begin & Prepare to have a side-splittingly funny time as we uncover the most clever and hilarious alternatives for these artist or movie names, challenging your wit to guess the correct one with a burst of creativity, humor, and imaginative twists! & 86.8 \\ [12ex]
\texttt{text-bison} & \texttt{PaLM 2-L-IT} & Q\_begin & A humorous edit of an artist or movie name can be created by replacing one or more letters to form a new word or phrase that sounds similar but has a different meaning. The new word or phrase should be relevant to the original word, but it should also be a surprise, which makes the edit funny. For example, the artist or movie name "Rocky" can be changed to "Ricky," and "Schindler's List" can be changed to "Schindler's Lift." Be creative and have fun! & 83.6 \\ [22ex]
\texttt{text-bison} & \texttt{gpt-3.5-turbo} & Q\_begin & Choose the option that offers the most clever and humorous alteration of the given artist or movie name. Let your creativity shine and select the answer that will undoubtedly bring a smile to your face! Make sure to think outside the box! & 75.2 \\
\midrule
\multicolumn{5}{l}{\textit{temporal\_sequences} (no \texttt{PaLM 2-L} as scorer results because its training accuracy on empty string is 100.0)} \\
\hdashline\noalign{\vskip 0.5ex}
\texttt{text-bison} & \texttt{PaLM 2-L-IT} & Q\_begin & To determine the time period when a person went to a place, first identify all the time periods when the person's whereabouts are unknown. Then, rule out any time periods during which the person was seen doing something else or the place was closed. The remaining time periods are the possible times when the person could have gone to the place. & 80.4 \\ [18ex]
\texttt{text-bison} & \texttt{gpt-3.5-turbo} & Q\_begin & Identify the optimal time slot for the individual to engage in the mentioned location/activity considering the given sightings and waking up time, taking into account the opening and closing times of the location and the duration of each event. & 53.6 \\
\bottomrule
\end{tabular}
}
\label{table:top_instructions_on_bbh_tasks_main_paper}
\end{table}

\subsubsection{Semantically similar instructions may achieve drastically different accuracies}
\label{sec:ins-acc-variance}
One challenge of prompt optimization is the sensitivity of model performance to subtle changes in the instruction. 
For example, with the \texttt{PaLM 2-L} scorer on the GSM8K test set, ``Let's think step by step.'' achieves accuracy 71.8, ``Let's solve the problem together.'' has accuracy 60.5, while the accuracy of ``Let's work together to solve this problem step by step.'' is only 49.4, although it is the semantic combination of the two upper instructions.
This behavior increases both the variance across single-step instructions and the oscillation during optimization, and motivates us to generate multiple instructions at each step to improve the optimization stability.

\subsubsection{Transferability of found instructions}

\begin{table}
\footnotesize
\caption{Transferability across datasets: accuracies of top instructions found for GSM8K on MultiArith and AQuA.
}
\begin{center}
\scalebox{0.9}{
\begin{tabular}{cP{2.2cm}P{1.5cm}P{5cm}cc}
\toprule
\multirow{2}{*}{Scorer} & \multirow{2}{*}{Source} & \multirow{2}{*}{\parbox{1.5cm} {\centering Instruction position}} & \multirow{2}{*}{Instruction} & \multicolumn{2}{c}{Accuracy} \\ \cmidrule{5-6}
& & & & MultiArith & AQuA \\
\midrule
\multicolumn{3}{l}{\textit{Baselines}} \\
\hdashline\noalign{\vskip 0.5ex}
\texttt{PaLM 2-L} & \citep{kojima2022large} & A\_begin & Let's think step by step. & 85.7 & 44.9 \\ [1ex]
\texttt{PaLM 2-L} & \citep{zhou2022large} & A\_begin & Let’s work this out in a step by step way to be sure we have the right answer. & 72.8 & 48.4 \\ [3ex]
\texttt{PaLM 2-L} & & A\_begin & Let's solve the problem. & 87.5 & 44.1 \\ [1ex]
\texttt{PaLM 2-L} & & A\_begin & (empty string) & 69.3 & 37.8 \\ [1ex]
\texttt{text-bison} & \citep{kojima2022large} & Q\_begin & Let's think step by step. & 92.5 & 31.9 \\ [1ex]
\texttt{text-bison} & \citep{zhou2022large} & Q\_begin & Let’s work this out in a step by step way to be sure we have the right answer. & 93.7 & 32.3 \\ [3ex]
\texttt{text-bison} & & Q\_begin & Let's solve the problem. & 85.5 & 29.9 \\ [1ex]
\texttt{text-bison} & & Q\_begin & (empty string) & 82.2 & 33.5 \\
\midrule
\multicolumn{3}{l}{\textit{Ours}} \\
\hdashline\noalign{\vskip 0.5ex}
\texttt{PaLM 2-L} & \texttt{PaLM 2-L-IT} on GSM8K & A\_begin & Take a deep breath and work on this problem step-by-step. & \textbf{95.3} & \textbf{54.3} \\ [4ex]
\texttt{text-bison} & \texttt{PaLM 2-L-IT} on GSM8K & Q\_begin & Let's work together to solve math word problems! First, we will read and discuss the problem together to make sure we understand it. Then, we will work together to find the solution. I will give you hints and help you work through the problem if you get stuck. & \textbf{96.8} & \textbf{37.8} \\
\bottomrule
\end{tabular}
}
\end{center}
\label{table:ins_performance_on_multiarith}
\end{table}

We assess the transferability of found prompts to different datasets of the same domain, where we evaluate the top instructions found for GSM8K on two more math reasoning benchmarks MultiArith~\citep{roy2016solving} and AQuA~\citep{ling2017program}.
Table~\ref{table:ins_performance_on_multiarith} shows that our optimized prompts also outperform baseline prompts with different scorer LLMs on these two benchmarks.

\subsection{Ablation Studies}
\label{sec:ablation}
We use \texttt{text-bison} as the scorer and \texttt{PaLM 2-L} as the optimizer for all ablation studies.
The tasks we evaluate are GSM8K (math reasoning) and BBH sports\_understanding (non-math reasoning).

\begin{figure}
\centering
\subfigure[instruction ordering (GSM8K)]{\label{fig:ablation_meta_prompt_ins_ordering_gsm8k}\includegraphics[width=.46\linewidth]{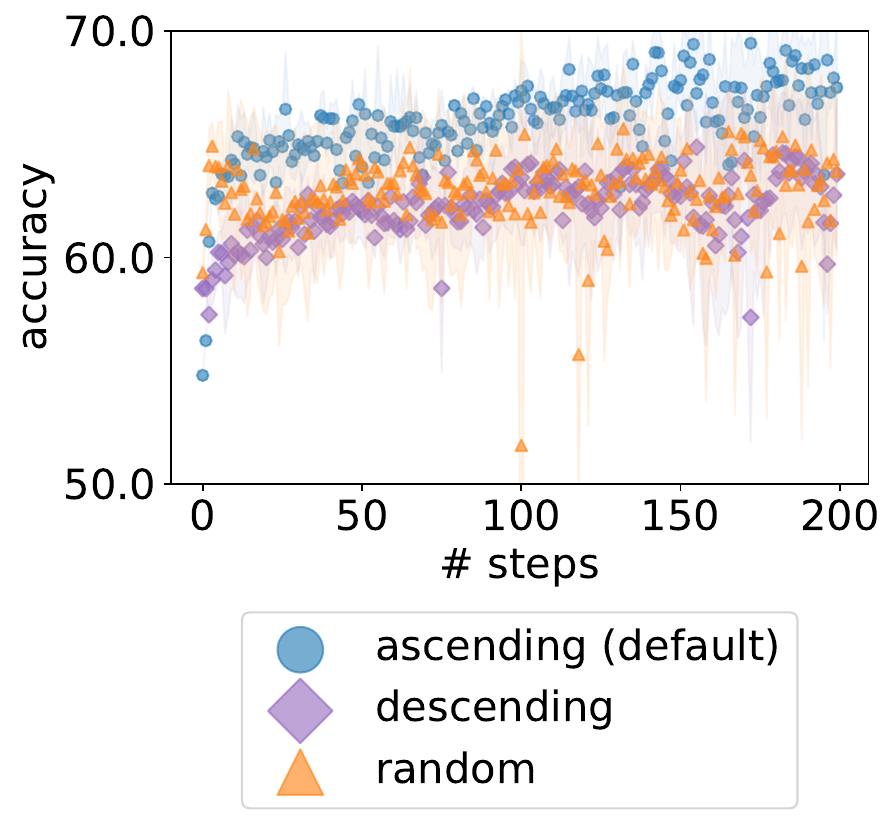}}
\hspace{.01\linewidth}
\subfigure[instruction ordering (BBH sports\_understanding)]{\label{fig:ablation_meta_prompt_ins_ordering_sports}\includegraphics[width=.47\linewidth]{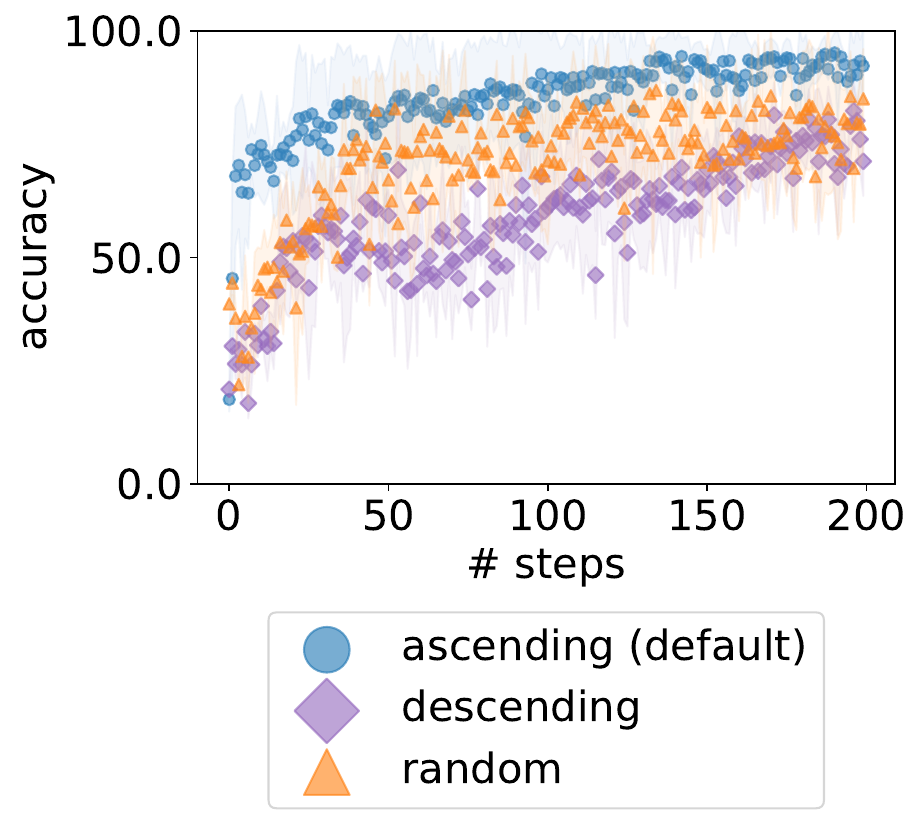}}

\subfigure[instruction scores (GSM8K)]{\label{fig:ablation_meta_prompt_ins_scores_gsm8k}\includegraphics[width=.46\linewidth]{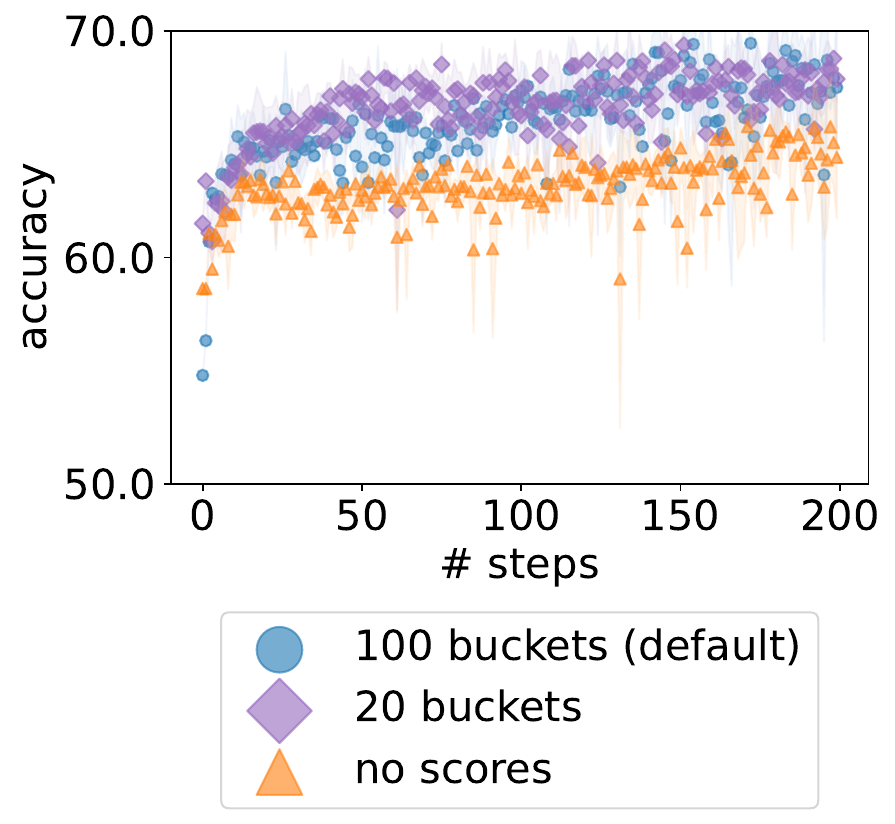}}
\hspace{.01\linewidth}
\subfigure[instruction scores (BBH sports\_understanding)]{\label{fig:ablation_meta_prompt_ins_scores_sports}\includegraphics[width=.46\linewidth]{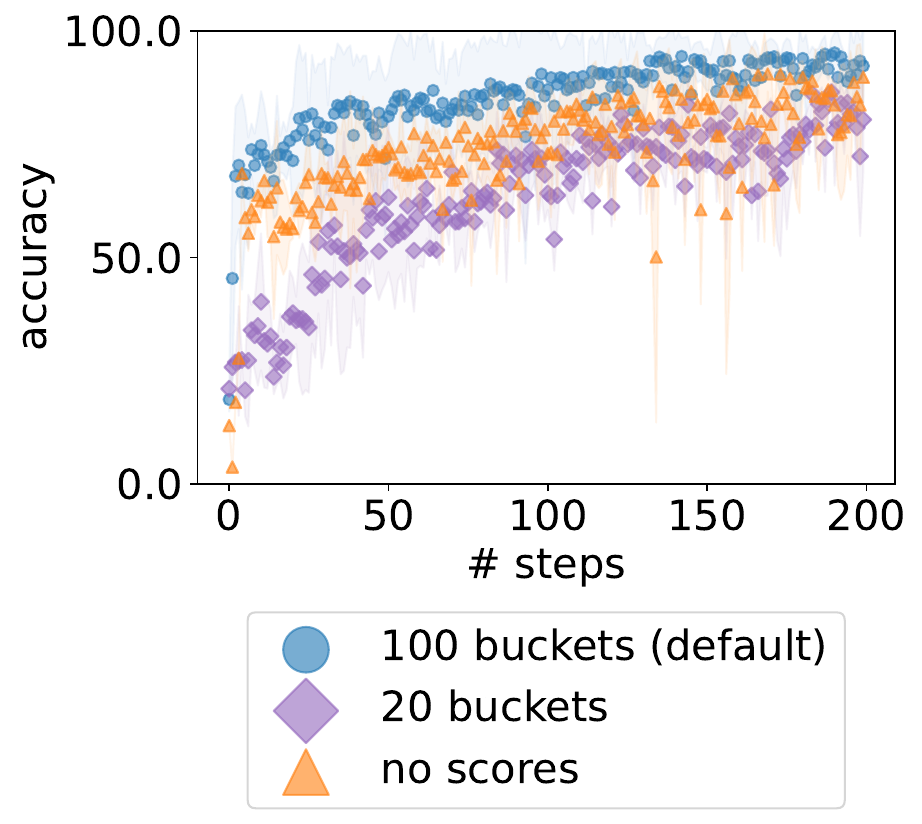}}

\subfigure[\# exemplars (GSM8K)]{\label{fig:ablation_meta_prompt_exemplars_gsm8k}\includegraphics[width=.46\linewidth]{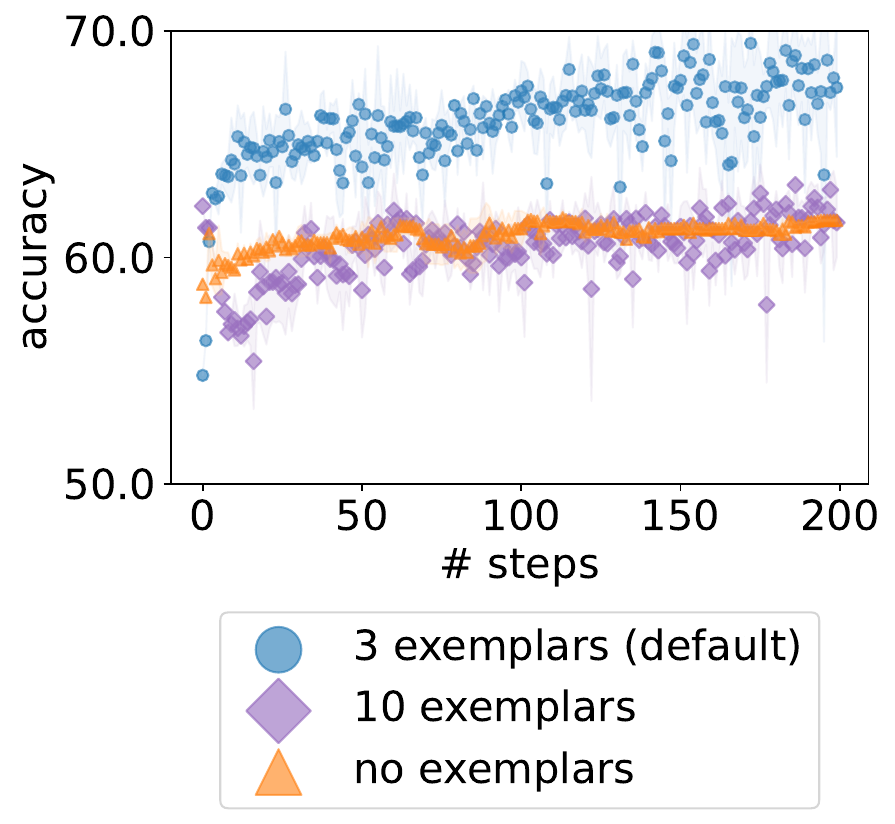}}
\hspace{.01\linewidth}
\subfigure[\# exemplars (BBH sports\_understanding)]{\label{fig:ablation_meta_prompt_exemplars_sports}\includegraphics[width=.46\linewidth]{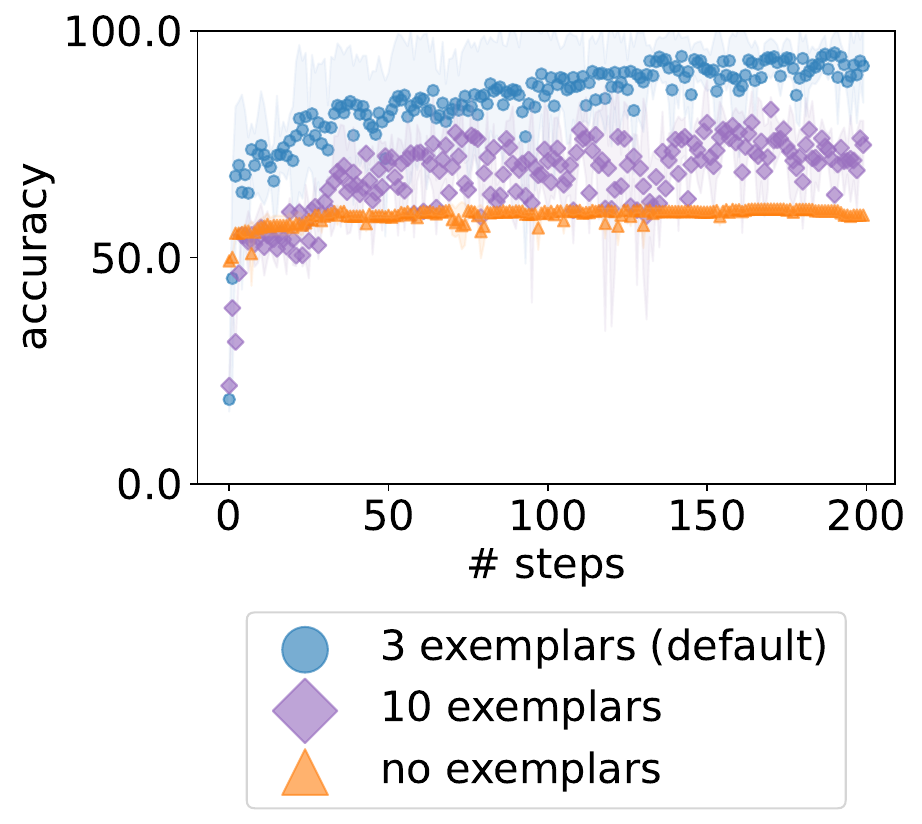}}

\caption{\textbf{Ablation studies: how each part of the meta-prompt matters.} 
The dots are the average values across 3 optimization repetitions, and the shaded regions represent standard deviations.}
\label{fig:ablation_meta_prompt}
\end{figure}

\myparagraph{Meta-prompt design}
The meta-prompt design is crucial in achieving good prompt optimization performance. We investigate the following core design choices:
\begin{itemize}[leftmargin=2em,topsep=0pt,partopsep=1ex,parsep=0ex]
\item \emph{The order of the previous instructions.}
We compare the following options: (1) from lowest to highest (our default setting); (2) from highest to lowest; (3) random.
Figures~\ref{fig:ablation_meta_prompt_ins_ordering_gsm8k} and~\ref{fig:ablation_meta_prompt_ins_ordering_sports} show that the default setting achieves better final accuracies and converges faster.
One hypothesis is that the optimizer LLM output is affected more by the past instructions closer to the end of the meta-prompt. 
This is consistent with the recency bias observed in~\citet{zhao2021calibrate}, which states that LLMs are more likely to generate tokens similar to the end of the prompt.

\item \emph{The effect of instruction scores.}
In terms of how to present the accuracy scores, we compare three options: (1) rounding the accuracies to integers, which is equivalent to bucketizing the accuracy scores to 100 buckets (our default setting); (2) bucketizing the accuracies to 20 buckets; (3) not showing the accuracies, only showing the instructions in the ascending order.
Figures~\ref{fig:ablation_meta_prompt_ins_scores_gsm8k} and~\ref{fig:ablation_meta_prompt_ins_scores_sports} show that the accuracy scores assists the optimizer LLM in better understanding the quality difference among previous instructions, and thus the optimizer LLM proposes better new instructions that are similar to the best ones in the input optimization trajectory.

\item \emph{The effect of exemplars.}
We compare three options: (1) showing 3 exemplars from the task (default); (2) showing 10 exemplars from the task; (3) no exemplars.
Figures~\ref{fig:ablation_meta_prompt_exemplars_gsm8k} and~\ref{fig:ablation_meta_prompt_exemplars_sports} show that presenting exemplars in the meta-prompt is critical, as it provides information on what the task looks like and helps the optimizer model phrase new instructions better.
However, more exemplars do not necessarily improve the performance, as a few exemplars are usually sufficient to describe the task. In addition, including more exemplars results in a longer meta-prompt with a dominating exemplar part, which may distract the optimizer LLM from other important components like the optimization trajectory.
\end{itemize}

\begin{figure}
\centering
\subfigure[GSM8K]{\label{fig:ablation_num_ins_in_each_step_gsm8k}\includegraphics[width=.46\linewidth]{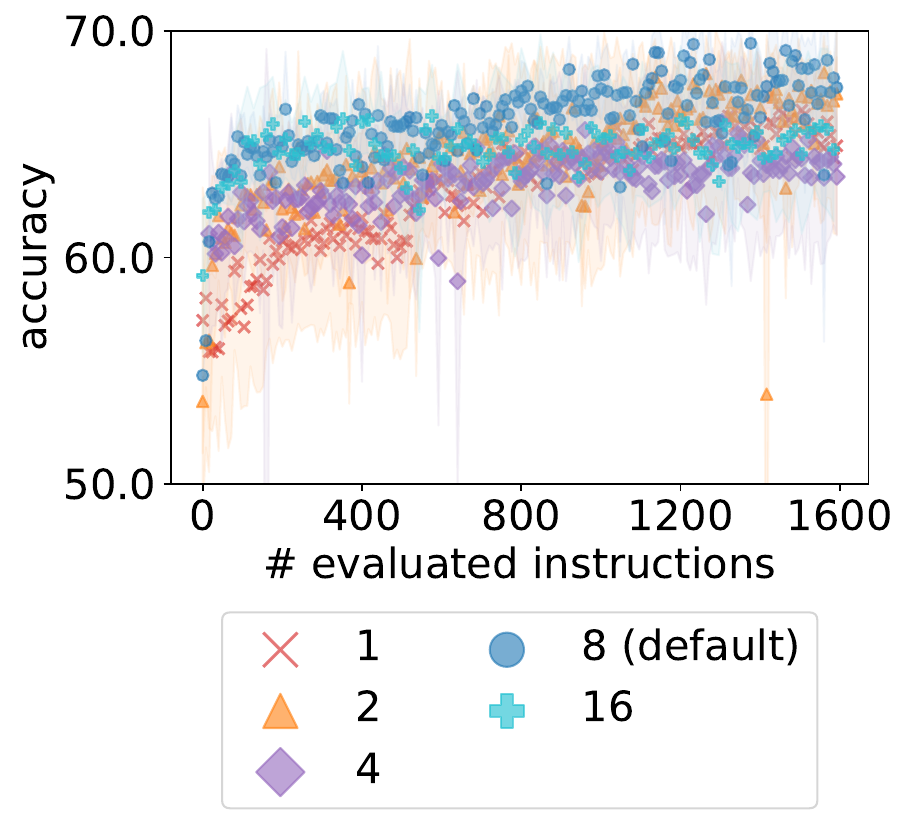}}
\hspace{.01\linewidth}
\subfigure[BBH sports\_understanding]{\label{fig:ablation_num_ins_in_each_step_sports}\includegraphics[width=.46\linewidth]{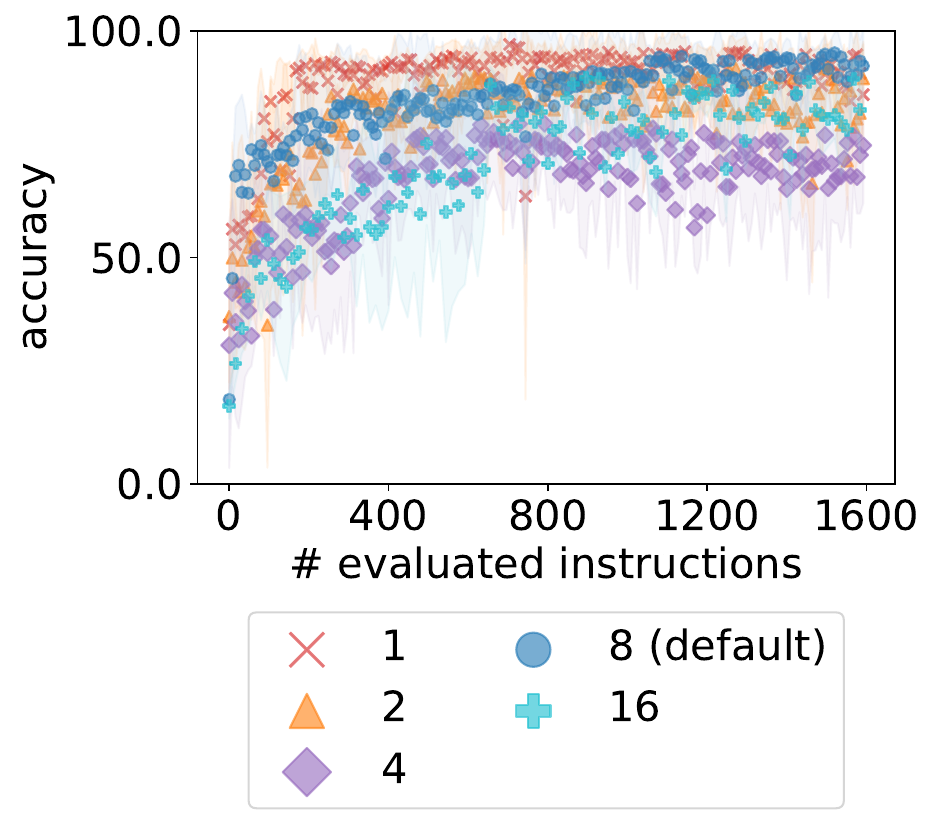}}
\caption{\textbf{Ablation studies: the number of generated instructions in each step.} 
The dots are the average values across 3 optimization repetitions, and the shaded regions represent standard deviations.
The x-axis represents the total number of evaluated instructions through the optimization; e.g., we run 200 optimization steps when sampling 8 instructions in each step, run 400 steps when sampling 4 instructions in each step, etc.}
\label{fig:ablation_num_ins_in_each_step}
\end{figure}

\myparagraph{The number of generated instructions per step}
Computing a mini-batch of gradients reduces the variance of a stochastic gradient descent procedure.
Similarly, generating multiple instructions in each step improves the optimization stability with LLMs.
On the other hand, to achieve better performance with a fixed budget for the number of instructions to evaluate, the number of per-step instructions should not be too large, so as to allow more optimization steps to incorporate richer information of past instructions with their accuracies.
Taking both aspects into consideration, Figure~\ref{fig:ablation_num_ins_in_each_step} compares the optimization performance of sampling 1 / 2 / 4 / 8 (default) / 16 instructions in each step, showing that sampling 8 instructions at each step overall achieves the best performance.

\begin{figure}[t]
\centering
\subfigure[GSM8K, \texttt{text-bison} scorer, Q\_begin]{\label{fig:ablation_starting_point_text_bison}\includegraphics[width=.43\linewidth]{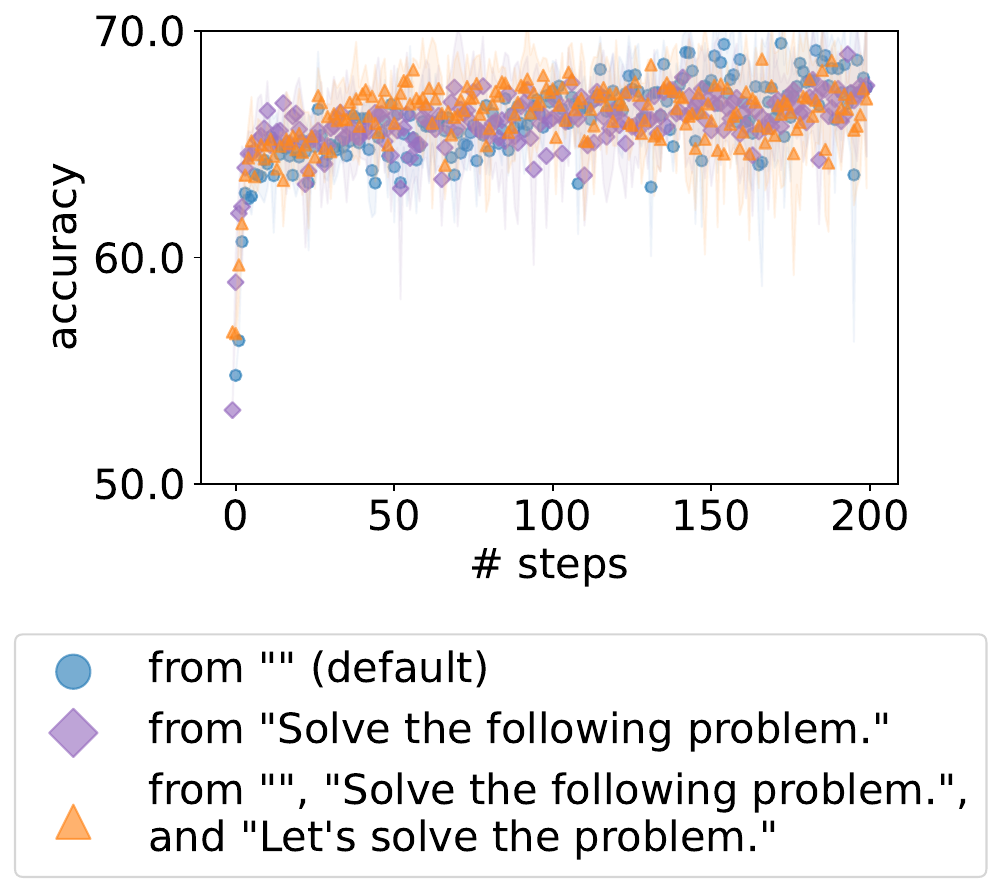}}
\hspace{.01\linewidth}
\subfigure[GSM8K, \texttt{PaLM 2-L} scorer, A\_begin]{\label{fig:ablation_starting_point_palm_2_l}\includegraphics[width=.5\linewidth]{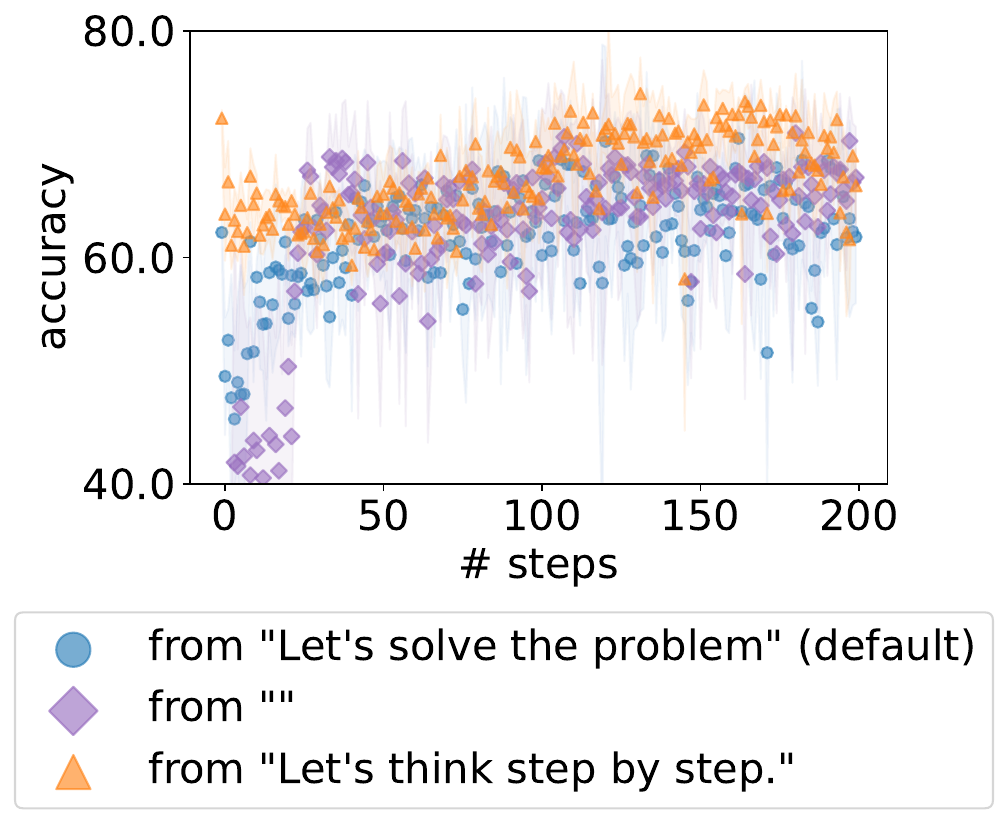}}
\caption{\textbf{Ablation studies: the initial instructions for prompt optimization.}
The dots are the average values across 3 optimization repetitions, and the shaded regions represent standard deviations.
}
\label{fig:ablation_starting_point}
\end{figure}

\myparagraph{Starting point}
We study the effect of different initial instructions for prompt optimization.
Our default setting is to start from an empty string when the scorer LLM is (instruction-tuned) \texttt{text-bison}, and to start from either the empty string (on BBH tasks) or ``Let's solve the problem.'' (on GSM8K) with instruction position A\_begin when the scorer LLM is the (pre-trained) \texttt{PaLM 2-L}.
Figure~\ref{fig:ablation_starting_point_text_bison} shows the performance of \texttt{text-bison} as the scorer LLM with 3 options of initial instructions: (1) the empty string; (2) ``Solve the following problem.''; or (3) ``Solve the following problem.'' and ``Let's solve the problem.''. We observe that the accuracies do not differ much with different starting points.
Interestingly, the styles of the generated instructions are also similar. For example, most of the generated instructions starting from (1) and (2) contain the phrase ``solve this problem'', like ``Let's work together to solve this problem.'' in Step 4 with training accuracy 64.8 from (1), and ``Let's solve the following problems using the given information.'' in Step 3 with training accuracy 62.8 from (2).

Figure~\ref{fig:ablation_starting_point_palm_2_l} presents the results of  of \texttt{PaLM 2-L} as the scorer LLM  with the following options of initial instructions: (1) ``Let's solve the problem.''; (2) the empty string; or (3) ``Let's think step by step.''. We notice that the performance differs much more with different initial instructions, especially at the beginning of the optimization.
Specifically, starting from (1) leads to better generated instructions than (2) in the first 30 steps, while the instructions optimized from both (1) and (2) are worse than (3) throughout.
A similar observation holds when using \texttt{PaLM 2-L} as scorer and \texttt{gpt-3.5-turbo} as optimizer for BBH tasks, by comparing the results starting from the empty string (Appendix~\ref{appsec:bbh_taskwise_detailed_results_gpt_3.5_turbo_optimizer_start_from_empty}) and from ``Let's solve the problem.'' (Appendix~\ref{appsec:bbh_taskwise_detailed_results_gpt_3.5_turbo_optimizer_start_from_solve}).
Taking a closer look into the optimization process of (2), we find that although both ``solve the problem'' and ``step by step'' show up in generated instructions at Step 5, it takes the optimizer LLM more steps to get rid of worse instructions presented in the meta-prompt when starting from instructions with lower accuracies.
Therefore, one direction for future work is to accelerate convergence from weaker starting points.

\begin{figure}[t]
\centering
\subfigure[GSM8K]{\label{fig:ablation_temperature_gsm8k}\includegraphics[width=.4\linewidth]{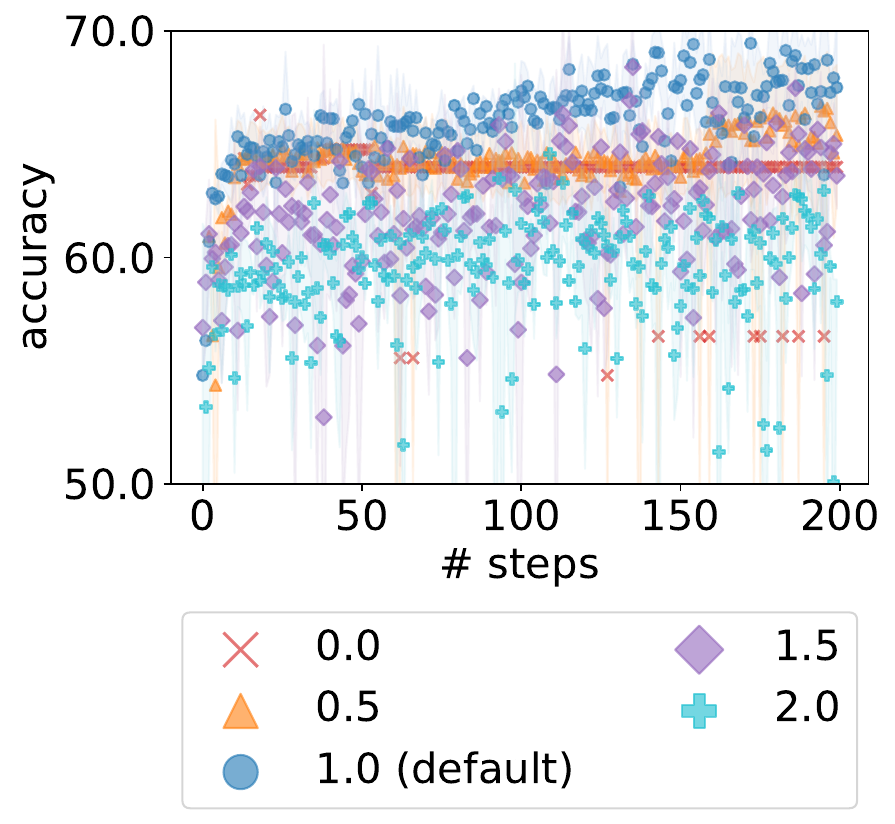}}
\hspace{.01\linewidth}
\subfigure[BBH sports\_understanding]{\label{fig:ablation_temperature_bbh_sports}\includegraphics[width=.4\linewidth]{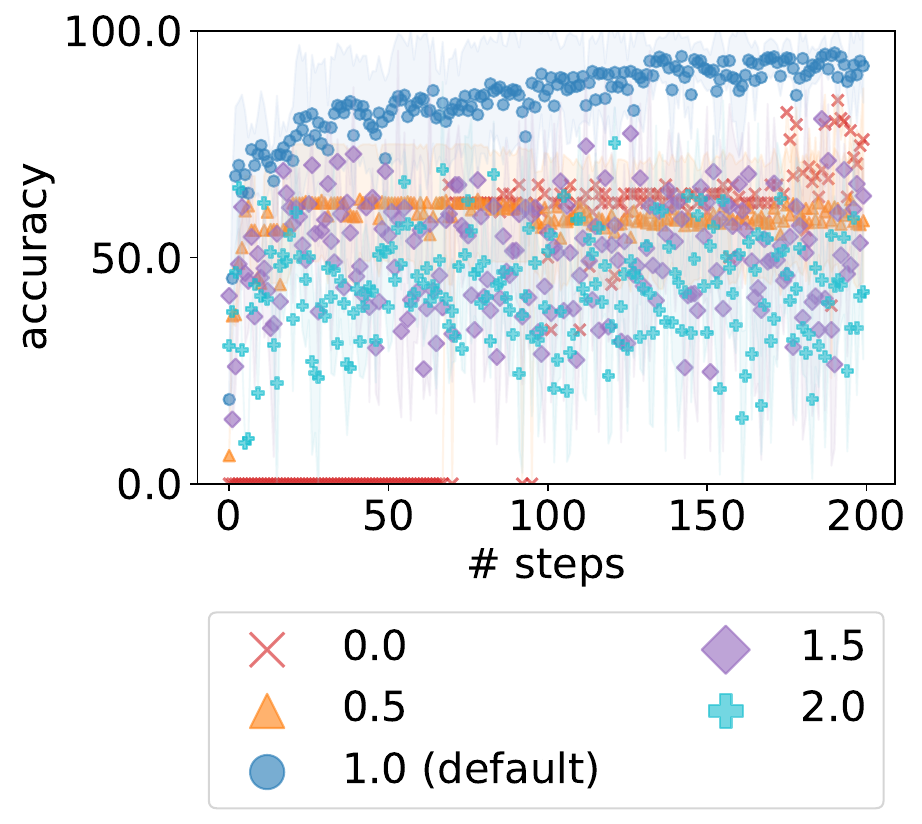}}
\caption{\textbf{Ablation studies: temperature of the optimizer model.} 
The dots are the average values across 3 optimization repetitions, and the shaded regions represent standard deviations.
}
\label{fig:ablation_temperature}
\end{figure}

\myparagraph{Diversity per step} 
We evaluate the following temperatures of the optimizer LLM: \{0.0, 0.5, 1.0 (default), 1.5, 2.0\}.
Figure~\ref{fig:ablation_temperature} shows the default temperature 1.0 achieves the best performance.
Specifically, optimizations with smaller temperatures (0.0 and 0.5) lack exploration and thus creativity, and the optimizer LLM often gets stuck at the same instruction for tens of steps, resulting in flat optimization curves.
On the other hand, with larger temperatures (1.5 and 2.0), the optimizer LLM more often ignores the trajectory of previous instructions presented in the meta-prompt and thus lacks exploitation, therefore the optimization curve does not have a steady upward trend.

\myparagraph{Comparison with one-step instruction generation}
Our current iterative procedure runs for multiple steps and generates a new batch of solutions in each step.
To validate the importance of leveraging the optimization trajectory for generating new prompts, we compare to a baseline that generates all instructions in a single step without entering into the optimization procedure.
We compare these two approaches on GSM8K and BBH sports\_understanding with the \texttt{PaLM 2-L-IT} optimizer. 
For GSM8K the scorer LLM is pre-trained \texttt{PaLM 2-L} and the initial instruction is “Let’s solve the problem”, and for BBH sports\_understanding the scorer LLM is \texttt{text-bison} and the initial instruction is the empty string. 
The baseline generates 50 instructions in a single step, thus its meta-prompt only includes task exemplars, the initial instruction with its accuracy, and the same meta-instructions as our full meta-prompt for performing optimization. 
All the other hyperparameters remain the same.

Our results show that this one-step instruction generation performs much worse than our optimization approach. Specifically:
(1) On GSM8K, the best instruction among all 50 is still ``Let's solve the problem'', with a 64.4 training accuracy and a 60.8 test accuracy. On the other hand, our approach (corresponding to Figure~\ref*{fig:prompt_optimization_graph_in_intro_gsm8k} in the main paper) found ``Let’s do the math!'' with a 78.2 training accuracy and a 76.3 test accuracy at the 5th step by generating 8 instructions at each step.
(2) Similarly, on BBH sports\_understanding, the best instruction among all 50 achieved a 84.0 training accuracy and 80.0 test accuracy. This is again worse than the instruction found by our approach at Step 4, which achieved a 88.0 training accuracy and a 84.5 test accuracy.

\subsection{Overfitting Analysis in Prompt Optimization}
\label{sec:overfitting_analysis_in_prompt_optimization}
For simplicity, we do not set aside a validation set in our default setting of prompt optimization.
We made this decision based on the experiments when a validation set is present.

\begin{figure}[t]
\centering
\subfigure[BBH snarks, \texttt{PaLM 2-L} as scorer, \texttt{PaLM 2-L-IT} as optimizer, starting from ``Let’s solve the problem.'']{\label{fig:overfitting_analysis_bbh_snarks}\includegraphics[width=.43\linewidth]{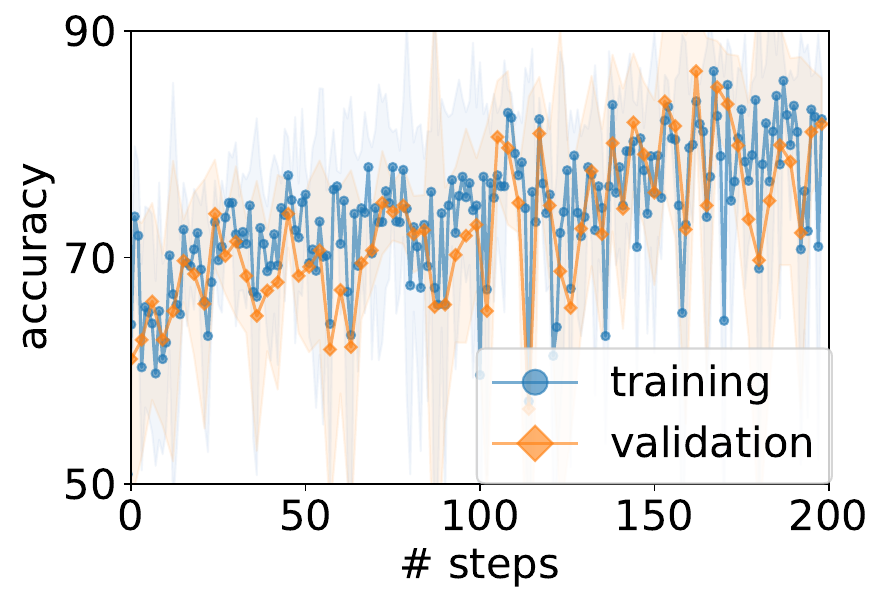}}
\hspace{.05\linewidth}
\subfigure[BBH sports\_understanding, \texttt{text-bison} as scorer, \texttt{gpt-3.5-turbo} as optimizer, starting from the empty string]{\label{fig:overfitting_analysis_bbh_sports}\includegraphics[width=.43\linewidth]{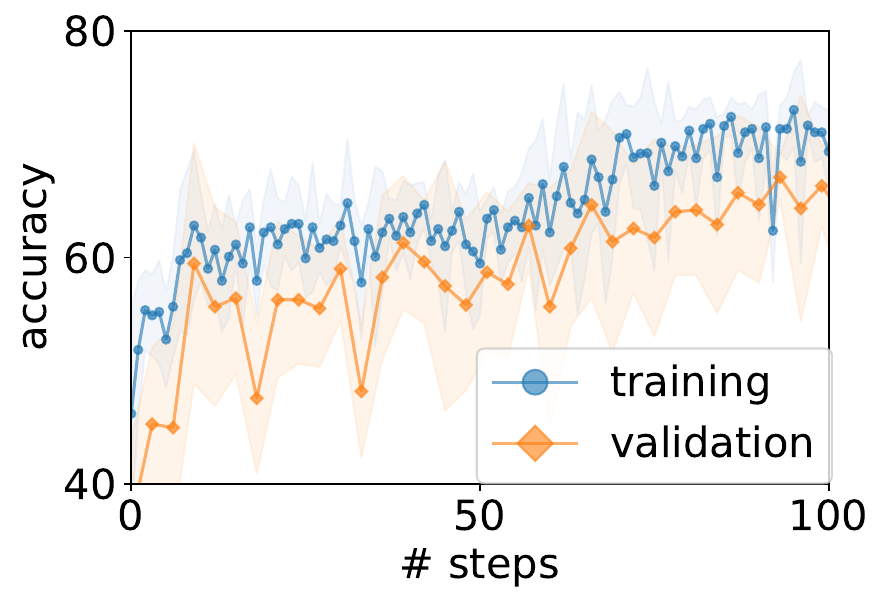}}
\caption{\textbf{Overfitting analysis.}
The exemplars are splitted to 1/3 training, 1/3 validation and 1/3 test.
We compute the validation accuracy every 3 steps.
The training/validation dots are the average training/validation accuracies across 3 optimization repetitions, respectively, and the shaded regions represent standard deviations.
}
\label{fig:overfitting_analysis}
\end{figure}

Overfitting may result in training accuracy being much higher than the validation/test accuracy.
It is difficult to avoid overfitting, but overfitting is less harmful when each candidate solution (natural language instruction in the prompt optimization context) overfits to a similar extent.
In this case, a higher training accuracy solution still achieves a higher validation/test accuracy, and one can adopt solutions with the highest training accuracies as the final result.
Figure~\ref{fig:overfitting_analysis} shows this is the case for OPRO in prompt optimization: when setting aside a validation set with the same size as the training set, the validation accuracy curves trend up and down alongside the training curves in both prompt optimization settings.

Of course, overfitting still occurs in the instructions found by our prompt optimization: in Table~\ref{table:palm2_scores_on_bbh_tasks} and~\ref{table:gpt_scores_on_bbh_tasks_starting_from_empty}, our training accuracies are often 5\%-20\% higher than our test accuracies, despite that our test and overall accuracies are still mostly higher than human-written counterparts.
Setting aside a larger training set and optimizing for fewer steps (early stopping) may help reduce overfitting.

\subsection{Comparison with EvoPrompt}
\label{sec:comparison_with_evoprompt}
Some concurrent works on prompt optimization propose meta-prompts that explicitly ask the LLM to perform mutation and crossovers of existing prompts~\citep{fernando2023promptbreeder,guo2023connecting}. In our evaluation, we compare our approach to the Genetic Algorithm (GA) and Differential Evolution (DE) versions of EvoPrompt~\citep{guo2023connecting}. Specifically, in the GA meta-prompt, given two prompts, the meta-prompt instructs the LLM to cross over the two prompts and generates a new one, then mutates the newly generated prompt to produce the final prompt. DE extends the GA meta-prompt to include more detailed instructions, e.g., asking the LLM to identify different parts between the two given prompts before performing the mutation. This is in contrast with OPRO, which leverages the optimization trajectory including multiple past prompts, instead of only 2 previous prompts. Meanwhile, OPRO also provides the LLM with richer information to facilitate the understanding of the optimization problem, including exemplars and task accuracies of different prompts.

\begin{figure}[t]
\centering
\subfigure[GSM8K, \texttt{PaLM 2-L} scorer, A\_begin]{\label{fig:comparison_with_evoprompt_gsm8k}\includegraphics[width=.43\linewidth]{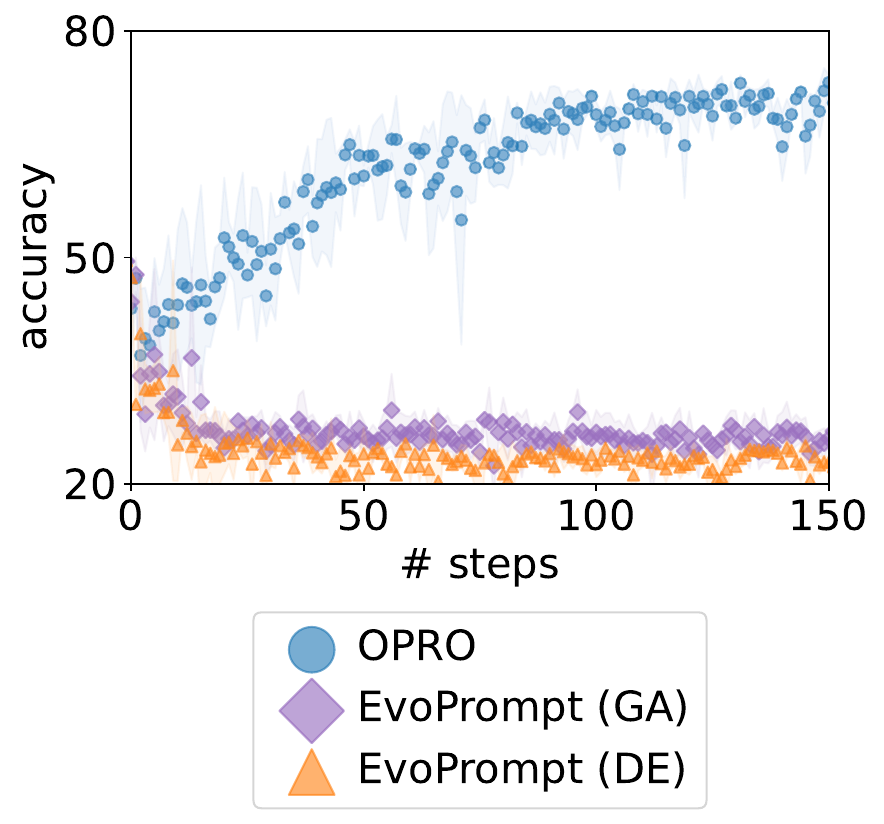}}
\hspace{.01\linewidth}
\subfigure[BBH sports\_understanding, \texttt{text-bison} scorer, Q\_begin]{\label{fig:comparison_with_evoprompt_bbh_sports}\includegraphics[width=.43\linewidth]{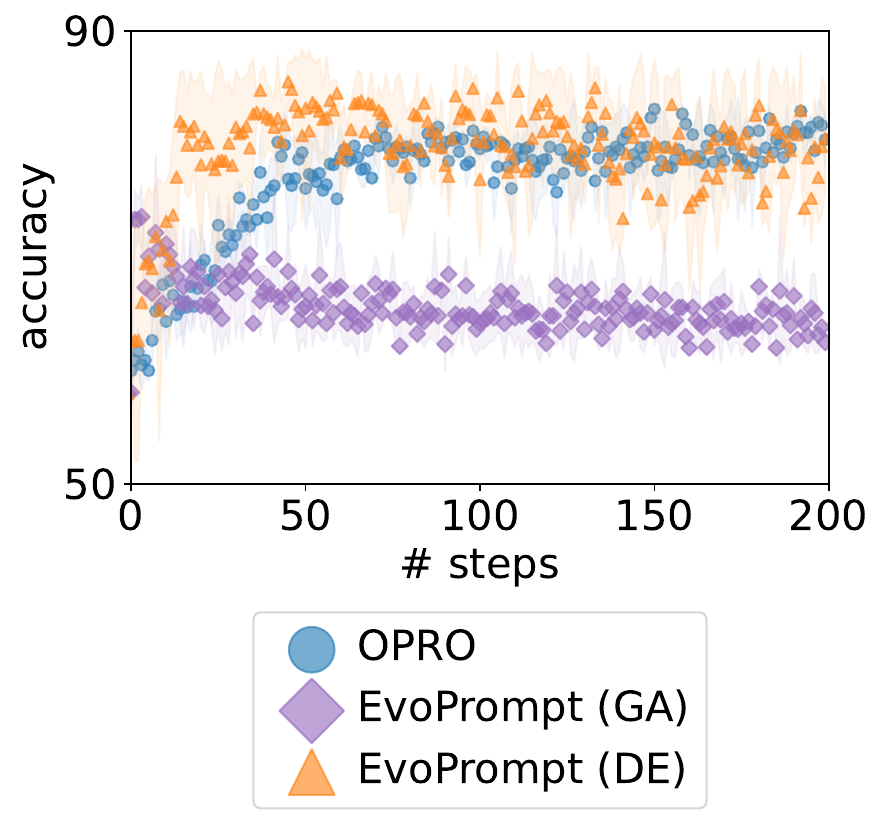}}
\caption{\textbf{Comparison with EvoPrompt in prompt optimization.}
We use the \texttt{gpt-3.5-turbo} optimizer for both experiments.
``EvoPrompt (GA)'' uses the meta-prompt from~\citet{guo2023connecting}, Figure 1; ``EvoPrompt (DE)'' uses the meta-prompt from~\citet{guo2023connecting}, Figure 2.
All optimizations in~\subref{fig:comparison_with_evoprompt_gsm8k} use the pre-trained \texttt{PaLM 2-L} scorer and start from two simple instructions ``Let's solve the problem.'' and ``Here is the answer.''; all optimizations in~\subref{fig:comparison_with_evoprompt_bbh_sports} use the \texttt{text-bison} scorer and start from two richer (task-specific) instructions ``Solve the sports understanding problem.'' and ``Give me the answer to sports understanding.''.
The dots are the average values across 3 optimization repetitions, and the shaded regions represent standard deviations.
We use temperature 1.0 for OPRO and temperature 0.5 for EvoPrompt, same as the default settings in respective works.
}
\label{fig:comparison_with_evoprompt}
\end{figure}

Figure~\ref{fig:comparison_with_evoprompt} presents the results on GSM8K and BBH sports\_understanding benchmarks, where we use \texttt{gpt-3.5-turbo} as the optimizer. On GSM8K, the initial instructions of all approaches are ``Let's solve the problem.'' and ``Here is the answer.'', which are simple and generic. Again, we observe that OPRO performance steadily improves with more optimization steps. On the other hand, both versions of EvoPrompt even degrade the performance on GSM8K. The main reason is because EvoPrompt does not utilize exemplars for prompt optimization, thus it lacks the understanding of the task to optimize for. In this way, EvoPrompt relies on good-quality and task-specific initial prompts to optimize from.

Given this observation, we provide more task-specific initial instructions for experiments on BBH sports\_understanding, which are ``Solve the sports understanding problem.'' and ``Give me the answer to sports understanding.'' In this case, EvoPrompt (DE) is able to find better prompts than the initial ones, but the optimization curve is less stable than OPRO. This indicates that leveraging the optimization trajectory helps the LLM to identify promising directions to improve existing prompts.

\section{Related Work}
\label{sec:work}

\myparagraph{Prompt optimization} Prior works have developed soft prompt-tuning methods that optimize the prompt represented as task-specific continuous vectors~\citep{lester2021power,li2021prefix,liu2021gpt,qin2021learning}, as well as performing discrete prompt optimization by gradient-guided search~\citep{shin2020autoprompt,wen2023hard,gao2020making,chen2023instructzero} and reinforcement learning~\citep{deng2022rlprompt,zhang2022tempera}. 
These approaches become inapplicable when there is only API access to the LLM. 
Other works designed edit-based approaches for gradient-free prompt optimization~\citep{xu2022gps,prasad2022grips}, where the editing can be done with human-defined operations (e.g., swapping two phrases)~\citep{prasad2022grips} or language models (e.g., back translation)~\citep{xu2022gps}. 
Some recent works investigate LLMs for prompt optimization~\citep{zhou2022large,pryzant2023automatic,xu2023wizardlm}. 
Specifically, APE~\citep{zhou2022large} first uses the LLM to generate initial instructions. Afterwards, APE selects top instructions with the highest accuracies, then prompts the LLM with each individual instruction to generate a semantically similar variant of the initial instruction.
APO~\citep{pryzant2023automatic} in each step instructs the LLM to produce text feedback on how to update an old instruction. 
Different from edit-based approaches, the optimizer LLM in our work directly generates new instructions at each optimization step, and the optimizer LLM is merely asked to improve the task accuracy without being required to imitate past instructions. Compared to \citet{zhou2022large} and \citet{pryzant2023automatic}, our optimization process incorporates the past generated instructions with their scores in the meta-prompt, enabling the optimizer LLM to discover common patterns of high-quality instructions.

\myparagraph{Prompting with natural language feedback} A recent line of work investigates approaches to improve the LLM performance by prompting with natural language feedback to revise the model output, which has shown effectiveness in reducing harmful LLM outputs~\citep{bai2022constitutional,ganguli2023capacity}, improving reasoning~\citep{shinn2023reflexion,madaan2023self} and code generation performance~\citep{chen2023teaching,olausson2023demystifying,shinn2023reflexion,chen2023improving}, dialogue applications~\citep{nair2023dera,madaan2023self,yuan2023system}, and so on~\citep{kim2023language,wang2023voyager}. Specifically, \citet{yuan2023system} develops a human-in-the-loop framework for deriving system-level feedback from a collection of instance-level feedback, which is then used for refining data. In our work, the optimizer LLM utilizes the optimization trajectory in the prompt, which implicitly requires the LLM to summarize the common characteristics among solutions with similar scores. We consider incorporating explicit natural language feedback on generated solutions for later optimization steps as future work.

\myparagraph{Tuning language models for optimization}
Some previous works tune or prompt language models to behave as mutation and crossover operators in evolutionary algorithms.
\citet{meyerson2023language} utilizes language models with few-shot exemplars to propose evolutionary cross-overs on tasks such as image and code generation.
In \citet{lehman2022evolution}, the large language model trained on code diff generation is used as the mutation operator, and they further design a fine-tuning method to improve performance in the Sodarace domain for robot simulation.
EvoPrompting~\citep{chen2023evoprompting} uses large language models to evolve neural network architectures, where they combine evolutionary search with soft prompt tuning.
With respect to taking the trajectory as the input for optimization, OptFormer~\citep{chen2022towards} trains a transformer model on large collections of hyperparameter optimization data.
On the other hand, our work performs optimization solely by prompting without additional training.

\section{Conclusion}
\label{sec:conclusion}
We embark on employing LLMs as optimizers, where the LLM progressively generates new solutions to optimize an objective function.
We first motivate \name{} with linear regression and traveling salesman problems, then proceed to prompt optimization as a concrete application.
Our evaluation demonstrates that LLMs have the capacity of gradually improving the generated solutions based on the past optimization trajectory.
Interestingly, on small-scale traveling salesman problems, \name{} performs on par with some hand-crafted heuristic algorithms.
For prompt optimization, optimized prompts outperform human-designed prompts on GSM8K and Big-Bench Hard by a significant margin, sometimes over $50\%$.

A number of unresolved questions are open for future research on LLMs for optimization. 
In general, how to reduce the sensitivity to initialization and better balance exploitation with exploration remains a challenge. 
Specifically, for prompt optimization, one limitation of our current implementation is that the optimizer LLM does not effectively utilize error cases in the training set to infer promising directions to improve the generated instructions. 
In our experiments, we tried including error cases in the meta-prompt rather than randomly sampling from the training set at each optimization step, but the results are similar, indicating that the error cases alone are not informative enough for the optimizer LLM to grasp the cause of the wrong prediction.
Another limitation is that prompt optimization requires a training set to compute the accuracy that guides the optimization process. 
Currently the training set at least contains tens of samples, so that the optimized prompt does not severely overfit to the training samples. 
A promising direction is to incorporate richer feedback about the error cases besides the aggregated accuracy, and summarize the key features that distinguish between high-quality and low-quality generated prompts in the optimization trajectory.
Such information may inform the optimizer LLM of how to more efficiently improve over the past generated instructions, and potentially further reduce the example set size needed for prompt optimization.

\section*{Ethics Statement}
This work uses synthetic math problems for linear regression and traveling salesman problems, and uses public datasets like GSM8K and Big-Bench Hard for prompt optimization.
These tasks have been commonly used in similar works and should not be regarded controversial.
There is a peril that LLMs may generate harmful information that poses safety risks; how to safeguard model behavior remains valuable future work.

\section*{Reproducibility Statement}

We evaluate on public benchmarks. The \texttt{text-bison} API is available at:~\url{https://cloud.google.com/vertex-ai/docs/generative-ai/learn/models}. The GPT models are available here:~\url{http://openai.com/api/}. 
This work uses \texttt{gpt-3.5-turbo-0613} and \texttt{gpt-4-0613}.

\section*{Acknowledgments}

We thank Daiyi Peng, Yanqi Zhou, Jerry Wei, Shuo Chen, Tim Rocktäschel, Chrisantha Fernando, Dylan Banarse, Henryk Michalewski, Simon Osindero, and Ed H. Chi for their valuable feedback, and thank several anonymous reviewers for helpful comments.

\bibliography{scholar}
\bibliographystyle{arxiv}

\newpage
\appendix
\section{Some Failure Cases}
\label{appsec:some_failure_cases}
Although LLMs show the power of optimizing basic math problems (Section~\ref{sec:motivating_example}) and prompts (Section~\ref{sec:application_prompt_opt}), we see some limitations across all optimizer LLMs that may impede their power of solving more challenging problems.
These limitations include:
\begin{itemize}[leftmargin=2em,topsep=0pt,partopsep=1ex,parsep=0ex]
\item \textbf{Hallucinating the values that need to come from math calculation}: The optimizer LLMs often output contents like ``the function value at (5, 3) is 15'' despite that the true value is not 15. 
The model will get it right if external tools that can reliably calculate the value are triggered.
When and how to trigger such tool use cases remains an interesting topic (see e.g., \citep{schick2023toolformer,cai2023large}).
\item \textbf{Generating solutions already appeared in context even if we tell it to "Give me a new (w, b) pair that is different from all pairs above"}: the optimizer LLMs do not 100\% reliably follow this instruction even if its own outputs often include sentences like ``I will provide a new pair that is different'', making the output self-contradictory.
The output is almost guaranteed to be different from in-context old solutions when the model output contains a comparison of the new pair and all old pairs, though.
Thus (implicitly) triggering such behaviors may be a solution.
How to implement this feature without harming the instruction following performance of other parts remains an interesting topic to study.
\item \textbf{In black-box math optimization, getting stuck at a point that is neither global nor local optimal}: This often occurs in two linear regression cases: (a) The in-context exemplars all share the same $w$ or $b$ that is different from $w_\text{true}$ or $b_\text{true}$.
This case is more likely to be avoided when a larger number of past solutions are included in the meta-prompt; (b) one or several of the best previous solutions in the meta-prompt have $w$s and $b$s in quantitatively opposite directions from the global optima $w_\text{true}$ and $b_\text{true}$: for example, the $w$s are all smaller than $w_\text{true}$ while the $b$s are all larger than $b_\text{true}$.
Since the optimizer model often proposes to only increase $w$ or decrease $b$ when the past solutions in meta-prompt share $w$ or $b$, the optimization will get stuck if either increasing $w$ or decreasing $b$ would increase the objective value.
This issue is mitigated by sampling multiple new solutions (thus more exploration) at each step.
\item \textbf{Hard to navigate a bumpy loss landscape}: Like other optimizers, it is harder for the optimizer LLM to optimize black-box functions when the loss landscape gets more complicated.
For example, when minimizing the Rosenbrock function $f(x, y) = (a - x)^2 + b (y - x^2)^2$ with $a=20$ (whose global optimal point is $x=20$, $y=400$) with 5 starting points in $[10, 20] \times [10, 20]$, the optimization often gets stuck at around $(0, 0)$.
This is because the optimizer LLM sees a decrease of objective value when it drastically decreases both $x$ and $y$ to $0$.
Then starting from $(0, 0)$, the optimizer LLM is hard to further navigate $x$ and $y$ along the narrow valley in the loss landscape towards $(20, 400)$ (Figure~\ref{fig:rosenbrock_visualization}).

\begin{figure}[H]
\centering
\subfigure{\includegraphics[width=.4\linewidth]{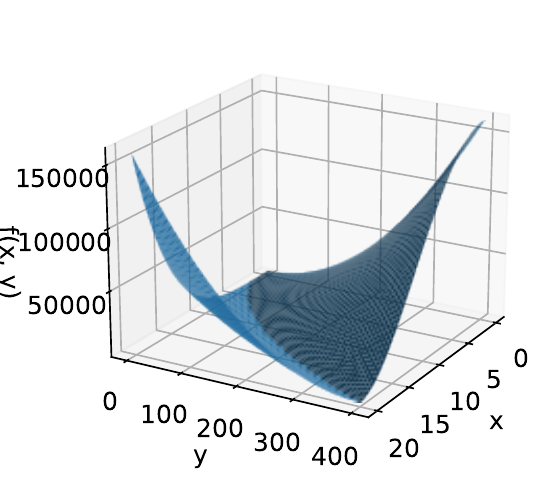}}
\caption{A visualization of the landscape of the Rosenbrock function $f(x, y) = (a - x)^2 + b (y - x^2)^2$ with $a=20$ and $b=1$.
The global optima is at $x=20$, $y=400$ with function value $0$.
The function value at $x=0$, $y=0$ is $400$.
The landscape has a narrow valley between $(0, 0)$ and $(20, 400)$.
}
\label{fig:rosenbrock_visualization}
\end{figure}

\end{itemize}

\section{Prompting Formats for Scorer LLM}
\label{appsec:scorer_prompting_formats}
Figure~\ref{fig:scorer_prompting_format_Q_begin}, \ref{fig:scorer_prompting_format_Q_end}, and~\ref{fig:scorer_prompting_format_A_begin} show examples of the Q\_begin, Q\_end, and A\_begin prompting formats when the ``QA'' pattern is present.
The ``QA'' pattern is eliminated when prompting instruction-tuned scorer models like \texttt{text-bison} with the Q\_begin and Q\_end formats (Figure~\ref{fig:scorer_prompting_format_Q_begin_no_QA} and~\ref{fig:scorer_prompting_format_Q_end_no_QA}).

\begin{figure}[H]
\noindent\fbox{
\parbox{\textwidth}{
Q: \textcolor{burntorange}{\{instruction\}}

Janet’s ducks lay 16 eggs per day. She eats three for breakfast every morning and bakes muffins for her friends every day with four. She sells the remainder at the farmers' market daily for \$2 per fresh duck egg. How much in dollars does she make every day at the farmers' market?

\vspace{1em}

A:
}
}
\caption{The Q\_begin prompting format on a GSM8K test exemplar with the "QA" pattern.}
\label{fig:scorer_prompting_format_Q_begin}
\end{figure}

\begin{figure}[H]
\noindent\fbox{
\parbox{\textwidth}{
Q: Janet’s ducks lay 16 eggs per day. She eats three for breakfast every morning and bakes muffins for her friends every day with four. She sells the remainder at the farmers' market daily for \$2 per fresh duck egg. How much in dollars does she make every day at the farmers' market?

\textcolor{burntorange}{\{instruction\}}
\vspace{1em}

A:
}
}
\caption{The Q\_end prompting format on a GSM8K test exemplar with the "QA" pattern.}
\label{fig:scorer_prompting_format_Q_end}
\end{figure}

\begin{figure}[H]
\noindent\fbox{
\parbox{\textwidth}{
Q: Janet’s ducks lay 16 eggs per day. She eats three for breakfast every morning and bakes muffins for her friends every day with four. She sells the remainder at the farmers' market daily for \$2 per fresh duck egg. How much in dollars does she make every day at the farmers' market?

\vspace{1em}

A: \textcolor{burntorange}{\{instruction\}}
}
}
\caption{The A\_begin prompting format on a GSM8K test exemplar.}
\label{fig:scorer_prompting_format_A_begin}
\end{figure}

\begin{figure}[H]
\noindent\fbox{
\parbox{\textwidth}{
\textcolor{burntorange}{\{instruction\}}

Janet’s ducks lay 16 eggs per day. She eats three for breakfast every morning and bakes muffins for her friends every day with four. She sells the remainder at the farmers' market daily for \$2 per fresh duck egg. How much in dollars does she make every day at the farmers' market?
}
}
\caption{The Q\_begin prompting format on a GSM8K test exemplar without the "QA" pattern.}
\label{fig:scorer_prompting_format_Q_begin_no_QA}
\end{figure}

\begin{figure}[H]
\noindent\fbox{
\parbox{\textwidth}{
Janet’s ducks lay 16 eggs per day. She eats three for breakfast every morning and bakes muffins for her friends every day with four. She sells the remainder at the farmers' market daily for \$2 per fresh duck egg. How much in dollars does she make every day at the farmers' market?

\textcolor{burntorange}{\{instruction\}}
}
}
\caption{The Q\_end prompting format on a GSM8K test exemplar without the "QA" pattern.}
\label{fig:scorer_prompting_format_Q_end_no_QA}
\end{figure}

\section{Meta-Prompts}
\label{appsec:meta_prompts}
\subsection{Meta-Prompt for Math Optimization}
\label{appsec:meta_prompts_for_math_opt}

\begin{figure}[H]
\noindent\fbox{
\parbox{\textwidth}{
\color{burntorange}{
Now you will help me minimize a function with two input variables w, b. I have some (w, b) pairs and the function values at those points. The pairs are arranged in descending order based on their function values, where lower values are better.
}

\vspace{1em}

\color{blue}{
input:

w=18, b=15

value:

10386334

\vspace{1em}

input:

w=17, b=18

value:

9204724
}

\vspace{1em}
\color{burntorange}{
Give me a new (w, b) pair that is different from all pairs above, and has a function value lower than any of the above. Do not write code. The output must end with a pair [w, b], where w and b are numerical values.
}
}
}
\caption{An example of the meta-prompt for linear regression. The \textcolor{blue}{blue} text contains solution-score pairs; the \textcolor{burntorange}{orange} text are meta-instructions.}
\label{fig:meta_prompt_example_linear_regression}
\end{figure}

\begin{figure}[H]
\noindent\fbox{
\parbox{\textwidth}{
\color{burntorange}{
You are given a list of points with coordinates below:
(0): (-4, 5), (1): (17, 76), (2): (-9, 0), (3): (-31, -86), (4): (53, -35), (5): (26, 91), (6): (65, -33), (7): (26, 86), (8): (-13, -70), (9): (13, 79), (10): (-73, -86), (11): (-45, 93), (12): (74, 24), (13): (67, -42), (14): (87, 51), (15): (83, 94), (16): (-7, 52), (17): (-89, 47), (18): (0, -38), (19): (61, 58).

Below are some previous traces and their lengths. The traces are arranged in descending order based on their lengths, where lower values are better.
}

\vspace{1em}

\color{blue}{
<trace> 0,13,3,16,19,2,17,5,4,7,18,8,1,9,6,14,11,15,10,12 </trace>

length:

2254

\vspace{1em}

<trace> 0,18,4,11,9,7,14,17,12,15,10,5,19,3,13,16,1,6,8,2 </trace>

length:

2017

\vspace{1em}

<trace> 0,11,4,13,6,10,8,17,12,15,3,5,19,2,1,18,14,7,16,9 </trace>

length:

1953

\vspace{1em}

<trace> 0,10,4,18,6,8,7,16,14,11,2,15,9,1,5,19,13,12,17,3 </trace>

length:

1840
}

\vspace{1em}
\color{burntorange}{
Give me a new trace that is different from all traces above, and has a length lower than any of the above. The trace should traverse all points exactly once. The trace should start with <trace> and end with </trace>.
}
}
}
\caption{An example of the meta-prompt for Traveling Salesman Problems with problem size $n=20$. 
The \textcolor{blue}{blue} text contains solution-score pairs; the \textcolor{burntorange}{orange} text are meta-instructions.}
\label{fig:meta_prompt_example_tsp}
\end{figure}

\subsection{Meta-Prompt for Prompt Optimization}
\label{appsec:meta_prompts_for_prompt_opt}
Different optimizer models work the best on different styles of meta-prompts.
Figure~\ref{fig:meta_prompt_example} in the main paper shows the meta-prompt for \texttt{PaLM 2-L-IT};
Figure~\ref{fig:meta_prompt_example_pretrained_palm_2_l} shows that for pre-trained \texttt{PaLM 2-L}; Figure~\ref{fig:meta_prompt_example_gpt} shows that for GPT models.

\begin{figure}[H]
\noindent\fbox{
\parbox{\textwidth}{
\color{burntorange}{Create a piece of text at the beginning of the answer to enhance the precision in solving diverse grade school math problems.}

\vspace{1em}

\color{blue}{
Precision: 4 <TEXT>A dime</TEXT>

\vspace{1em}

Precision: 17 <TEXT>The answer is a function. It is</TEXT>

\vspace{1em}

Precision: 19 <TEXT>So how can we find out what this equation means?</TEXT>

\vspace{1em}

Precision: 20 <TEXT>Solutions:</TEXT>
}}
}
\caption{An example of the meta-prompt for prompt optimization with pre-trained \texttt{PaLM 2-L} on GSM8K, where the generated instruction will be prepended to the beginning of the scorer LLM output (\emph{A\_begin} in Section~\ref{sec:setup}).}
\label{fig:meta_prompt_example_pretrained_palm_2_l}
\end{figure}

\begin{figure}[H]
\noindent\fbox{
\parbox{\textwidth}{
\color{burntorange}{
Your task is to generate the instruction <INS>. Below are some previous instructions with their scores. The score ranges from 0 to 100.
}

\vspace{1em}
\color{blue}{
text:

Let's figure it out!

score:

61

\vspace{1em}

text:

Let's solve the problem.

score:

63

\vspace{1em}
(… more instructions and scores …)
\vspace{1em}
}

\color{burntorange}{
Below are some problems.
}
\color{violet}{
\vspace{1em}

Problem:

Q: Alannah, Beatrix, and Queen are preparing for the new school year and have been given books by their parents. Alannah has 20 more books than Beatrix. Queen has 1/5 times more books than Alannah. If Beatrix has 30 books, how many books do the three have together?

A: <INS>

\vspace{1em}
Ground truth answer:

140

\vspace{1em}
(… more exemplars …)
\vspace{1em}
}

\color{burntorange}{
Generate an instruction that is different from all the instructions <INS> above, and has a higher score than all the instructions <INS> above. The instruction should begin with <INS> and end with </INS>. The instruction should be concise, effective, and generally applicable to all problems above.
}
}
}
\caption{An example of the meta-prompt for prompt optimization with GPT models (\texttt{gpt-3.5-turbo} or \texttt{gpt-4}) on GSM8K, where the generated instruction will be prepended to the beginning of the scorer LLM output (\emph{A\_begin} in Section~\ref{sec:setup}). The \textcolor{blue}{blue} text contains solution-score pairs; the \textcolor{violet}{purple} text describes the optimization task and output format; the \textcolor{burntorange}{orange} text are meta-instructions.}
\label{fig:meta_prompt_example_gpt}
\end{figure}

\section{Prompt Optimization Curves on the Remaining BBH Tasks}
\label{appsec:bbh_optimization_curves}

\begin{figure}[H]
\centering
\subfigure[BBH boolean\_expressions]{\label{fig:prompt_optimization_graph_bbh_boolean_expressions}\includegraphics[width=.31\linewidth]{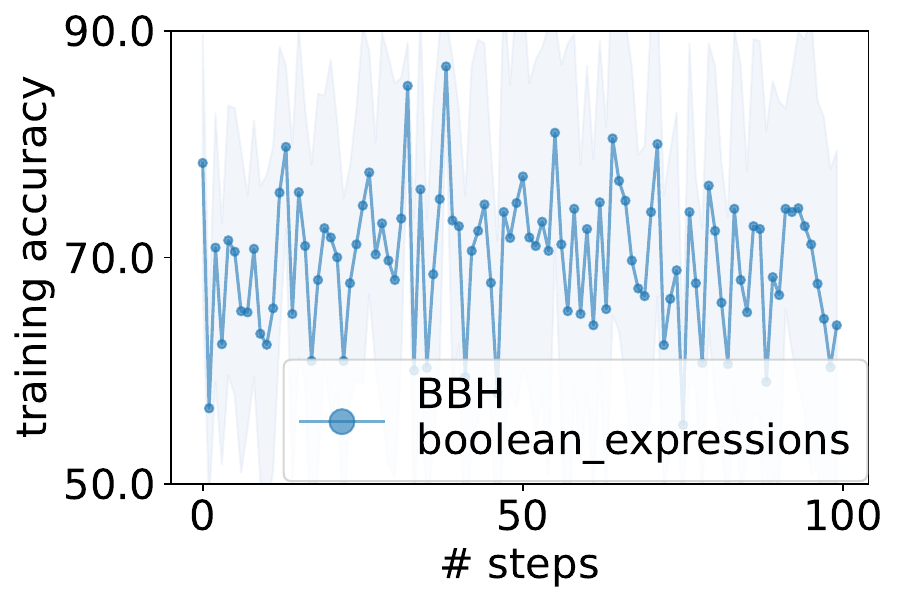}}
\hspace{.01\linewidth}
\subfigure[BBH causal\_judgement]{\label{fig:prompt_optimization_graph_bbh_causal_judgement}\includegraphics[width=.31\linewidth]{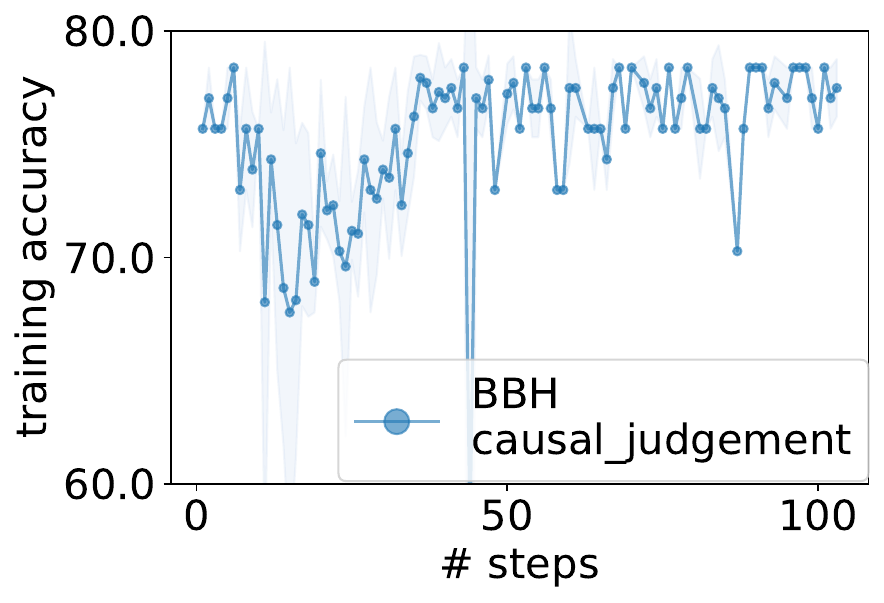}}
\hspace{.01\linewidth}
\subfigure[BBH date\_understanding]{\label{fig:prompt_optimization_graph_bbh_date_understanding}\includegraphics[width=.31\linewidth]{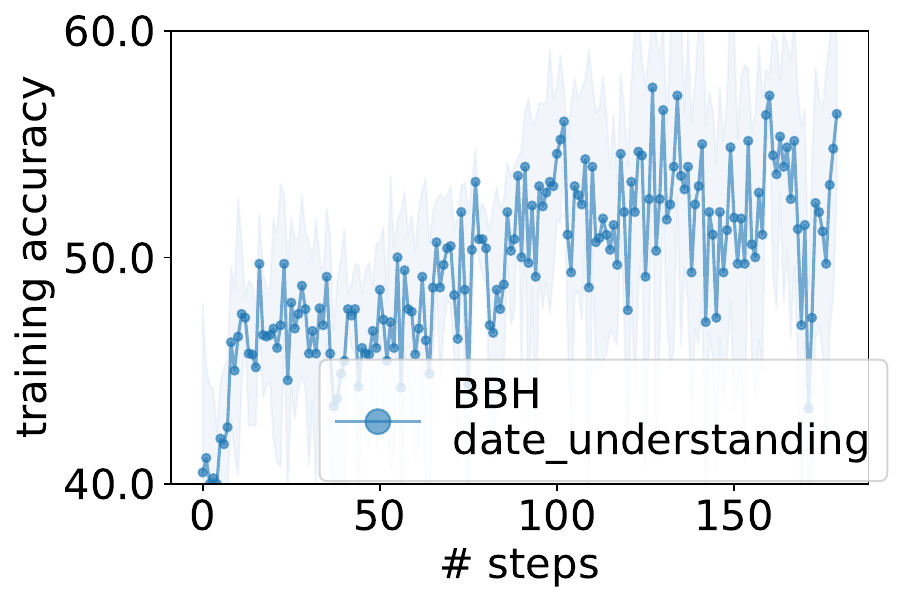}}

\subfigure[BBH disambiguation\_qa]{\label{fig:prompt_optimization_graph_bbh_disambiguation_qa}\includegraphics[width=.31\linewidth]{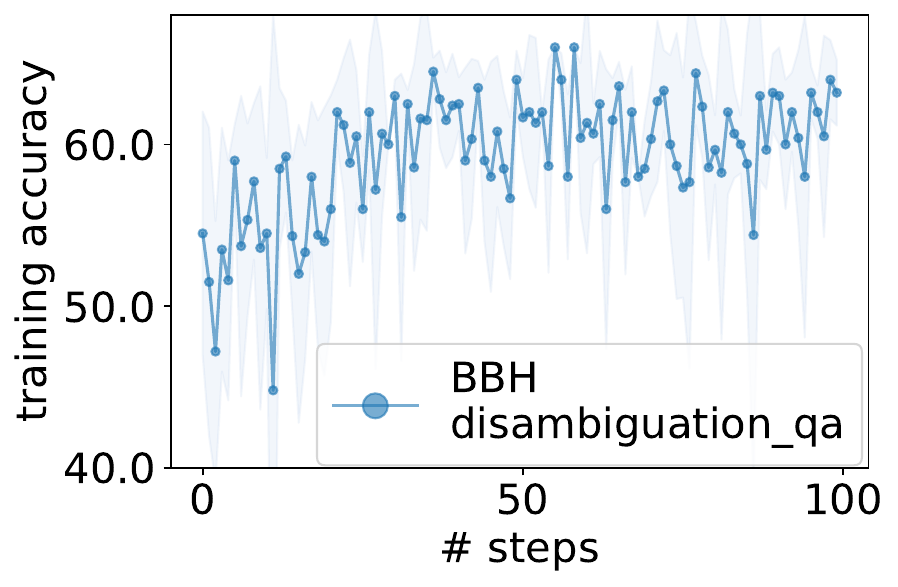}}
\hspace{.01\linewidth}
\subfigure[BBH dyck\_languages]{\label{fig:prompt_optimization_graph_bbh_dyck_languages}\includegraphics[width=.31\linewidth]{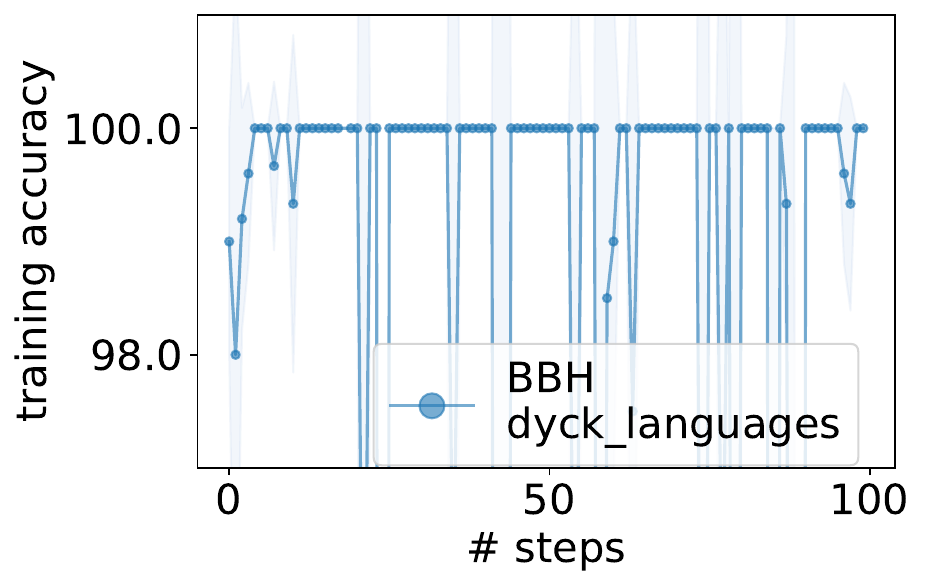}}
\hspace{.01\linewidth}
\subfigure[BBH formal\_fallacies]{\label{fig:prompt_optimization_graph_bbh_formal_fallacies}\includegraphics[width=.31\linewidth]{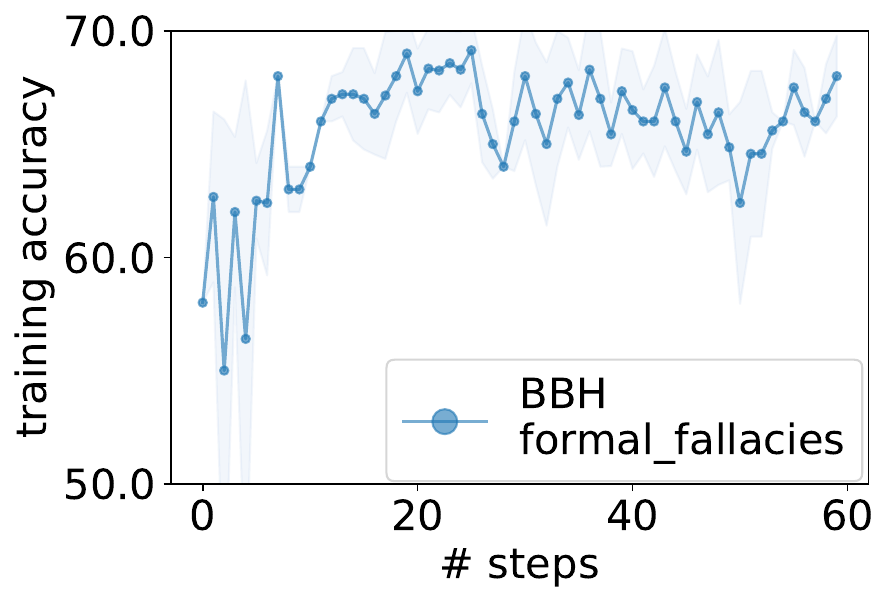}}

\subfigure[BBH geometric\_shapes]{\label{fig:prompt_optimization_graph_bbh_geometric_shapes}\includegraphics[width=.31\linewidth]{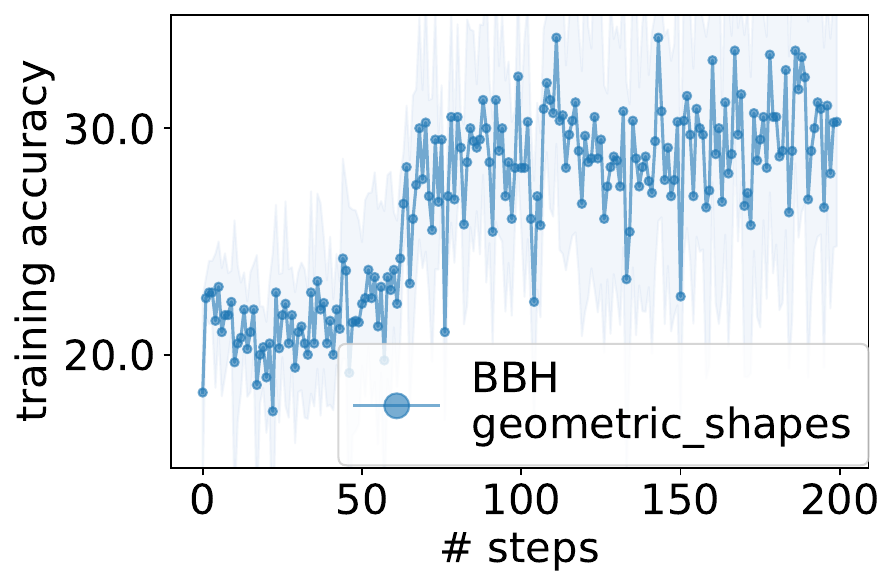}}
\hspace{.01\linewidth}
\subfigure[BBH hyperbaton]{\label{fig:prompt_optimization_graph_bbh_hyperbaton}\includegraphics[width=.31\linewidth]{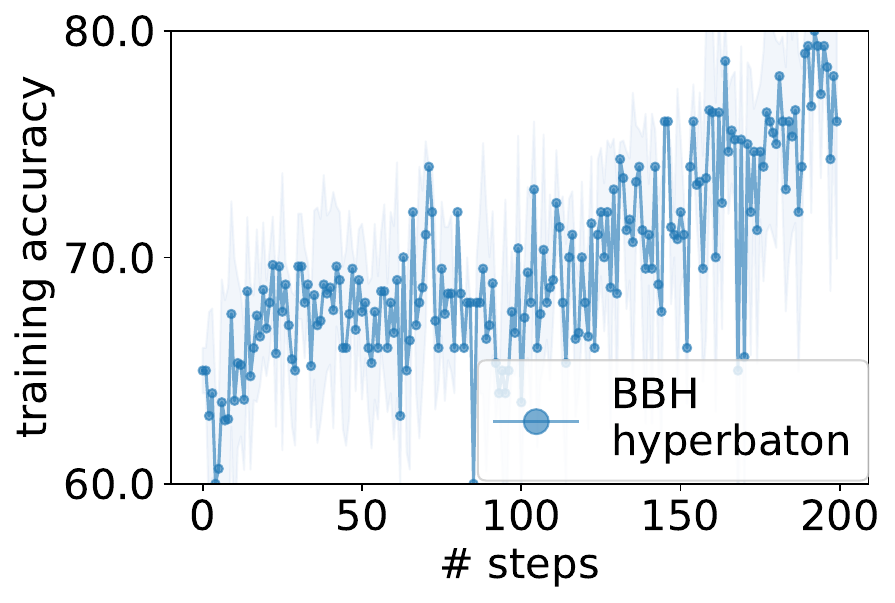}}
\hspace{.01\linewidth}
\subfigure[\scriptsize BBH logical\_deduction\_seven\_objects]{\label{fig:prompt_optimization_graph_bbh_logical_deduction_seven_objects}\includegraphics[width=.31\linewidth]{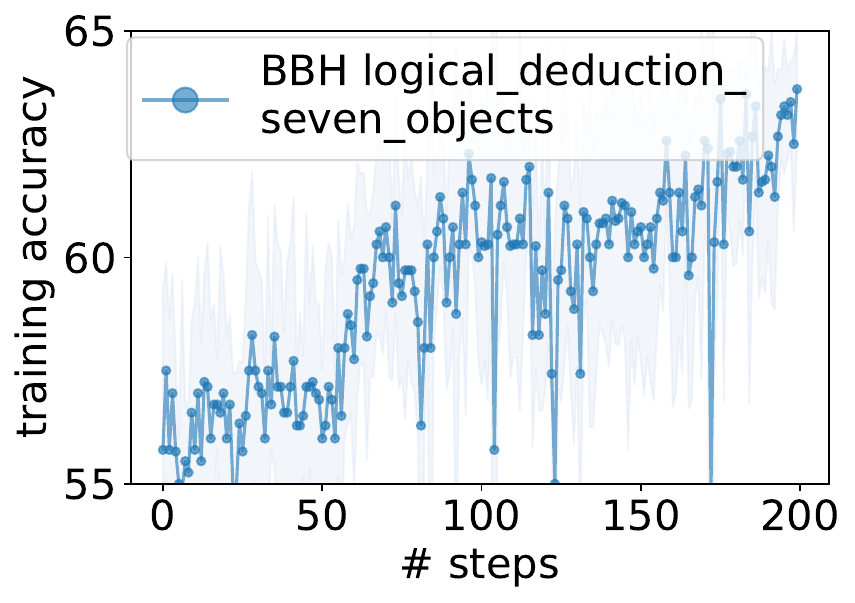}}

\subfigure[BBH movie\_recommendation]{\label{fig:prompt_optimization_graph_bbh_movie_recommendation}\includegraphics[width=.31\linewidth]{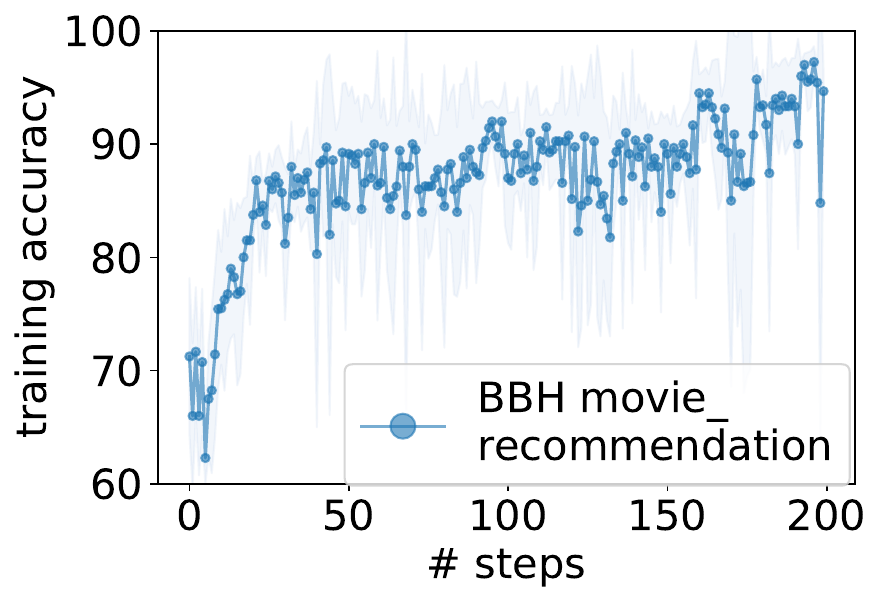}}
\hspace{.01\linewidth}
\subfigure[\scriptsize BBH multistep\_arithmetic\_two]{\label{fig:prompt_optimization_graph_multistep_arithmetic_two}\includegraphics[width=.31\linewidth]{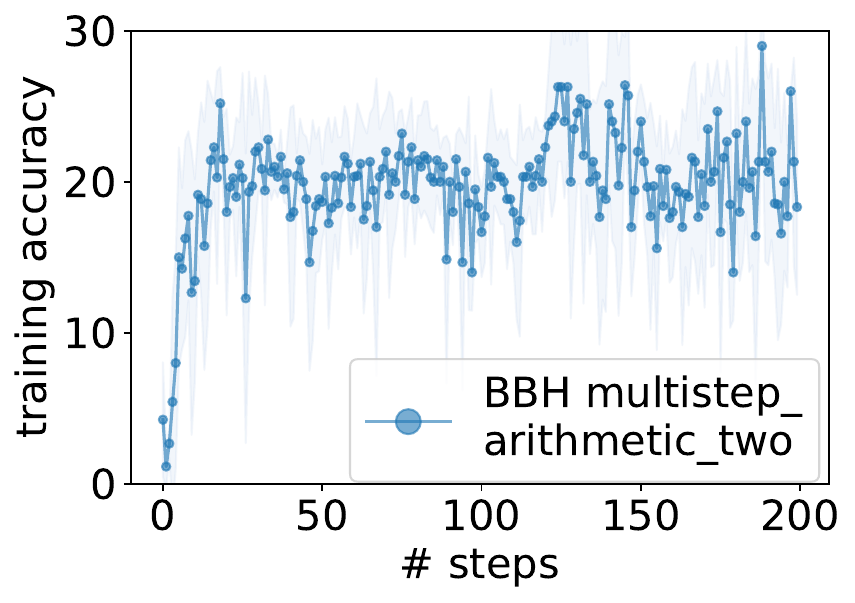}}
\hspace{.01\linewidth}
\subfigure[BBH navigate]{\label{fig:prompt_optimization_graph_bbh_navigate}\includegraphics[width=.31\linewidth]{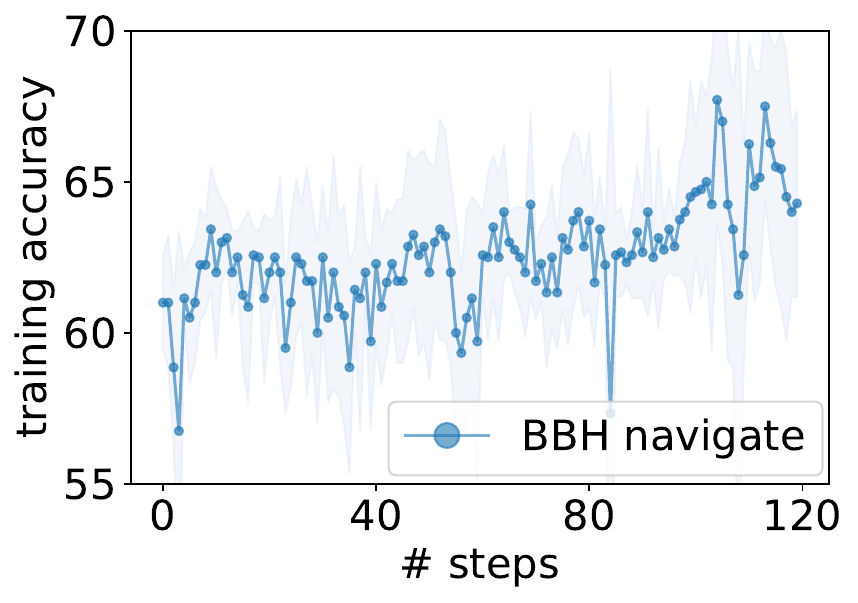}}

\subfigure[BBH object\_counting]{\label{fig:prompt_optimization_graph_bbh_object_counting}\includegraphics[width=.31\linewidth]{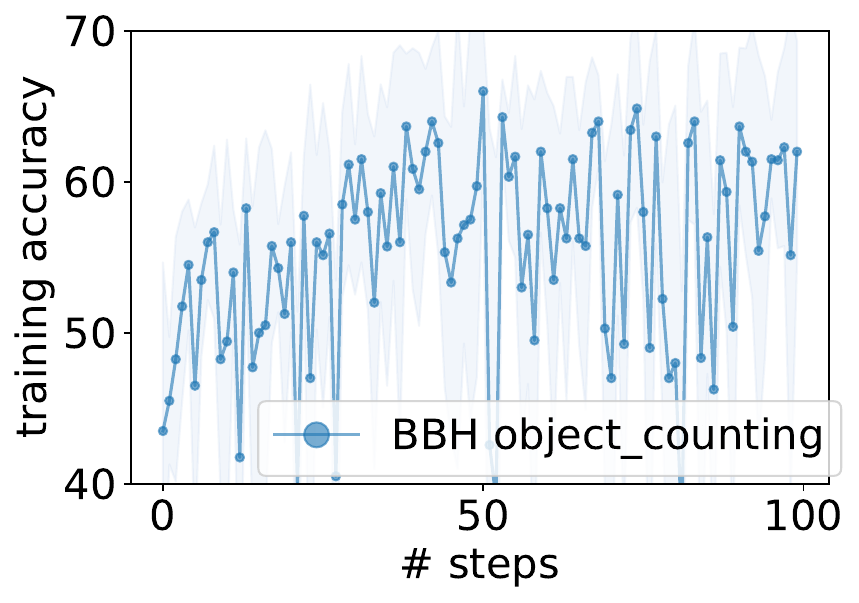}}
\hspace{.01\linewidth}
\subfigure[BBH penguins\_in\_a\_table]{\label{fig:prompt_optimization_graph_bbh_penguins_in_a_table}\includegraphics[width=.31\linewidth]{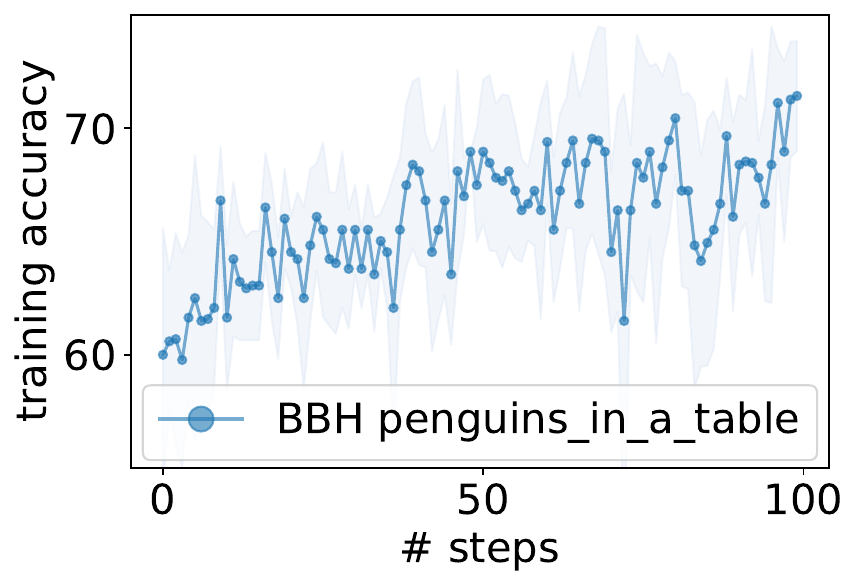}}
\hspace{.01\linewidth}
\subfigure[\scriptsize BBH reasoning\_about\_colored\_objects]{\label{fig:prompt_optimization_graph_bbh_reasoning_about_colored_objects}\includegraphics[width=.31\linewidth]{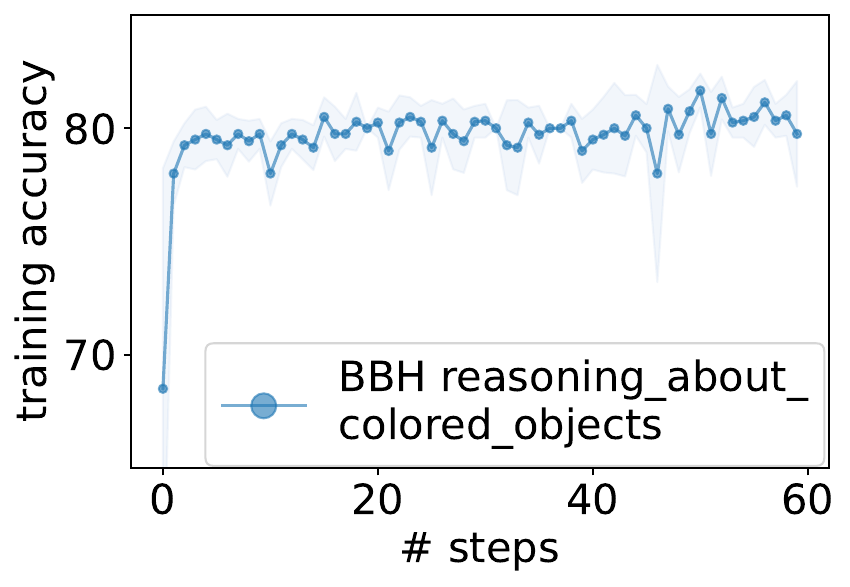}}

\caption{Prompt optimization on 21 BBH tasks (except ruin\_names and temporal\_sequences already shown in Figure~\ref{fig:prompt_optimization_in_main_results_bbh}) with the \texttt{text-bison} scorer and the \texttt{PaLM 2-L-IT} optimizer, Part I.
Most curves have upward trends.
}
\label{fig:prompt_optimization_curves_bbh_text_bison_scorer_all_tasks_appendix_part_one}
\end{figure}

\begin{figure}[H]
\centering
\subfigure[\tiny BBH salient\_translation\_error\_detection]{\label{fig:prompt_optimization_graph_bbh_salient_translation_error_detection}\includegraphics[width=.31\linewidth]{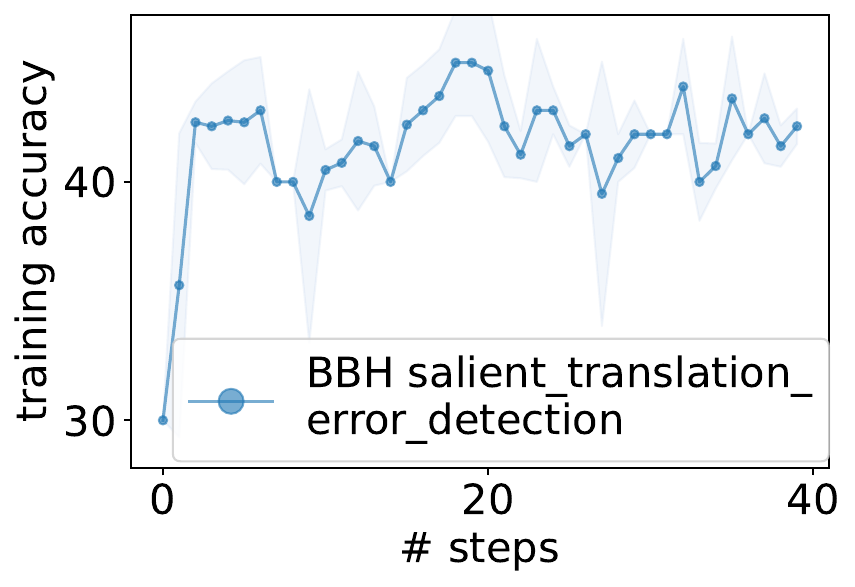}}
\hspace{.01\linewidth}
\subfigure[BBH snarks]{\label{fig:prompt_optimization_graph_bbh_snarks}\includegraphics[width=.31\linewidth]{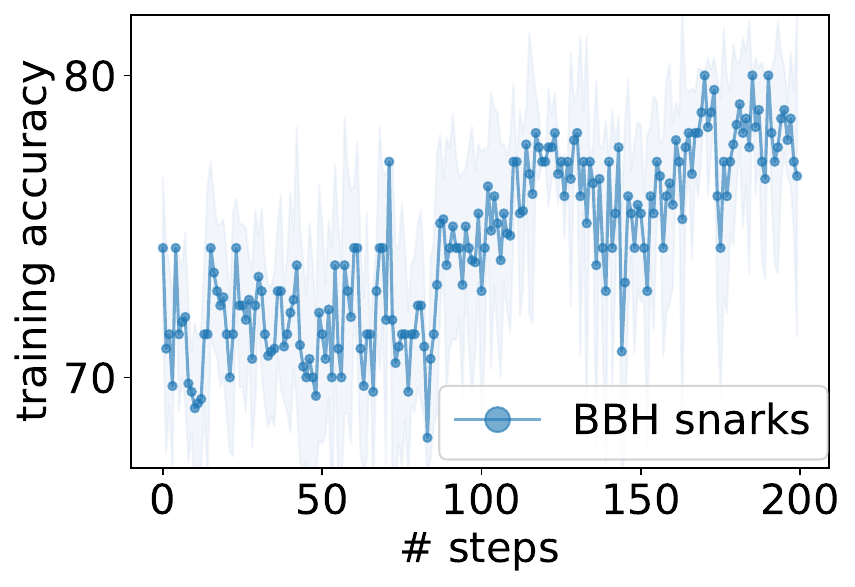}}
\hspace{.01\linewidth}
\subfigure[BBH sports\_understanding]{\label{fig:prompt_optimization_graph_bbh_sports_understanding}\includegraphics[width=.31\linewidth]{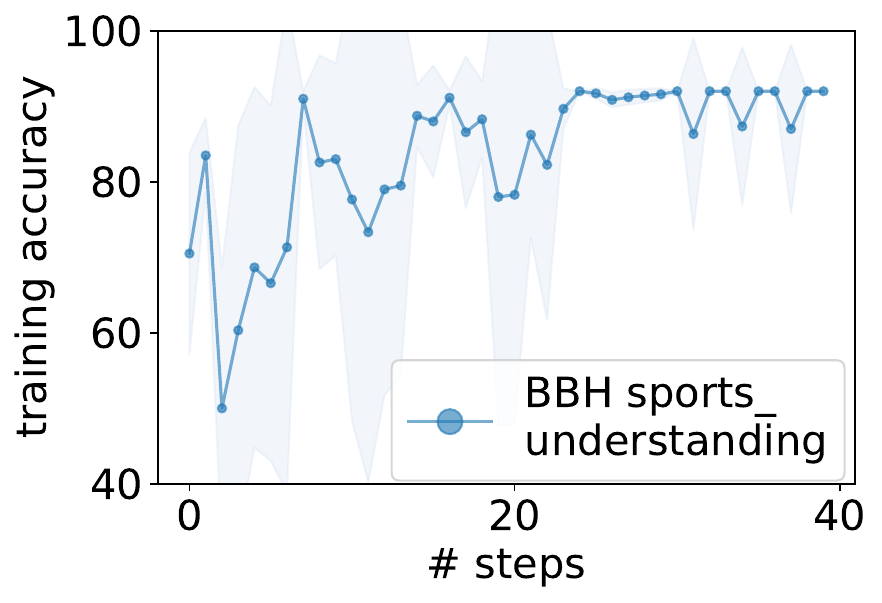}}

\subfigure[BBH tracking\_shuffled\_ \linebreak objects\_seven\_objects]{\label{fig:prompt_optimization_graph_bbh_tracking_shuffled_objects_seven_objects}\includegraphics[width=.31\linewidth]{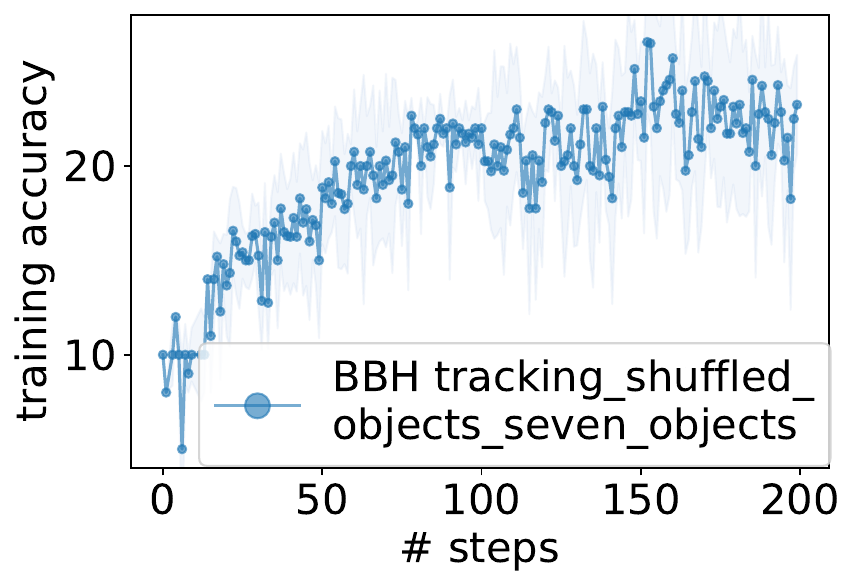}}
\hspace{.01\linewidth}
\subfigure[BBH web\_of\_lies]{\label{fig:prompt_optimization_graph_bbh_web_of_lies}\includegraphics[width=.31\linewidth]{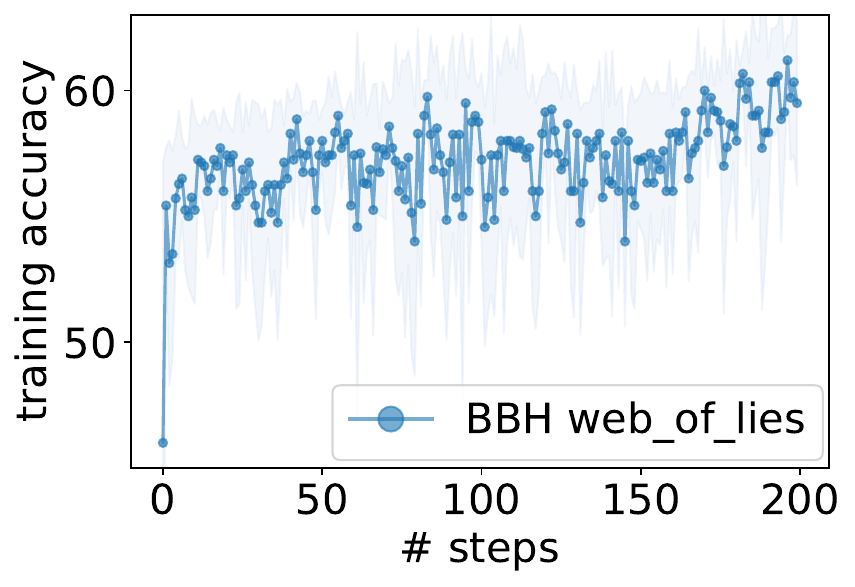}}
\hspace{.01\linewidth}
\subfigure[BBH word\_sorting]{\label{fig:prompt_optimization_graph_bbh_word_sorting}\includegraphics[width=.31\linewidth]{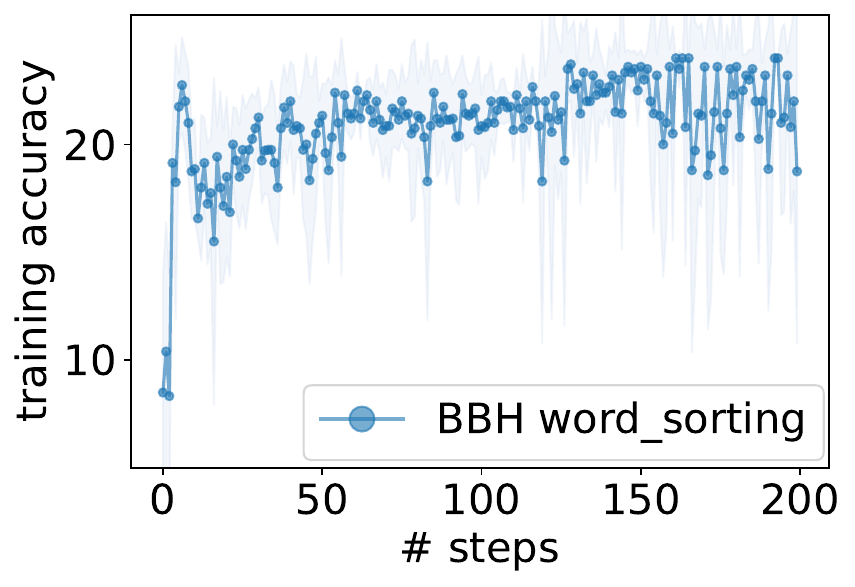}}

\caption{Prompt optimization on 21 BBH tasks (except ruin\_names and temporal\_sequences in Figure~\ref{fig:prompt_optimization_in_main_results_bbh}) with the \texttt{text-bison} scorer and the \texttt{PaLM 2-L-IT} optimizer, Part II.
All curves have upward trends.
}
\label{fig:prompt_optimization_curves_bbh_text_bison_scorer_all_tasks_appendix_part_two}
\end{figure}

\section{Prompt Optimization on BBH Tasks -- Tabulated Accuracies and Found Instructions}
\label{appsec:bbh_taskwise_detailed_results}

\subsection{\texttt{PaLM 2-L-IT} as optimizer, optimization starting from the empty string}
\label{appsec:bbh_taskwise_detailed_results_palm_2_l_it_optimizer}
Table~\ref{table:found_instructions_on_bbh_tasks_palm_2_l} and~\ref{table:found_instructions_on_bbh_tasks_text_bison} show the instructions found by prompt optimization.
A comparison of their accuracies with baselines ``Let's think step by step.''~\citep{kojima2022large}, ``Let’s work this out in a step by step way to be sure we have the right answer.''~\citep{zhou2022large}, and the empty string is in Table~\ref{table:palm2_scores_on_bbh_tasks}; a visualization is in Section~\ref{sec:main_results} Figure~\ref{fig:accuracy_comparison_bar_charts_o_palm_2_l_it}.

\newpage
\begin{table}[H]
\scriptsize
\caption{Accuracies on BBH tasks: our found instructions with the \texttt{PaLM 2-L-IT} optimizer vs baseline.
The optimization starts from the empty string.
Because of the 20-80 train-test split, we show accuracies with the format ``training / test / overall (training + test)''.
The \texttt{PaLM 2-L} scores are from A\_begin instructions; the \texttt{text-bison} scores are from Q\_begin instructions.
Bold numbers indicate the best for the corresponding task.
}
\begin{center}
\scalebox{0.85}{
\begin{tabular}{cP{1.2cm}P{2.2cm}P{2.2cm}P{2.2cm}P{2.2cm}}
\toprule
\multirow{2}{*}{Task} & \multirow{2}{*}{Scorer} & Our Acc & ``Let's think step by step.'' Acc & ``Let’s work this out in a step by step way to be sure we have the right answer.'' Acc & empty string ``'' Acc \\ \cmidrule(lr){3-3} \cmidrule(lr){4-4} \cmidrule(lr){5-5} \cmidrule(lr){6-6}
& & training / test / overall & training / test / overall & training / test / overall & training / test / overall \\
\midrule
boolean\_expressions & \texttt{PaLM 2-L} & \textbf{90.0 / 83.5 / 84.8} & 90.0 / 83.0 / 84.4 & 82.0 / 74.0 / 75.6 & 74.0 / 71.0 / 71.6 \\
causal\_judgement & \texttt{PaLM 2-L} & \textbf{84.8 / 58.0 / 63.1} & 73.0 / 55.3 / 58.8 & 59.5 / 57.3 / 57.8 & 29.7 / 49.3 / 45.5 \\
date\_understanding & \texttt{PaLM 2-L} & \textbf{86.0 / 84.5 / 84.8} & 76.0 / 80.0 / 79.2 & 74.0 / 77.0 / 76.4 & 70.0 / 74.0 / 73.2 \\
disambiguation\_qa & \texttt{PaLM 2-L} & \textbf{80.0 / 69.0 / 71.2} & 40.0 / 52.5 / 50.0 & 48.0 / 47.0 / 47.2 & 54.0 / 57.5 / 56.8 \\
dyck\_languages & \texttt{PaLM 2-L} & \textbf{100.0 / 100.0 / 100.0} & 96.0 / 94.5 / 94.8 & 100.0 / 93.5 / 94.8 & 94.0 / 95.0 / 94.8 \\
formal\_fallacies & \texttt{PaLM 2-L} & \textbf{84.0 / 64.0 / 68.4} & 78.0 / 59.5 / 63.2 & 68.0 / 63.0 / 64.0 & 66.0 / 59.0 / 60.4 \\
geometric\_shapes & \texttt{PaLM 2-L} & \textbf{76.0 / 57.0 / 60.8} & 42.0 / 33.0 / 34.8 & 42.0 / 32.0 / 34.0 & 34.0 / 33.0 / 33.2 \\
hyperbaton & \texttt{PaLM 2-L} & \textbf{100.0 / 96.0 / 96.8} & 78.0 / 75.0 / 75.6 & 74.0 / 72.5 / 72.8 & 88.0 / 89.0 / 88.8 \\
logical\_deduction\_seven\_objects & \texttt{PaLM 2-L} & \textbf{74.0 / 57.0 / 60.4} & 46.0 / 37.0 / 38.8 & 34.0 / 30.5 / 31.2 & 46.0 / 45.5 / 45.6 \\
movie\_recommendation & \texttt{PaLM 2-L} & \textbf{92.0 / 90.5 / 90.8} & 62.0 / 52.5 / 54.4 & 52.0 / 48.0 / 48.8 & 80.0 / 83.0 / 82.4 \\
multistep\_arithmetic\_two & \texttt{PaLM 2-L} & \textbf{72.0 / 55.5 / 58.8} & 42.0 / 46.0 / 45.2 & 60.0 / 50.5 / 52.4 & 4.0 / 3.5 / 3.6 \\
navigate & \texttt{PaLM 2-L} & \textbf{92.0 / 75.0 / 78.4} & 68.0 / 62.0 / 63.2 & 70.0 / 64.0 / 65.2 & 38.0 / 37.5 / 37.6 \\
object\_counting & \texttt{PaLM 2-L} & \textbf{84.0 / 86.5 / 86.0} & 36.0 / 46.5 / 44.4 & 60.0 / 62.0 / 61.6 & 28.0 / 27.0 / 27.2 \\
penguins\_in\_a\_table & \texttt{PaLM 2-L} & \textbf{86.2 / 71.8 / 74.7} & 79.3 / 64.1 / 67.1 & 62.1 / 58.1 / 58.9 & 72.4 / 69.2 / 69.9 \\
reasoning\_about\_colored\_objects & \texttt{PaLM 2-L} & \textbf{98.0 / 85.5 / 88.0} & 82.0 / 79.5 / 80.0 & 82.0 / 75.0 / 76.4 & 42.0 / 35.0 / 36.4 \\
ruin\_names & \texttt{PaLM 2-L} & \textbf{88.0 / 88.0 / 88.0} & 70.0 / 55.0 / 58.0 & 80.0 / 75.5 / 76.4 & 88.0 / 76.5 / 78.8 \\
salient\_translation\_error\_detection & \texttt{PaLM 2-L} & \textbf{62.0 / 67.0 / 66.0} & 42.0 / 50.0 / 48.4 & 58.0 / 46.0 / 48.4 & 56.0 / 56.5 / 56.4 \\
snarks & \texttt{PaLM 2-L} & \textbf{85.7 / 83.2 / 83.7} & 60.0 / 62.2 / 61.8 & 54.3 / 53.1 / 53.4 & 51.4 / 60.1 / 58.4 \\
sports\_understanding & \texttt{PaLM 2-L} & \textbf{98.0 / 88.0 / 90.0} & 50.0 / 46.5 / 47.2 & 60.0 / 52.5 / 54.0 & 52.0 / 41.5 / 43.6 \\
temporal\_sequences & \texttt{PaLM 2-L} & \textbf{100.0 / 100.0 / 100.0} & 100.0 / 96.0 / 96.8 & 90.0 / 87.0 / 87.6 & 100.0 / 99.5 / 99.6 \\
tracking\_shuffled\_objects\_seven\_objects & \texttt{PaLM 2-L} & 32.0 / 16.5 / 19.6 & \textbf{58.0 / 61.5 / 60.8} & 54.0 / 55.5 / 55.2 & 14.0 / 23.5 / 21.6 \\
web\_of\_lies & \texttt{PaLM 2-L} & \textbf{62.0 / 52.0 / 54.0} & 46.0 / 41.5 / 42.4 & 24.0 / 31.0 / 29.6 & \textbf{54.0 / 54.0 / 54.0} \\
word\_sorting & \texttt{PaLM 2-L} & \textbf{54.0 / 54.5 / 54.4} & 2.0 / 4.5 / 4.0 & 12.0 / 9.5 / 10.0 & 20.0 / 22.5 / 22.0 \\
\hdashline\noalign{\vskip 0.5ex}
boolean\_expressions & \texttt{text-bison} & \textbf{98.0 / 87.0 / 89.2} & 72.0 / 61.5 / 63.6 & 88.0 / 78.0 / 80.0 &  80.0 / 68.5 / 70.8 \\
causal\_judgement & \texttt{text-bison} & \textbf{78.4 / 58.0 / 62.0} & 70.3 / 50.7 / 54.5 & 73.0 / 55.3 / 58.8 & \textbf{78.4 / 58.0 / 62.0} \\
date\_understanding & \texttt{text-bison} & \textbf{60.0 / 50.0 / 52.0} & 44.0 / 45.5 / 45.2 & 48.0 / 45.0 / 45.6 &  44.0 / 45.0 / 44.8 \\
disambiguation\_qa & \texttt{text-bison} & \textbf{68.0 / 73.0 / 72.0} & 4.0 / 6.0 / 5.6 & 4.0 / 15.5 / 13.2 & 52.0 / 68.5 / 65.2 \\
dyck\_languages & \texttt{text-bison} & \textbf{100.0 / 100.0 / 100.0} & 100.0 / 95.5 / 96.4 & 100.0 / 94.5 / 95.6 & 100.0 / 98.5 / 98.8 \\
formal\_fallacies & \texttt{text-bison} & 70.0 / 53.0 / 56.4 & 64.0 / 54.5 / 56.4 & \textbf{84.0 / 82.5 / 82.8} & 70.0 / 54.5 / 57.6 \\
geometric\_shapes & \texttt{text-bison} & \textbf{40.0 / 19.5 / 23.6} & 22.0 / 13.0 / 14.8 & 18.0 / 12.0 / 13.2 & 20.0 / 14.5 / 15.6 \\
hyperbaton & \texttt{text-bison} & \textbf{80.0 / 79.5 / 79.6} & 64.0 / 67.5 / 66.8 & 64.0 / 69.0 / 68.0 & 64.0 / 64.0 / 64.0 \\
logical\_deduction\_seven\_objects & \texttt{text-bison} & 66.0 / 53.5 / 56.0 & \textbf{56.0 / 58.0 / 57.6} & 56.0 / 56.0 / 56.0 & 58.0 / 56.5 / 56.8 \\
movie\_recommendation & \texttt{text-bison} & \textbf{98.0 / 90.0 / 91.6} & 68.0 / 63.0 / 64.0 & 66.0 / 62.0 / 62.8 & 68.0 / 64.0 / 64.8 \\
multistep\_arithmetic\_two & \texttt{text-bison} & \textbf{32.0 / 16.5 / 19.6} & 12.0 / 18.0 / 16.8 & 18.0 / 17.5 / 17.6 & 16.0 / 18.5 / 18.0 \\
navigate & \texttt{text-bison} & \textbf{72.0 / 61.0 / 63.2} & 56.0 / 55.0 / 55.2 & 60.0 / 56.5 / 57.2 & 56.0 / 57.0 / 56.8 \\
object\_counting & \texttt{text-bison} & \textbf{72.0 / 62.0 / 64.0} & 58.0 / 57.0 / 57.2 & 62.0 / 55.5 / 56.8 & 50.0 / 57.0 / 55.6 \\
penguins\_in\_a\_table & \texttt{text-bison} & \textbf{72.4 / 56.4 / 59.6} & 58.6 / 53.0 / 54.1 & 55.2 / 55.6 / 55.5 & 58.6 / 53.0 / 54.1 \\
reasoning\_about\_colored\_objects & \texttt{text-bison} & \textbf{82.0 / 77.0 / 78.0} & 76.0 / 72.5 / 73.2 & 78.0 / 73.0 / 74.0 & 74.0 / 69.5 / 70.4 \\
ruin\_names & \texttt{text-bison} & \textbf{88.0 / 82.5 / 83.6} & 66.0 / 65.5 / 65.6 & 66.0 / 62.5 / 63.2 & 64.0 / 66.0 / 65.6 \\
salient\_translation \_error\_detection & \texttt{text-bison} & \textbf{46.0 / 50.5 / 49.6} & 42.0 / 47.5 / 46.4 & 42.0 / 49.5 / 48.0 & 44.0 / 50.0 / 48.8 \\
snarks & \texttt{text-bison} & \textbf{80.0 / 81.8 / 81.5} & 68.6 / 77.6 / 75.8 & 71.4 / 76.2 / 75.3 & 77.1 / 84.6 / 73.1 \\
sports\_understanding & \texttt{text-bison} & \textbf{94.0 / 82.5 / 84.8} & 86.0 / 79.0 / 80.4 & 90.0 / 81.0 / 82.8 & 38.0 / 44.5 / 43.2 \\
temporal\_sequences & \texttt{text-bison} & \textbf{78.0 / 81.0 / 80.4} & 36.0 / 43.5 / 42.0 & 32.0 / 45.0 / 42.4 & 36.0 / 43.0 / 41.6 \\
tracking\_shuffled\_objects\_seven\_objects & \texttt{text-bison} & \textbf{32.0 / 15.5 / 18.8} & 10.0 / 17.0 / 15.6 & 10.0 / 18.0 / 16.4 & 12.0 / 15.5 / 14.8 \\
web\_of\_lies & \texttt{text-bison} & \textbf{62.0 / 50.0 / 52.4} & 48.0 / 45.5 / 46.0 & 48.0 / 44.0 / 44.8 & 52.0 / 51.5 / 51.2 \\
word\_sorting & \texttt{text-bison} & \textbf{24.0 / 17.5 / 18.8} & 10.0 / 12.0 / 11.6 & 4.0 / 8.0 / 7.2 & 4.0 / 7.5 / 6.8 \\
\bottomrule
\end{tabular}
}
\end{center}
\label{table:palm2_scores_on_bbh_tasks}
\end{table}

\newpage
\begin{table}[H]
\scriptsize
\caption{BBH task-wise instructions found by prompt optimization with the \texttt{PaLM 2-L} scorer and the \texttt{PaLM 2-L-IT} optimizer.
The optimization starts from the empty string.
}
\centering
\scalebox{0.99}{
\begin{tabular}{P{2.3cm}P{11.6cm}}
\toprule
Task & Our Instruction \\
\midrule
boolean\_expressions & A Boolean expression is a well-formed expression consisting of variables, values, and logical operators. The expression must evaluate to a single True or False value. The order of precedence of the logical operators is as follows: NOT, AND, OR, XOR, IMP. Parentheses can be used to group subexpressions and to control the order of evaluation.\\ [2ex]
causal\_judgement & When considering questions about causation, a typical person would consider the following factors: whether the action or event was a necessary condition for the outcome to occur, a sufficient condition, a proximate cause, or a foreseeable cause. \\ [2ex]
date\_understanding & To find the date X time ago from today, first find today's date. Then subtract X time from today's date. If the current date is the last day of a month, then the date a month ago is the last day of the previous month. If the current date is not the last day of a month, then the date a month ago is the same day of the previous month. For example, if today is March 31, 2023, then the date a month ago is February 28, 2023. If today is April 1, 2023, then the date a month ago is March 1, 2023.\\ [2ex]
disambiguation\_qa & Identifying Antecedents of Pronouns: A Comprehensive Guide \\ [2ex]
dyck\_languages & First, look for the opening parentheses. Then, count the number of opening parentheses. Finally, close the parentheses in the reverse order that they were opened.\\ [2ex]
formal\_fallacies & A deductive argument is one where the conclusion follows necessarily from the premises. If the premises are true, then the conclusion must also be true. An invalid argument is one where it is possible for the premises to be true and the conclusion to be false.\\ [2ex]
geometric\_shapes & A closed polygonal chain is a series of connected line segments. The line segments can be straight or curved. The first and last line segments are connected. The line segments do not intersect each other except at their endpoints. A closed polygon can be described by an SVG path element, which starts at a given point, goes to one or more additional points, and then ends at the starting point. The path element can consist of straight line segments, curved segments, or a mixture of both.\\ [2ex]
hyperbaton & The correct adjective order in English is opinion, size, shape, age, color, origin, material, and purpose. If you have more than one adjective of the same type, they are usually placed in order of importance. For example, you would say "a large, old, Pakistani ship" rather than "an old, large, Pakistani ship." There are a few exceptions to these rules, but they are generally followed in most cases.\\ [2ex]
logical\_deduction \_seven\_objects & The following questions will test your ability to use deductive reasoning. You will be given a set of statements about a group of objects. You will then be asked to answer questions about the objects based on the statements. The statements in the questions are logically consistent, so you can use them to deduce the order of the objects. For each question, you must choose the option that is logically consistent with the information in the questions.\\ [2ex]
movie\_recommendation & Based on your input, I have analyzed the given movies in terms of genre, plot, tone, audience rating, year of release, director, cast, and reviews. I have also taken into account the given options. The movie that is most similar to the given movies in terms of all these factors is:\\ [2ex]
multistep\_arithmetic \_two & The order of operations in mathematics is PEMDAS, which stands for Parentheses, Exponents, Multiplication, Division, Addition, and Subtraction. When there are multiple operations of the same precedence, they must be performed from left to right. Note that multiplication and division have the same precedence, as do addition and subtraction.\\ [2ex]
navigate & You will return to the starting point if and only if (1) the total number of steps you take forward is equal to the total number of steps you take back, and (2) the total number of turns you make is a multiple of 180 degrees.\\ [2ex]
object\_counting & Here is a list of the objects you mentioned and their corresponding counts:\\ [2ex]
penguins\_in\_a\_table & Here is my new text:\\ [2ex]
reasoning\_about \_colored\_objects & Starting from the leftmost object in the row, I observe the following objects arranged in this order:\\ [4ex]
ruin\_names & Which is the funniest pun on the artist or movie name?\\ [2ex]
salient\_translation \_error\_detection & Instructions: Read the German sentence and its English translation carefully, then identify the type of error in the translation and select the correct option. There are six possible types of errors: Named Entities, Numerical Values, Modifiers or Adjectives, Negation or Antonyms, Facts, and Dropped Content.\\ [2ex]
snarks & Identify the sarcastic statement by considering the following factors: incongruity, exaggeration, understatement, context, speaker's intent, and audience's reaction. I will also consider the speaker's tone of voice, facial expressions, and body language.\\ [2ex]
sports\_understanding & I will determine if a sentence about an athlete is plausible by first checking if it is grammatically correct. If it is, I will then check if it is consistent with the athlete's sport, position, and real-world statistics. I will also check if it is consistent with the rules of the athlete's sport. If the sentence is consistent with all of these things, I will answer "yes", otherwise I will answer "no".\\ [2ex]
temporal\_sequences & The answer is the time that is not mentioned in the given statements. \\ [2ex]
tracking\_shuffled\_objects \_seven\_objects & Claire has the blue ball, Gertrude has the black ball, and Dave has the green ball. They are all happy with their new balls. \\ [2ex]
web\_of\_lies & The answer to a question is yes if there are an odd number of liars before the current speaker, and no if there are an even number of liars before the current speaker. If the current speaker is a truth-teller, they will say the opposite of what the previous person said, while a liar will say the same thing as the previous person said. \\ [2ex]
word\_sorting & Alphabetical order of given words: \\
\bottomrule
\end{tabular}
}
\label{table:found_instructions_on_bbh_tasks_palm_2_l}
\end{table}

\newpage
\begin{table}[H]
\scriptsize
\caption{BBH task-wise instructions found by prompt optimization with the \texttt{text-bison} scorer and the \texttt{PaLM 2-L-IT} optimizer.
The optimization starts from the empty string.
}
\begin{center}
\begin{tabular}{P{2.3cm}P{11.6cm}}
\toprule
Task& Our Instruction \\
\midrule
boolean\_expressions & Not (not False) and not not False is False \\ [2ex]
causal\_judgement & A typical person would likely answer the questions about causation as follows: \\ [2ex]
date\_understanding & Today is February 28, 2023. It is a Tuesday. Yesterday was Monday, February 27, 2023. Tomorrow will be Wednesday, March 1, 2023. A week ago, it was February 21, 2023, and a month ago, it was January 28, 2023. A year from now, it will be February 28, 2024. The day of the week is important to note because it will help us to correctly answer the questions below. Not all years are leap years that contain February 29. \\ [2ex]
disambiguation\_qa & A pronoun is a word that stands in for a noun. The noun that a pronoun refers to is called its antecedent. To identify the antecedent of a pronoun, look for the noun that the pronoun could be referring to. If there is only one possible noun, then that is the antecedent. If there are two or more possible nouns, then the antecedent is ambiguous. Use the context of the sentence to help you determine the correct antecedent. \\ [2ex]
dyck\_languages & \{ \} \\ [2ex]
formal\_fallacies & How to Evaluate Deductive Validity of an Argument \\ [2ex]
geometric\_shapes & What shape is this SVG code drawing, and how many sides does it have? \\ [2ex]
hyperbaton & In English, adjectives are typically placed before nouns in a specific order. The order is: opinion, size, shape, age, color, origin, material, purpose, noun. For example, the sentence "the big, old, red barn" would be considered grammatically correct, while the sentence "the old, big, red barn" would not. Adjectives that come before nouns are called attributive adjectives, while adjectives that come after nouns are called predicative adjectives. \\ [2ex]
logical\_deduction \_seven\_objects & In this logical reasoning task, you will be given a series of paragraphs, each of which describes a set of objects arranged in a fixed order. The statements in each paragraph are logically consistent. You must read each paragraph carefully and use the information given to determine the logical relationships between the objects. You will then be asked a question about the order of the objects. Read each question carefully and choose the option that answers the question correctly. \\ [2ex]
movie\_recommendation & What is the highest-rated movie similar to the given movies, with a similar IMDb rating and released in the same year? \\ [2ex]
multistep\_arithmetic\_two & Let's solve these equations using PEMDAS order of operations. Remember that PEMDAS stands for parentheses, exponents, multiplication and division, and addition and subtraction. \\ [2ex]
navigate & Starting at the origin, facing north, follow the instructions. If your displacement from the origin is zero and your direction is unchanged, then your answer is Yes. Otherwise, your answer is No. \\ [2ex]
object\_counting & Let me help you count the items you have. Just list them one by one, separated by commas. I will then count each item and tell you how many items there are in total. \\ [2ex]
penguins\_in\_a\_table & This table shows information about penguins. The columns show the penguin’s name, age, height (in cm), and weight (in kg). The penguins are listed in order of their age, from youngest to oldest. \\ [2ex]
reasoning\_about \_colored\_objects & First, read the input carefully. Then, identify all the objects mentioned, their colors, and their positions. Next, visualize the objects and their positions in your mind. Finally, answer the questions accurately based on the information given. Make sure to pay attention to the order of the objects. \\ [2ex]
ruin\_names & A humorous edit of an artist or movie name can be created by replacing one or more letters to form a new word or phrase that sounds similar but has a different meaning. The new word or phrase should be relevant to the original word, but it should also be a surprise, which makes the edit funny. For example, the artist or movie name "Rocky" can be changed to "Ricky," and "Schindler's List" can be changed to "Schindler's Lift." Be creative and have fun! \\ [2ex]
salient\_translation \_error\_detection & The following translations from German to English contain a particular error. The error may be one of the following types: Named Entities, Numerical Values, Modifiers or Adjectives, Negation or Antonyms, Facts, or Dropped Content. Please identify the error. \\ [2ex]
snarks & The statement \\ [2ex]
sports\_understanding & To determine the plausibility of a sports sentence, I will first identify the sport, athletes, teams, and events mentioned in the sentence. Then, I will use my knowledge of the rules of the sport, the context of the sentence, common sense, and my knowledge of the world to determine whether the sentence is plausible. I will also consider the time period and location, as well as any other relevant information. Finally, I will return a score of 1 for plausible sentences and 0 for implausible ones. \\ [2ex]
temporal\_sequences & To determine the time period when a person went to a place, first identify all the time periods when the person's whereabouts are unknown. Then, rule out any time periods during which the person was seen doing something else or the place was closed. The remaining time periods are the possible times when the person could have gone to the place. \\ [2ex]
tracking\_shuffled\_objects \_seven\_objects & At the start of the game, Claire has a blue ball. Throughout the game, pairs of people swap balls. Claire ends up with the yellow ball. \\ [2ex]
web\_of\_lies & People in a group either tell the truth or lie. The truthfulness of a person's statement is determined by the statement of the previous person. If the previous person told the truth, then the current person who says the opposite is lying. If the previous person lied, then the current person who says the opposite is telling the truth. This rule applies to all subsequent statements. \\ [2ex]
word\_sorting & Sort the following words alphabetically, ignoring case and punctuation. Print the sorted list. \\
\bottomrule
\end{tabular}
\end{center}
\label{table:found_instructions_on_bbh_tasks_text_bison}
\end{table}

\newpage
\subsection{\texttt{gpt-3.5-turbo} as optimizer, optimization starting from the empty string}
\label{appsec:bbh_taskwise_detailed_results_gpt_3.5_turbo_optimizer_start_from_empty}

\begin{figure}[H]
\centering
\scalebox{0.9}{
\subfigure[\texttt{PaLM 2-L}, ours minus ``Let's think step by step.'']{\label{fig:accuracy_comparison_palm_2_l_found_minus_step_by_step_o_gpt_3.5_turbo}\includegraphics[width=.48\linewidth]{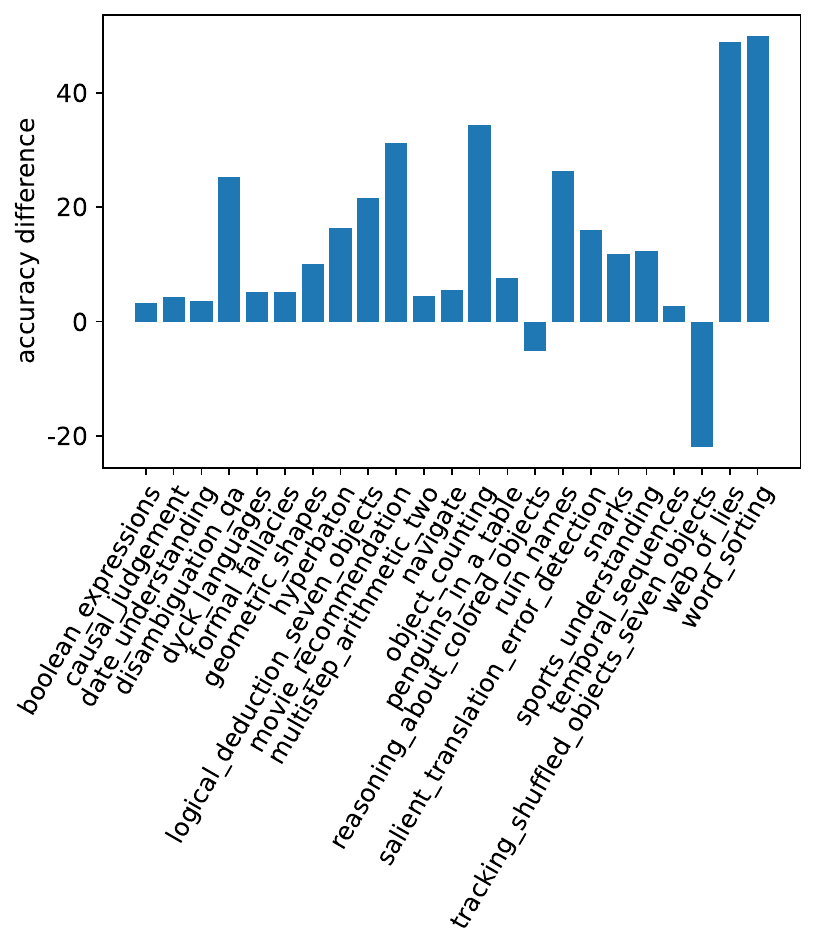}}
\hspace{.01\linewidth}
\subfigure[\texttt{PaLM 2-L}, ours minus empty starting point]{\label{fig:accuracy_comparison_palm_2_l_found_minus_empty_o_gpt_3.5_turbo}\includegraphics[width=.48\linewidth]{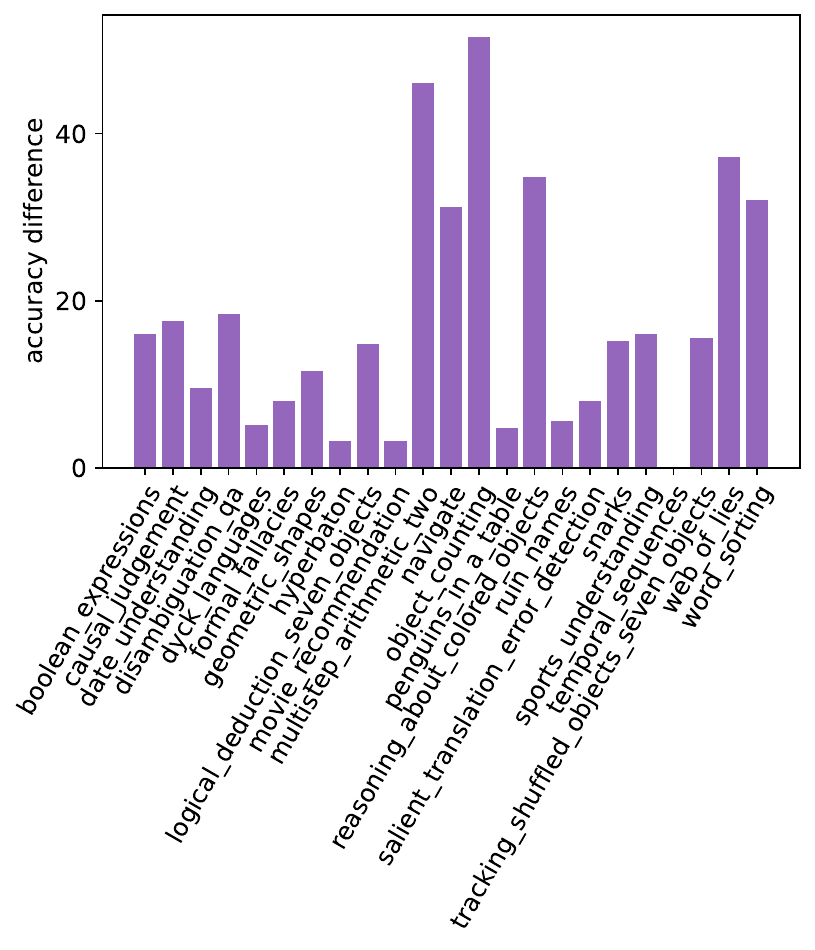}}}

\scalebox{0.91}{
\subfigure[\scriptsize \texttt{text-bison}, ours minus ``Let's think step by step.'']{\label{fig:accuracy_comparison_text_bison_found_minus_step_by_step_o_gpt_3.5_turbo}\includegraphics[width=.48\linewidth]{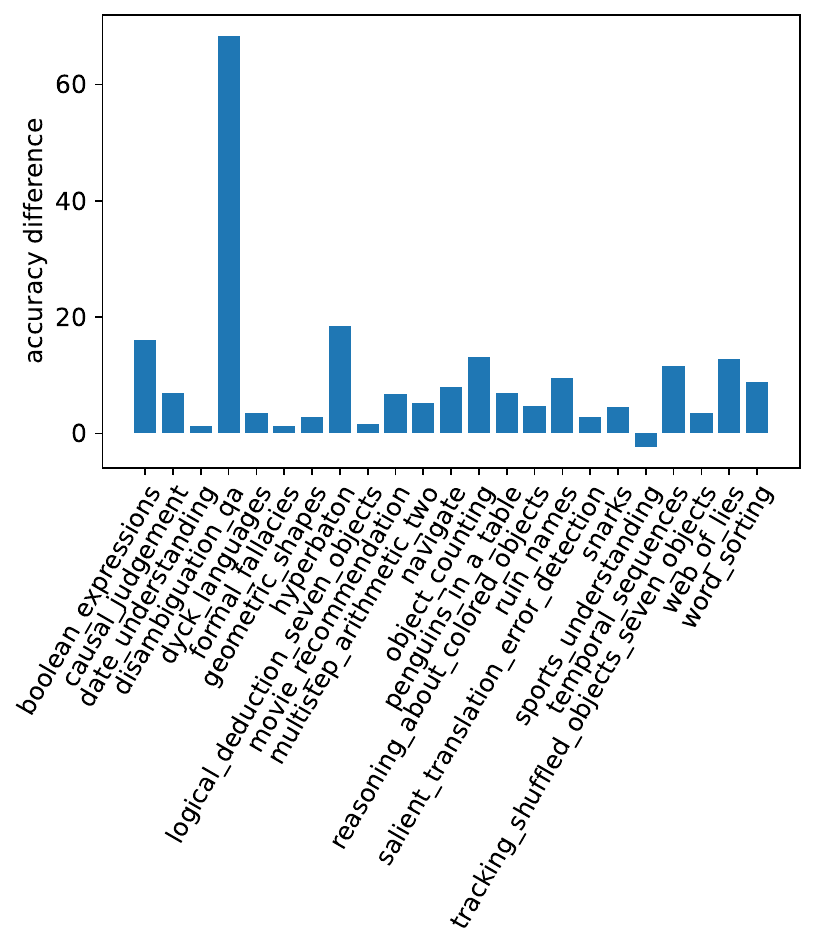}}
\hspace{.01\linewidth}
\subfigure[\texttt{text-bison}, ours minus empty starting point]{\label{fig:accuracy_comparison_text_bison_found_minus_empty_o_gpt_3.5_turbo}\includegraphics[width=.48\linewidth]{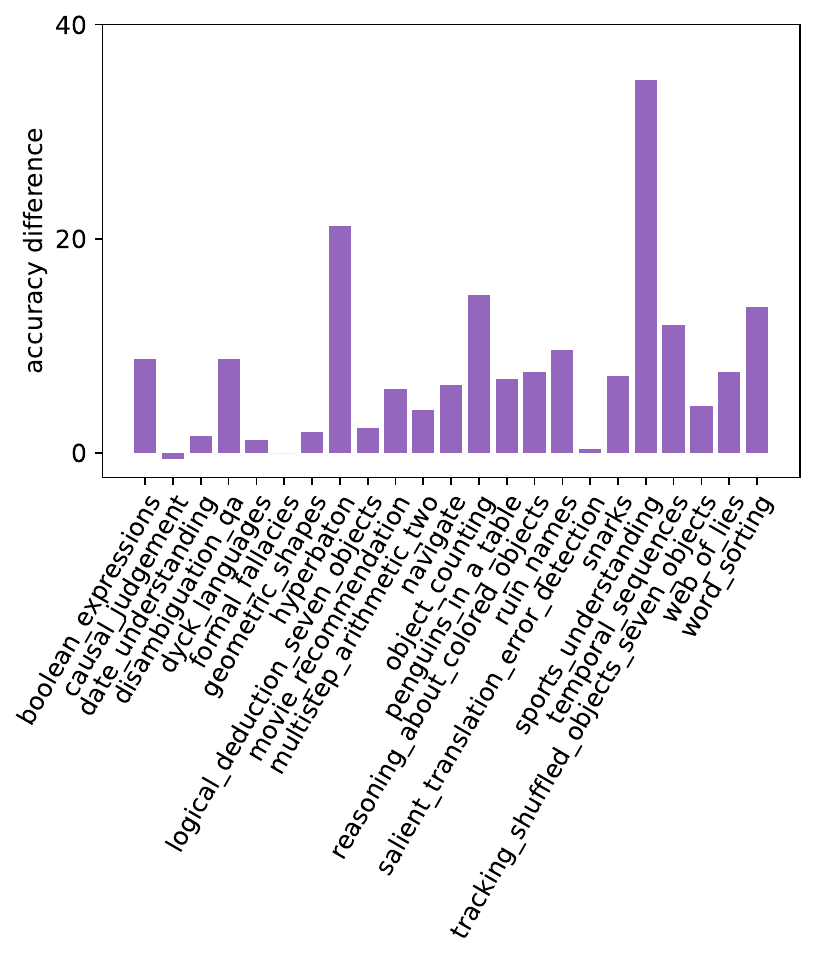}}
}
\caption{On 23 BBH tasks, the accuracy differences among instructions found by prompt optimization (with the \texttt{gpt-3.5-turbo} optimizer), ``Let's think step by step.'', and the empty string (optimization starting point).
}
\label{fig:accuracy_comparison_bar_charts_o_gpt_3.5_turbo}
\end{figure}

Table~\ref{table:found_instructions_on_bbh_tasks_s_palm_2_l_o_gpt_3.5_turbo_from_empty}, \ref{table:found_instructions_on_bbh_tasks_s_text_bison_o_gpt_3.5_turbo_from_empty} and~\ref{table:found_instructions_on_bbh_tasks_s_text_bison_o_gpt_3.5_turbo_from_empty_q_end} show the instructions found by prompt optimization.
Their accuracies are listed in Table~\ref{table:gpt_scores_on_bbh_tasks_starting_from_empty}.
Figure~\ref{fig:accuracy_comparison_bar_charts_o_gpt_3.5_turbo} visualizes the difference between their accuracies and those of the baselines ``Let's think step by step.'' and the empty string.
The optimizations find instructions better than the empty starting point, and most of the found instructions are better than ``Let's think step by step''.

One caveat in the A\_begin instructions (Table~\ref{table:found_instructions_on_bbh_tasks_s_palm_2_l_o_gpt_3.5_turbo_from_empty}) is that a lot of the found instructions are imperative or interrogative sentences that are more suitable to be put into ``Q:'' rather than ``A:'', like ``Solve the sequence by properly closing the parentheses.'' for dyck\_languages and ``Which movie option from the given choices ...?'' for movie\_recommendation.
Such styles appear more often here than the \texttt{PaLM 2-L-IT} optimizer results (Table~\ref{table:found_instructions_on_bbh_tasks_palm_2_l}), showing \texttt{PaLM 2-L-IT} understands the needed style better.
In Section~\ref{appsec:bbh_taskwise_detailed_results_gpt_3.5_turbo_optimizer_start_from_solve}, we show the A\_begin optimization results with the non-empty starting point ``Let's solve the problem.''.
Most results there are declarative sentences -- more suitable for A\_begin.

\newpage
\begin{table}[H]
\scriptsize
\caption{Accuracies on BBH tasks with the \texttt{gpt-3.5-turbo} optimizer that starts from the empty string.
The \texttt{PaLM 2-L} scores are from A\_begin (left) instructions; the \texttt{text-bison} scores include Q\_begin (left) and Q\_end (right) instructions.
}
\begin{center}
\begin{tabular}{cP{1.2cm}P{2.2cm}P{2.2cm}}
\toprule
\multirow{2}{*}{Task} & \multirow{2}{*}{Scorer} & Our Acc (\texttt{begin}) & Our Acc (\texttt{end})\\ \cmidrule(lr){3-3} \cmidrule(lr){4-4}
& & training / test / overall & training / test / overall \\
\midrule
boolean\_expressions & \texttt{PaLM 2-L} & 92.0 / 86.5 / 87.6 & N/A \\
causal\_judgement & \texttt{PaLM 2-L} & 81.1 / 58.7 / 63.1 & N/A \\
date\_understanding & \texttt{PaLM 2-L} & 86.0 / 82.0 / 82.8 & N/A \\
disambiguation\_qa & \texttt{PaLM 2-L} & 80.0 / 74.0 / 75.2 & N/A \\
dyck\_languages & \texttt{PaLM 2-L} & 100.0 / 100.0 / 100.0 & N/A \\
formal\_fallacies & \texttt{PaLM 2-L} & 88.0 / 63.5 / 68.4 & N/A \\
geometric\_shapes & \texttt{PaLM 2-L} & 60.0 / 41.0 / 44.8 & N/A \\
hyperbaton & \texttt{PaLM 2-L} & 88.0 / 93.0 / 92.0 & N/A \\
logical\_deduction\_seven\_objects & \texttt{PaLM 2-L} & 76.0 / 56.5 / 60.4 & N/A \\
movie\_recommendation & \texttt{PaLM 2-L} & 84.0 / 86.0 / 85.6 & N/A \\
multistep\_arithmetic\_two & \texttt{PaLM 2-L} & 52.0 / 49.0 / 49.6 & N/A \\
navigate & \texttt{PaLM 2-L} & 76.0 / 67.0 / 68.8 & N/A \\
object\_counting & \texttt{PaLM 2-L} & 78.0 / 79.0 / 78.8 & N/A \\
penguins\_in\_a\_table & \texttt{PaLM 2-L} & 82.8 / 72.6 / 74.7 & N/A \\
reasoning\_about \_colored\_objects & \texttt{PaLM 2-L} & 86.0 / 67.5 / 71.2 & N/A \\
ruin\_names & \texttt{PaLM 2-L} & 90.0 / 83.0 / 84.4 & N/A \\
salient\_translation\_error\_detection & \texttt{PaLM 2-L} & 62.0 / 65.0 / 64.4 & N/A \\
snarks & \texttt{PaLM 2-L} & 85.7 / 70.6 / 73.6 & N/A \\
sports\_understanding & \texttt{PaLM 2-L} & 68.0 / 57.5 / 59.6 & N/A \\
temporal\_sequences & \texttt{PaLM 2-L} & 100.0 / 99.5 / 99.6 & N/A \\
tracking\_shuffled\_objects\_seven\_objects & \texttt{PaLM 2-L} & 44.0 / 34.5 / 36.4 & N/A \\
web\_of\_lies & \texttt{PaLM 2-L} & 92.0 / 91.0 / 91.2 & N/A \\
word\_sorting & \texttt{PaLM 2-L} & 62.0 / 52.0 / 54.0 & N/A \\
\hdashline\noalign{\vskip 0.5ex}
boolean\_expressions & \texttt{text-bison} & 84.0 / 78.5 / 79.6 & 80.0 / 78.0 / 78.4\\
causal\_judgement & \texttt{text-bison} & 78.4 / 57.3 / 61.5 & 83.8 / 53.3 / 59.4\\
date\_understanding & \texttt{text-bison} & 52.0 / 45.0 / 46.4 & 64.0 / 52.4 / 54.8\\
disambiguation\_qa & \texttt{text-bison} & 68.0 / 75.5 / 74.0 & 64.0 / 71.5 / 70.0\\
dyck\_languages & \texttt{text-bison} & 100.0 / 99.5 / 99.6 & 100.0 / 100.0 / 100.0\\
formal\_fallacies & \texttt{text-bison} & 70.0 / 54.5 / 57.6 & 74.0 / 53.5 / 57.6\\
geometric\_shapes & \texttt{text-bison} & 28.0 / 15.0 / 17.6 & 48.0 / 28.0 / 32.0\\
hyperbaton & \texttt{text-bison} & 86.0 / 85.0 / 85.2 & 80.0 / 76.5 / 77.2 \\
logical\_deduction\_seven\_objects & \texttt{text-bison} & 66.0 / 57.5 / 59.2 & 62.0 / 55.0 / 56.4\\
movie\_recommendation & \texttt{text-bison} & 76.0 / 69.5 / 70.8 & 82.0 / 70.5 / 72.8\\
multistep\_arithmetic\_two & \texttt{text-bison} & 28.0 / 20.5 / 22.0 & 28.0 / 22.5 / 23.6\\
navigate & \texttt{text-bison} & 72.0 / 61.0 / 63.2 & 68.0 / 59.5 / 61.2\\
object\_counting & \texttt{text-bison} & 68.0 / 71.0 / 70.4 & 72.0 / 69.0 / 69.6 \\
penguins\_in\_a\_table & \texttt{text-bison} & 65.5 / 59.8 / 61.0 & 79.3 / 53.0 / 58.2 \\
reasoning\_about\_colored\_objects & \texttt{text-bison} & 84.0 / 76.5 / 78.0 & 86.0 / 74.0 / 76.4\\
ruin\_names & \texttt{text-bison} & 80.0 / 74.0 / 75.2 & 74.0 / 75.0 / 74.8\\
salient\_translation\_error\_detection & \texttt{text-bison} & 44.0 / 50.5 / 49.2 & 48.0 / 51.0 / 50.4\\
snarks & \texttt{text-bison} & 82.9 / 79.7 / 80.3 & 88.6 / 84.6 / 85.4 \\
sports\_understanding & \texttt{text-bison} & 84.0 / 76.5 / 78.0 & 90.0 / 80.0 / 82.0 \\
temporal\_sequences & \texttt{text-bison} & 50.0 / 54.5 / 53.6 & 64.0 / 61.5 / 62.0\\
tracking\_shuffled\_objects\_seven\_objects & \texttt{text-bison} & 22.0 / 18.5 / 19.2 & 30.0 / 21.5 / 23.2 \\
web\_of\_lies & \texttt{text-bison} & 64.0 / 57.5 / 58.8 & 68.0 / 55.0 / 57.6\\
word\_sorting & \texttt{text-bison} & 26.0 / 19.0 / 20.4 & 32.0 / 25.5 / 26.8 \\
\bottomrule
\end{tabular}
\end{center}
\label{table:gpt_scores_on_bbh_tasks_starting_from_empty}
\end{table}

\newpage
\begin{table}[H]
\scriptsize
\caption{BBH task-wise instructions found by prompt optimization with the \texttt{PaLM 2-L} scorer and the \texttt{gpt-3.5-turbo} optimizer.
The optimizations start from the empty string.
}
\centering
\begin{tabular}{P{2.3cm}P{11.6cm}}
\toprule
Task & Our Instruction \\
\midrule
boolean\_expressions & An accurate evaluation of logical expressions involves correctly applying Boolean operators, considering the order of operations, and analyzing the truth values of the operands in accordance with Boolean logic principles. \\ [2ex]
causal\_judgement & Understanding causality is critical for accurately assessing cause and effect relationships in various scenarios, leading to well-informed judgments, precise conclusions, and definitive answers to questions about the outcomes involved. \\ [2ex]
date\_understanding & What is the specific date mentioned or required in each given problem or question, taking into account all relevant information, available options, and the provided context? Please provide the accurate answer in the format MM/DD/YYYY. \\ [2ex]
disambiguation\_qa & Accurately analyze and clarify the pronoun-antecedent relationship in the given sentences, identifying the appropriate referent to eliminate any potential confusion or ambiguity and ensure a precise understanding of the intended meaning. \\ [2ex]
dyck\_languages & Solve the sequence by properly closing the parentheses. \\ [2ex]
formal\_fallacies & In determining the deductive validity of arguments based on explicit premises, a meticulous analysis of the logical relationships and implications is essential for definitively establishing their soundness, confirming their validity or invalidity, and ensuring a reliable and robust assessment of the arguments at hand. \\ [2ex]
geometric\_shapes & The SVG path element with the "d" attribute plays a crucial role in web development, allowing for the precise definition and rendering of various shapes on a webpage. \\ [2ex]
hyperbaton & Understanding the correct order of adjectives is crucial for constructing grammatically accurate and coherent sentences that effectively convey the intended meaning in diverse contexts while ensuring clarity, cohesion, and consistency throughout consistently and effortlessly. \\ [2ex]
logical\_deduction \_seven\_objects & By conducting a meticulous analysis of the given information and ensuring logical consistency within each paragraph, we can accurately determine the precise order or ranking of the mentioned objects, allowing us to confidently and consistently identify the correct answer in every presented scenario with utmost precision and confidence. \\ [2ex]
movie\_recommendation & Which movie option from the given choices closely matches the mentioned films in terms of themes, storylines, and characteristics, guaranteeing the highest possible similarity score among them all? \\ [2ex]
multistep\_arithmetic\_two & Evaluate the given mathematical expressions step by step to determine the correct solutions accurately. \\ [2ex]
navigate & Is it possible to determine, with absolute certainty, whether strictly adhering to the given instructions will unfailingly bring you back to the original starting point without any exceptions, errors, or deviations? \\ [2ex]
object\_counting & Determine the total number of objects or entities mentioned in the given list, covering various categories and types, to accurately calculate the overall count. \\ [2ex]
penguins\_in\_a\_table & From the given table, what information can we gather about the mentioned animals and their respective attributes, including names, ages, heights, and weights? \\ [2ex]
reasoning\_about \_colored\_objects & By thoroughly examining the given information, accurately determine the answers for each question by considering the specific characteristics, colors, and positions of the mentioned objects. \\ [2ex]
ruin\_names & Select the most amusing and clever alteration from the options provided for the given artist, movie, or title name, and accurately choose the correct answer to test your wit and creativity. \\ [2ex]
salient\_translation \_error\_detection & Thoroughly examine the given translations from German to English and accurately identify any errors by carefully analyzing the text and selecting the appropriate option with meticulous attention to detail, precision, utmost accuracy, and comprehensive understanding of the language for precise evaluation and categorization. \\ [2ex]
snarks & Which option delivers the most devastatingly sarcastic response, brilliantly exposing the sheer absurdity and leaving absolutely no doubt whatsoever in all the given situations? \\ [2ex]
sports\_understanding & Maintaining the accuracy, reliability, and integrity of sports event representation is essential for upholding the highest standards of credibility, trustworthiness, and overall quality in conveying information, without any compromise, misrepresentation, or distortion, thereby ensuring the factual accuracy of sports journalism. \\ [2ex]
temporal\_sequences & Based on the provided timeline and observed activities, we can accurately determine the possible time range when each individual could have visited their intended destinations and answer questions about their visitation time. \\ [2ex]
tracking\_shuffled\_objects \_seven\_objects & An important point to note is that each person in the group starts with one specific book at the beginning of the semester. \\ [2ex]
web\_of\_lies & Analyzing the consistency and accuracy of statements provided by each person is crucial for determining the truthfulness of individuals in every scenario. \\ [2ex]
word\_sorting & Please sort the given words in alphabetical order: The list of words to be sorted contains - \\ [2ex]

\bottomrule
\end{tabular}
\label{table:found_instructions_on_bbh_tasks_s_palm_2_l_o_gpt_3.5_turbo_from_empty}
\end{table}

\newpage
\begin{table}[H]
\scriptsize
\caption{BBH task-wise Q\_begin instructions found by prompt optimization with the \texttt{text-bison} scorer and the \texttt{gpt-3.5-turbo} optimizer.
The optimizations start from the empty string.
}
\begin{center}
\begin{tabular}{P{2.3cm}P{11.6cm}}
\toprule
Task & Our Instruction \\
\midrule
boolean\_expressions & Group sub-expressions with parentheses to accurately evaluate logical operations: not, and, and finally or. Determine the resulting value as either True or False. \\ [2ex]
causal\_judgement & Consider the intentions and actions of the individuals involved. \\ [2ex]
date\_understanding & Determine the one-day difference in the given date and express it in the format MM/DD/YYYY. \\ [2ex]
disambiguation\_qa & Determine the precise antecedent of the pronoun in the given sentence and select the correct option or state if it is ambiguous. \\ [2ex]
dyck\_languages & Ensure that all opening brackets have a corresponding closing bracket, and that the closing brackets are in the correct order. \\ [2ex]
formal\_fallacies & Thoroughly analyze the explicitly provided premises and determine the deductive validity of the argument based on all necessary conditions, implications, exclusions, and dependencies given. \\ [2ex]
geometric\_shapes & Analyze the given SVG path element carefully and confidently select the correct option from the provided choices to accurately determine the corresponding shape. Pay close attention to the specific path details and confidently make the most suitable choice. \\ [2ex]
hyperbaton & Select the sentence that strictly adheres to the standard order of adjectives: opinion, size, age, shape, color, origin, material, and purpose. Ensure there are no deviations or alterations in the adjective order. Choose the option without any changes. \\ [2ex]
logical\_deduction \_seven\_objects & Analyze the given information to accurately determine the precise order and ranking of the mentioned objects/people, considering their relationships, positions, and any provided comparisons, for a definitive and logical progression with maximum accuracy and efficiency. \\ [2ex]
movie\_recommendation & Based on the movie list provided, carefully consider your preferences and make a well-informed decision. \\ [2ex]
multistep\_arithmetic\_two & First, simplify any expressions within parentheses following the correct order of operations to accurately evaluate the final answer with efficiency and precision. \\ [2ex]
navigate & Always face forward. Take 10 steps forward. Turn left. Take 5 steps forward. Take 3 steps backward. Finally, take 7 steps forward. Turn around and take 1 step forward. Repeat the previous sequence three times. Follow the given path precisely without any deviations. At the end, turn right and take 11 steps forward. If you follow these instructions, will you return to the starting point?
Options:
- Yes
- No \\ [2ex]
object\_counting & Determine the total count of mentioned vegetables accurately and state the final count as the answer. \\ [2ex]
penguins\_in\_a\_table & Analyze the given table to accurately determine the required information based on the provided criteria and attributes of the penguins and giraffes. Utilize efficient problem-solving strategies to arrive at the correct answer. \\ [2ex]
reasoning\_about \_colored\_objects & State the color of the object mentioned in the given arrangement with utmost accuracy. \\ [2ex]
ruin\_names & Choose the option that offers the most clever and humorous alteration of the given artist or movie name. Let your creativity shine and select the answer that will undoubtedly bring a smile to your face! Make sure to think outside the box! \\ [2ex]
salient\_translation \_error\_detection & Analyze the translation and accurately identify the specific error type based on the source text, providing the most appropriate corresponding option. \\ [2ex]
snarks & Choose the option that wickedly embodies sarcasm. \\ [2ex]
sports\_understanding & Determine the plausibility of the given statement by evaluating factual accuracy, logical consistency, and contextual relevance, then provide a succinct and well-justified response. \\ [2ex]
temporal\_sequences & Identify the optimal time slot for the individual to engage in the mentioned location/activity considering the given sightings and waking up time, taking into account the opening and closing times of the location and the duration of each event. \\ [2ex]
tracking\_shuffled\_objects \_seven\_objects & Pay attention to the given information and track the swaps/exchanges carefully to accurately determine the final possession/position/outcome for the specified individual. \\ [2ex]
web\_of\_lies & To determine the truthfulness of the last person mentioned, analyze the consistency of each statement and count the number of individuals accusing the previous person of lying. If the count of accusers is even, that person tells the truth; if it is odd, that person lies. \\ [2ex]
word\_sorting & Alphabetically sort the given list of words, ensuring all words are included and in ascending order. \\ [2ex]

\bottomrule
\end{tabular}
\end{center}
\label{table:found_instructions_on_bbh_tasks_s_text_bison_o_gpt_3.5_turbo_from_empty}
\end{table}

\newpage
\begin{table}[H]
\scriptsize
\caption{BBH task-wise Q\_end instructions found by prompt optimization with the \texttt{text-bison} scorer and the \texttt{gpt-3.5-turbo} optimizer.
The optimizations start from the empty string.
}
\begin{center}
\begin{tabular}{P{2.3cm}P{11.6cm}}
\toprule
Task & Our Instruction \\
\midrule
boolean\_expressions & Accurately use order of operations and parentheses to evaluate logical expressions and determine truth values efficiently. \\ [2ex]
causal\_judgement & Consider all relevant factors, prioritize overall well-being and ethical considerations, make well-informed decisions while foreseeing potential consequences efficiently, and consistently strive for optimal outcomes with empathy and adaptability in a thoughtful and comprehensive manner. \\ [2ex]
date\_understanding & Subtract the specified number of days from the given date and format the outcome as MM/DD/YYYY to accurately determine the desired result in an efficient manner. \\ [2ex]
disambiguation\_qa & Clearly identify and select the unambiguous antecedent for the pronoun or designate it as "Ambiguous" if it is unclear. \\ [2ex]
dyck\_languages & Add the missing closing parentheses. \\ [2ex]
formal\_fallacies & Determine the deductive validity of the argument presented based on the explicitly stated premises and reach a definitive conclusion. \\ [2ex]
geometric\_shapes & Analyzing the given SVG path element, accurately determine its shape by closely examining its curves and coordinates, then select the correct option. \\ [2ex]
hyperbaton & Choose the option with the correct adjective order in each sentence, prioritizing specific attributes like size, color, and origin. Place the most specific adjective before the more general ones for precise and standardized ordering across all examples. Ensure accurate alignment of the adjectives based on their respective attributes for consistent and standardized ordering. \\ [2ex]
logical\_deduction \_seven\_objects & Determine the precise order of the given objects/participants based on the provided information and establish the final ranking accurately, considering all relevant factors, while maintaining logical consistency with maximum efficiency. \\ [2ex]
movie\_recommendation & Choose the most similar option from the choices provided that closely aligns with the given movies' themes, genres, and impact for the most accurate recommendation possible. Make your selection wisely. \\ [2ex]
multistep\_arithmetic\_two & Carefully follow the order of operations to precisely simplify the expressions within parentheses and efficiently find the accurate final answer. \\ [2ex]
navigate & Always face forward. Take 10 steps forward. Turn right and walk for 5 steps. Then, make a left turn and continue for 9 steps. Proceed by walking 6 steps backward. Finally, turn around and take 200 steps. Accurately track your movements, diligently adhere to the given path, and ensure to return to the starting point without any deviations or obstacles. \\ [2ex]
object\_counting & Determine the total count of items mentioned, including all listed items, using an efficient and concise method. State the final count as your answer. \\ [2ex]
penguins\_in\_a\_table & Identify the animal with the maximum measurement (weight, age, or height) in the table and state its name and species. \\ [2ex]
reasoning\_about \_colored\_objects & Determine the color of each item in the given scenario and select the correct color option from the provided choices for accurate responses, ensuring utmost precision and completeness. \\ [2ex]
ruin\_names & Choose the option that creatively and hilariously transforms the given artist or movie name. \\ [2ex]
salient\_translation \_error\_detection & Carefully analyze the translations and select the most suitable option from the given choices to rectify the specific error category, ensuring complete precision, accuracy, and faithful representation of the intended meaning, while considering all relevant information in the source text. \\ [2ex]
snarks & Choose the option that cleverly employs sarcasm to defy all expectations and leave everyone utterly dumbfounded, questioning the very essence of their own perception. \\ [2ex]
sports\_understanding & Evaluate the plausibility of each given statement and provide a well-supported justification based on logical reasoning, contextual understanding, and relevant evidence to arrive at a definitive and conclusive answer. \\ [2ex]
temporal\_sequences & Identify the possible time slot for the desired activity based on the given information and sightings, then select the correct option. \\ [2ex]
tracking\_shuffled\_objects \_seven\_objects & Thoroughly analyze the given scenarios, systematically consider all available information, and confidently determine the final outcome with exceptional precision and optimal efficiency, while maintaining a strategic and logical approach throughout the process. \\ [2ex]
web\_of\_lies & Examine each person's statements meticulously to accurately determine the truth and confidently identify who is telling the truth, enabling you to effectively solve the given problem. \\ [2ex]
word\_sorting & Sort the given words alphabetically using spaces as separators while maintaining their original order and including all words. \\ [2ex]

\bottomrule
\end{tabular}
\end{center}
\label{table:found_instructions_on_bbh_tasks_s_text_bison_o_gpt_3.5_turbo_from_empty_q_end}
\end{table}

\newpage
\subsection{\texttt{PaLM 2-L} as scorer, \texttt{gpt-3.5-turbo} as optimizer, optimization starting from ``Let's solve the problem.''}
\label{appsec:bbh_taskwise_detailed_results_gpt_3.5_turbo_optimizer_start_from_solve}

\begin{figure}[H]
\centering
\scalebox{0.91}{
\subfigure[ours minus ``Let's think step by step.'']{\label{fig:accuracy_comparison_palm_2_l_found_minus_step_by_step_o_gpt_3.5_turbo_start_from_solve}\includegraphics[width=.48\linewidth]{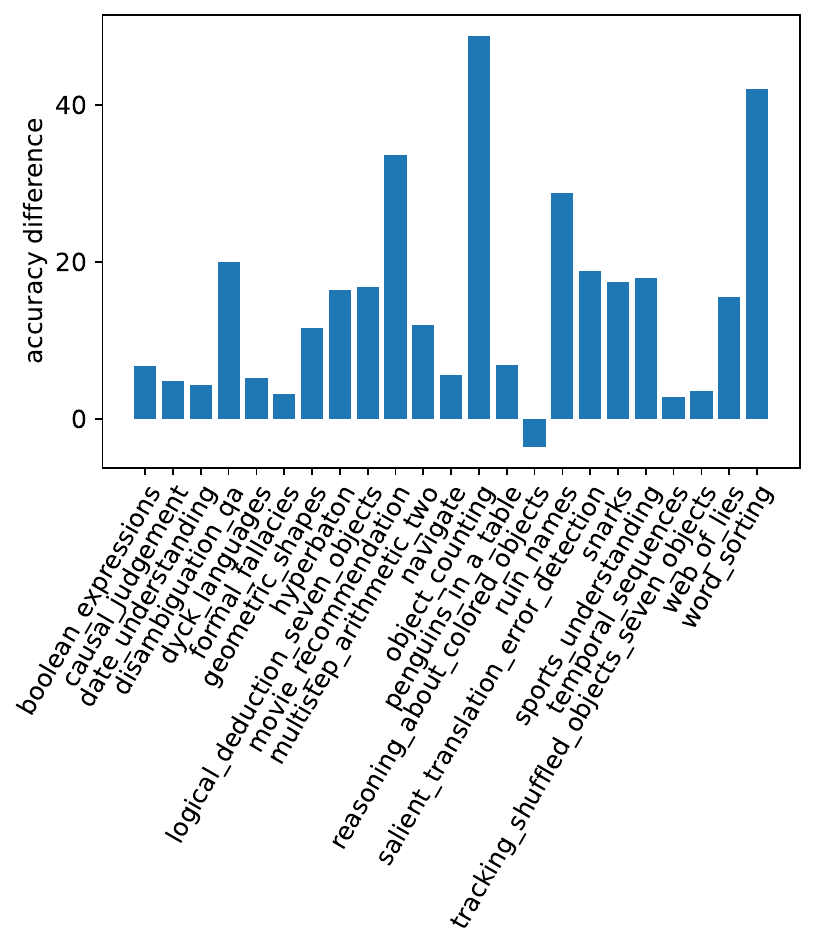}}
\hspace{.01\linewidth}
\subfigure[ours minus ``Let's solve the problem.'' starting point]{\label{fig:accuracy_comparison_palm_2_l_found_minus_empty_o_gpt_3.5_turbo_start_from_solve}\includegraphics[width=.48\linewidth]{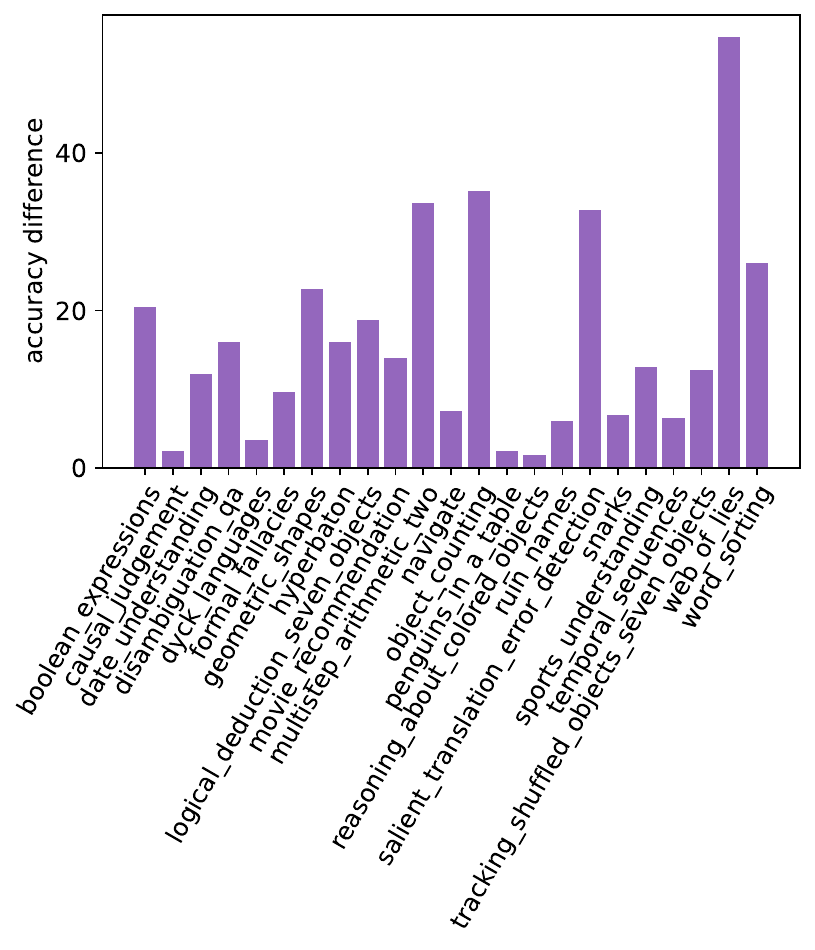}}
}

\scalebox{0.91}{
\subfigure[ours minus the instructions found with the empty starting point]{\label{fig:accuracy_comparison_text_bison_found_minus_found_from_empty_o_gpt_3.5_turbo}\includegraphics[width=.48\linewidth]{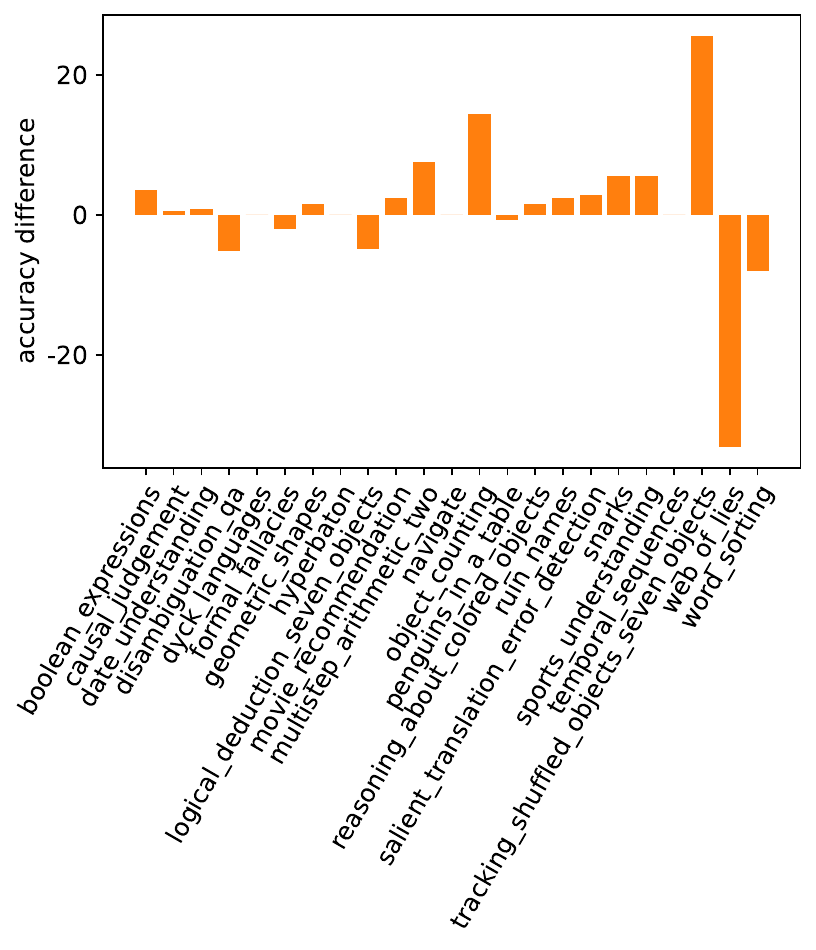}}
}

\caption{On 23 BBH tasks, the accuracy differences among instructions found by prompt optimization (with the \texttt{text-bison} scorer and the \texttt{gpt-3.5-turbo} optimizer), ``Let's think step by step.'', and ``Let's solve the problem.'' (optimization starting point).
The found instructions mostly outperform the ``Let's think step by step.'' baseline, the ``Let's solve the problem.'' starting point, and the instructions in Table~\ref{table:found_instructions_on_bbh_tasks_s_palm_2_l_o_gpt_3.5_turbo_from_empty} found by prompt optimization from the empty string.
}
\label{fig:accuracy_comparison_bar_charts_o_gpt_3.5_turbo_start_from_solve}
\end{figure}

Figure~\ref{fig:accuracy_comparison_bar_charts_o_gpt_3.5_turbo_start_from_solve} and Table~\ref{table:gpt_scores_on_bbh_tasks_starting_from_solve_the_problem} compare the accuracies of found instructions vs ``Let's solve the problem.'', ``Let's think step by step.'', and the instructions in Table~\ref{table:found_instructions_on_bbh_tasks_s_palm_2_l_o_gpt_3.5_turbo_from_empty}.
Table~\ref{table:found_instructions_on_bbh_tasks_s_palm_2_l_o_gpt_3.5_turbo_from_solve} details the found instructions.

The ``Let's'' pattern appears more often in the found instructions because of the starting points, and the instructions are more often declarative that are more suitable for A\_begin, even if some are semantically far from ``Let's solve the problem''.
In fact, ``Let's'' was adopted by~\citet{zhou2022large} as a fixed pattern in generated prompts, possibly because of the same reason.

\begin{table}[H]
\scriptsize
\caption{Accuracies on BBH tasks with the \texttt{PaLM 2-L} scorer and the \texttt{gpt-3.5-turbo} optimizer that starts from ``Let's solve the problem''.
The scores are from A\_begin instructions.
}
\begin{center}
\begin{tabular}{cP{1.2cm}P{2.2cm}P{2.2cm}}
\toprule
\multirow{2}{*}{Task} & \multirow{2}{*}{Scorer} & Our Acc & ``Let's solve the problem.'' Acc \\ \cmidrule{3-4}
& & training / test / overall & training / test / overall \\
\midrule
boolean\_expressions & \texttt{PaLM 2-L} & 98.0 / 89.5 / 91.2 & 78.0 / 69.0 / 70.8 \\
causal\_judgement & \texttt{PaLM 2-L} & 83.8 / 58.7 / 63.6 & 62.0 / 61.3 / 61.5 \\
date\_understanding & \texttt{PaLM 2-L} & 90.0 / 82.0 / 83.6 & 74.0 / 71.0 / 71.6 \\
disambiguation\_qa & \texttt{PaLM 2-L} & 78.0 / 68.0 / 70.0 & 52.0 / 54.5 / 54.0 \\
dyck\_languages & \texttt{PaLM 2-L} & 100.0 / 100.0 / 100.0 & 94.0 / 97.0 / 96.4 \\
formal\_fallacies & \texttt{PaLM 2-L} & 84.0 / 62.0 / 66.4 & 68.0 / 54.0 / 56.8 \\
geometric\_shapes & \texttt{PaLM 2-L} & 62.0 / 42.5 / 46.4 & 30.0 / 22.0 / 23.6 \\
hyperbaton & \texttt{PaLM 2-L} & 94.0 / 91.5 / 92.0 & 72.0 / 77.0 / 76.0 \\
logical\_deduction\_seven\_objects & \texttt{PaLM 2-L} & 66.0 / 53.0 / 55.6 & 38.0 / 36.5 / 36.8 \\
movie\_recommendation & \texttt{PaLM 2-L} & 88.0 / 88.0 / 88.0 & 66.0 / 76.0 / 74.0 \\
multistep\_arithmetic\_two & \texttt{PaLM 2-L} & 66.0 / 55.0 / 57.2  & 30.0 / 22.0 / 23.6 \\
navigate & \texttt{PaLM 2-L} & 76.0 / 67.0 / 68.8  & 54.0 / 63.5 / 61.6 \\
object\_counting & \texttt{PaLM 2-L} & 96.0 / 92.5 / 93.2 & 58.0 / 58.0 / 58.0 \\
penguins\_in\_a\_table & \texttt{PaLM 2-L} & 86.2 / 70.9 / 74.0 & 69.0 / 72.6 / 71.9 \\
reasoning\_about \_colored\_objects & \texttt{PaLM 2-L} & 88.0 / 69.0 / 72.8 & 78.0 / 69.5 / 71.2 \\
ruin\_names & \texttt{PaLM 2-L} & 92.0 / 85.5 / 86.8 & 76.0 / 79.5 / 80.8 \\
salient\_translation\_error\_detection & \texttt{PaLM 2-L} & 66.0 / 67.5 / 67.2 & 30.0 / 35.5 / 34.4 \\
snarks & \texttt{PaLM 2-L} & 88.6 / 76.9 / 79.2 & 80.0 / 70.6 / 72.5 \\
sports\_understanding & \texttt{PaLM 2-L} & 72.0 / 63.5 / 65.2 & 60.0 / 50.5 / 52.4 \\
temporal\_sequences & \texttt{PaLM 2-L} & 100.0 / 99.5 / 99.6 & 96.0 / 92.5 / 93.2 \\
tracking\_shuffled\_objects\_seven\_objects & \texttt{PaLM 2-L} & 56.0 / 63.5 / 62.0 & 42.0 / 51.5 / 49.6 \\
web\_of\_lies & \texttt{PaLM 2-L} & 56.0 / 58.5 / 58.0 & 0.0 / 4.0 / 3.2 \\
word\_sorting & \texttt{PaLM 2-L} & 52.0 / 44.5 / 46.0 & 18.0 / 20.5 / 20.0 \\
\bottomrule
\end{tabular}
\end{center}
\label{table:gpt_scores_on_bbh_tasks_starting_from_solve_the_problem}
\end{table}

\begin{table}[H]
\scriptsize
\caption{BBH task-wise Q\_begin instructions found by prompt optimization with the \texttt{PaLM 2-L} scorer and the \texttt{gpt-3.5-turbo} optimizer.
The optimizations start from ``Let's solve the problem''.
}
\centering
\begin{tabular}{P{2.3cm}P{11.6cm}}
\toprule
Task & Our Instruction \\
\midrule
boolean\_expressions & Let's accurately assess the given conditions and determine their corresponding Boolean values. \\ [2ex]
causal\_judgement & Let's conduct a meticulous evaluation of the given scenarios, accurately determine the causal relationships, and provide definitive answers through comprehensive analysis, ensuring a precise understanding of causation and a thorough determination of events in each situation. \\ [2ex]
date\_understanding & Let's accurately determine the correct date based on the given information and select the corresponding option in the standard MM/DD/YYYY format with utmost precision and reliability, ensuring the most definitive and reliable solution possible for accurate representation in all scenarios without any room for ambiguity, error, or confusion, and providing the highest level of accuracy and reliability. \\ [2ex]
disambiguation\_qa & Let's thoroughly analyze the given sentences to accurately determine the unambiguous antecedents of the pronouns used, ensuring clear understanding, effective communication, and leaving no room for any confusion or ambiguity. \\ [2ex]
dyck\_languages & Let's find the correct closing parentheses and brackets for the given sequences. \\ [2ex]
formal\_fallacies & Let's thoroughly analyze the explicitly stated premises and draw definitive conclusions to accurately determine the deductive validity of the arguments provided in each question, employing precise and logical reasoning in our assessments for unwavering confidence in our determinations. \\ [2ex]
geometric\_shapes & Let's accurately determine the shape represented by the given SVG path element by carefully analyzing its path data and considering all available options for a precise identification. \\ [2ex]
hyperbaton & Let's quickly identify the correct adjective order. \\ [2ex]
logical\_deduction \_seven\_objects & Let's methodically analyze the given information, employ logical reasoning, thoroughly evaluate all relevant details, and accurately determine the solutions for each problem by considering all provided options comprehensively and strategically, ensuring an efficient and effective approach towards arriving at the correct answers. \\ [2ex]
movie\_recommendation & Let's uncover the perfect movie recommendation from the options provided, ensuring an exceptional cinematic experience together as we select the most captivating and satisfying choice that will keep us thoroughly engaged and immersed until the very end. \\ [2ex]
multistep\_arithmetic\_two & Let's tackle the following calculations. \\ [2ex]
navigate & Let's accurately and efficiently determine the correct solution for each given scenario, ensuring the highest level of precision, reliability, and consistency throughout. \\ [2ex]
object\_counting & Let's determine the total count of various items/objects/ingredients/animals mentioned in order to accurately and efficiently find the answer. \\ [2ex]
penguins\_in\_a\_table & Let's analyze the given information and determine the correct answer. \\ [2ex]
reasoning\_about \_colored\_objects & Let's systematically analyze the given information and carefully evaluate each answer choice to confidently determine the accurate and optimal solutions, considering all available options and specific details provided in each question for precise and concise responses, ensuring complete accuracy and clarity in our answers. \\ [2ex]
ruin\_names & Prepare to have a side-splittingly funny time as we uncover the most clever and hilarious alternatives for these artist or movie names, challenging your wit to guess the correct one with a burst of creativity, humor, and imaginative twists! \\ [2ex]
salient\_translation \_error\_detection & Let's meticulously analyze the provided translations, accurately identifying any errors or discrepancies, and conduct a comprehensive evaluation to ensure the highest level of translation quality and fidelity. By considering contextual nuances, cultural references, linguistic conventions, potential factual errors, and any dropped content, our ultimate aim is to achieve precise and thorough assessments for optimal translation accuracy and adherence to the source text. \\ [2ex]
snarks & Let's expertly determine the sarcastic statement among the given options and confidently provide the definitive answer without any room for doubt or confusion, ensuring absolute precision, clarity, and unwavering expertise in our response, while carefully analyzing the context, tone, and intention behind each statement to achieve unrivaled accuracy and unwavering confidence. \\ [2ex]
sports\_understanding & Let's find the accurate information. \\ [2ex]
temporal\_sequences & The flawless approach \\ [2ex]
tracking\_shuffled\_objects \_seven\_objects & By meticulously analyzing the given scenarios and accurately determining the final outcomes through a series of trades, swaps, and exchanges among the individuals involved, let's ascertain the conclusive results. \\ [2ex]
web\_of\_lies & Let's scrutinize each statement provided to accurately determine the truth-teller and uncover the veracity behind their words with unwavering analysis. \\ [2ex]
word\_sorting & Employing efficient and precise measures, sort the given list of words in alphabetical order to provide an optimal solution for any sorting problem, ensuring maximum performance and effectiveness. \\ [2ex]

\bottomrule
\end{tabular}
\label{table:found_instructions_on_bbh_tasks_s_palm_2_l_o_gpt_3.5_turbo_from_solve}
\end{table}

%%%%%%%%%%%%%%%%%%%%%%%%%%%%%%%%%%%%%%%%%%%%%%%%%%%%%%%%%%%%
\end{document}